\PassOptionsToPackage{shortlabels}{enumitem}
\documentclass[11pt, a4paper]{include/gdm_format}

\usepackage{palatino}
\usepackage{authblk}

\usepackage[authoryear, sort&compress, round]{natbib}
\usepackage{xcolor}
\usepackage{graphicx}
\usepackage{booktabs}
\usepackage{url}
\usepackage{xurl}
\usepackage[parfill]{parskip}
\usepackage{hyperref}
\usepackage{textgreek}
\usepackage{cleveref}

\usepackage{tcolorbox}
\tcbset{summary/.style={colback=red!10!white, colframe=red!10!black, fonttitle=\bfseries, title=Summary}}
\DeclareUnicodeCharacter{0394}{\ensuremath{\Delta}}

\usepackage{hyunin}
\usepackage{algorithm2e}
\usepackage{amsmath, amsthm, amssymb, amsfonts}
\usepackage{mathtools}
\usepackage{bbm}
\usepackage{extarrows}
\usepackage{caption}
\usepackage{subcaption}
\usepackage{enumitem}
\usepackage{minted}
\tcbuselibrary{minted, breakable}
\usepackage{fvextra}

\newtcblisting{jsonbox}[2][]{
  listing engine=minted,
  colback=white,
  colframe=black,
  boxrule=0.5pt,
  arc=2pt,
  outer arc=2pt,
  boxsep=2mm,
  listing only,
  breakable,
  minted language=json,
  minted options={
    tabsize=4,
    fontsize=\small,
    breaklines,
    linenos=false
  },
  title={#2},
  #1
}

\tcbset{
  colback=gray!10,   
  colframe=gray!50,  
  boxrule=0.5pt,
  arc=4pt,           
}

\newcommand{\singledefault}{\texttt{single(default)}}
\newcommand{\multidefault}{\texttt{multi(default)} }
\newcommand{\multiours}{\texttt{multi(ours)}}
\newcommand{\singleunion}{\texttt{single(unionOur)} }

\newcommand{\hyunin}[1]{}
\newcommand{\yujin}[1]{}

\usepackage{tikz}
\definecolor{crownface}{HTML}{F4C542}
\definecolor{crownedge}{HTML}{B8860B}
\DeclareRobustCommand{\rhicrown}{\mbox{\tikz[x=1.5ex,y=1.5ex,baseline=-0.675ex]{%
  \fill[fill=crownface,draw=crownedge,line width=0.25pt]
    (-0.50,-0.45)--(0.50,-0.45)--(0.50,0.50)--(0.20,-0.05)--%
    (0.00,0.45)--(-0.20,-0.05)--(-0.50,0.50)--cycle;}}}

\title{Recursive Harness Self-Improvement}

\author[1,2]{Hyunin Lee}
\author[1]{Jinglue Xu}
\author[1]{Jeffrey Seely}
\author[2]{Donghyun Lee}
\author[2]{Matei Zaharia}
\author[1]{Yujin Tang}

\affil[1]{Sakana AI}
\affil[2]{UC Berkeley}

\begin{document}


\begin{abstract}
Under model--harness co-evolution, harnesses are not merely inference-time scaffolds but data-generating components whose execution traces can shape future foundation models. This motivates \emph{harness-in-the-loop} learning: optimizing harnesses for both immediate agent performance and the quality of traces used for future model training.
However, continually updating provider-built scaffolds is costly and labor-intensive. We therefore investigate whether optimizing user-constructed harnesses in a task-specific manner can improve execution-trace quality while remaining computationally lightweight and requiring only a few update iterations.
To this end, we introduce \emph{Recursive Harness Self-Improvement} (RHI), which represents the harness as a prompt-level specification of the agent loop and iteratively refines it using pairwise feedback over its own revision history. Across 30 synthetic machine-learning research tasks spanning quantitative finance, robotics, and pharmacy, a few RHI iterations suffice to substantially raise the performance ceiling of low-reasoning-effort agents, exceeding the corresponding maximum-reasoning-effort setting while reducing inference cost by up to 60\%. We show that these gains arise primarily from improved task-specific context management through more effective inter-agent information flow rather than longer reasoning traces. Finally, we formalize this behavior as an information-theoretic hypothesis for RHI's implicit optimization objective, suggesting RHI as a practical algorithm for continual learning within the paradigm of model--harness co-evolution.

\end{abstract}

\begingroup
\renewcommand{\thefootnote}{\fnsymbol{footnote}}
\maketitle
\footnotetext[1]{Work initiated and done during an internship at Sakana AI. Corresponding author: \texttt{hyunin@sakana.ai}}
\endgroup
\setcounter{footnote}{0}
\renewcommand{\thefootnote}{\arabic{footnote}}

\begin{figure}[h]
    \centering
    \includegraphics[width=\linewidth]{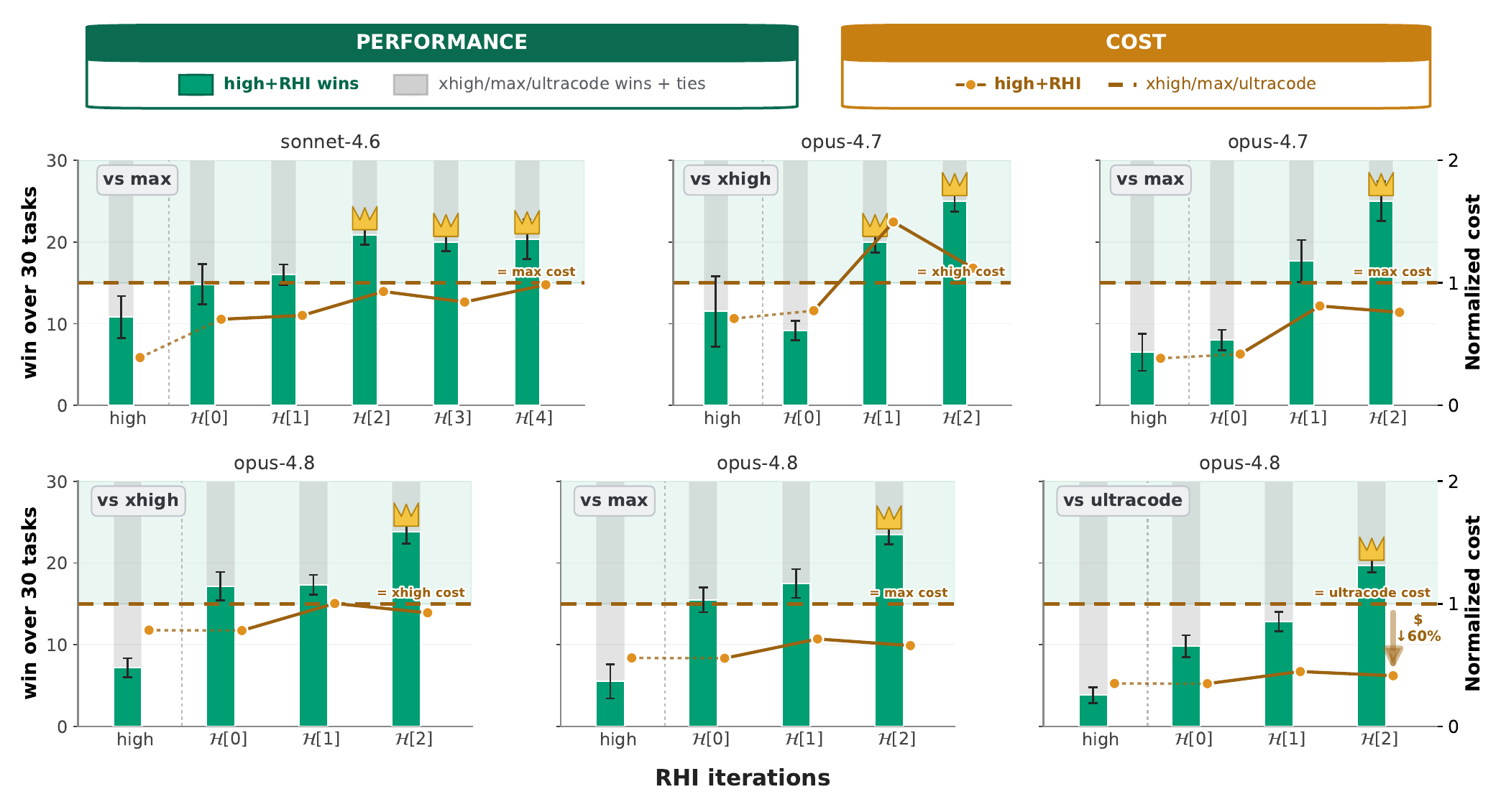}
    \caption{\textbf{Few-shot RHI raises the performance plateaus of test-time scaling.} Across $30$ synthetic ML research tasks, \texttt{high}-reasoning \texttt{sonnet-4.6}, \texttt{opus-4.7}, and \texttt{opus-4.8} agents with RHI-improved harnesses achieve higher pairwise win counts than evaluated test-time scaling baselines selected from \texttt{xhigh}, \texttt{max}, and \texttt{ultracode}. Bars report mean wins over two LLM judges and three seeds. A crown~(\rhicrown)~marks mean wins above $19.5$ out of $30$.}
    \label{fig:rhi_vs_tts_frontpage}
\end{figure}

\clearpage      

\tableofcontents

\clearpage      

\section{Introduction}

Scaling progress in AI is typically attributed to larger models \citep{kaplan2020scaling}, increased training compute \citep{hoffmann2022chinchilla}, and improved post-training \citep{deepseekai2025r1}. Increasingly, however, advances in deployed AI systems arise from the \emph{co-evolution} of foundation models and harnesses, rather than from foundation models alone \citep{anthropic2026harness_1,openai2026harness_1}. This co-evolution requires a \emph{recursive feedback loop}: for a fixed foundation model, stronger harnesses organize more effective agent workflows that generate substantially higher-quality execution traces than the standalone model can produce, which can subsequently serve as training data for future foundation models \citep{lin2026agentic_harness_engineering}. Modern coding agents generate such high-quality execution traces by leveraging harnesses from two complementary sources: providers ship built-in, general-purpose harnesses for coding and multi-agent workflows \citep{openai_codex_cli_2025,anthropic_claude_code_2025,openai_swarm,langgraph,google_adk}, while users further specialize these systems with task-specific harnesses, including reusable skills, subagents, and automation workflows \citep{openai_codex_agent_loop_2026,anthropic_claude_code_memory_2026,anthropic_claude_code_skills_2026,anthropic_claude_code_subagents_2026,n8n_ai_agents_2026}. Consequently, a key bottleneck in this harness–model co-evolutionary loop is the quality of the execution traces produced by the harness. 
This work focuses on the \emph{first half of this recursive loop}: for a fixed foundation model, how to improve the harness to generate higher-quality execution traces.

\hyunin{Explain that the co-evolution of the harness-model loop is the future of AI system development. Then, explain the trace quality produced by the harness is a bottleneck in that loop, so we have to improve the harness.}

Improving harnesses, however, is challenging in practice. Provider-built harnesses must generalize across diverse users and tasks, making continual updates prohibitively labor- and cost-intensive \citep{zhang2024aflow,shang2024agentsquare,zhang2026selfharness,anthropic2026harness_1,openai2026harness_1}. By contrast, \emph{user-constructed harnesses} can be optimized for individual tasks, making them a practical target for improving execution-trace quality \citep{openai_codex_agent_loop_2026,anthropic_claude_code_skills_2026,anthropic_claude_code_subagents_2026,n8n_ai_agents_2026}. Moreover, for such optimization to be practical, each update should be \emph{computationally lightweight}, and the optimization process should converge within only a \emph{few update iterations}, enabling users to specialize agents to many open-ended tasks under limited per-task budgets. Yet how to achieve such lightweight, iterative harness optimization remains largely unexplored. To address this execution-trace quality bottleneck, we investigate whether a \emph{few iterations} of \emph{lightweight} optimization of a \emph{user-constructed harness} can raise the performance plateaus of agent test-time scaling \citep{wang2025scaling}.

\hyunin{Explain why such harness improvement should be tailored to few iterations of lightweight updates to a user-constructed harness. Then, explain our approach to tackle that data quality bottleneck.}

To this end, we introduce \emph{Recursive Harness Self-Improvement} (RHI), which defines the harness as the agent loop itself, including the roles assigned to agents, the instructions they follow, the information exchanged between agents (contracts), and the workflow structure governing when reasoning is invoked (hops). Roles and instructions define task allocation, contracts specify inter-agent communication, and hops determine the control flow of the orchestrator--subagent workflow. RHI then iteratively refines the harness using pairwise feedback over its own revision history (Section~\ref{sec:RHI}). First, to enable efficient optimization of \emph{user-constructed harnesses}, our key design choice is to treat the \emph{harness as a prompt-level object} rather than executable code. Second, RHI is \emph{lightweight} because it replaces expensive population-level harness search with \emph{trajectory-local self-comparison}: at each iteration, the current harness is compared with its immediate predecessor, and the resulting preference history guides the next revision. Third, RHI requires only a \emph{few update iterations} to substantially \emph{raise the performance plateaus} of a low-reasoning-effort agent, often exceeding the corresponding maximum-reasoning-effort setting.

\hyunin{Introduce RHI, then explain why RHI satifies above conditions of harness optimization (lightweight, few-shot, user-side harness)}

We evaluate RHI on 30 synthetic, open-ended machine-learning research tasks spanning quantitative finance, robotics, and pharmacy. Each task requires generating a complete code repository with standardized deliverables, and outputs are evaluated using pairwise LLM-as-a-judge comparisons. Across \texttt{sonnet-4.6}, \texttt{opus-4.7}, and \texttt{opus-4.8}, few-shot RHI applied to \texttt{high}-reasoning agents achieves higher win rates than all stronger test-time-scaling settings, including \texttt{xhigh}, \texttt{max}, and \texttt{ultracode}. These gains persist across progressively stronger foundation models. For \texttt{opus-4.8}, RHI also reduces inference cost by up to $60\%$ relative to the \texttt{ultracode} baseline (Figure~\ref{fig:rhi_vs_tts_frontpage}; Section~\ref{sec:Experiment}). Our ablation studies show that these gains do not arise from generating longer outputs: across RHI iterations, output-token usage remains nearly constant while performance improves, and cache read/write usage and inference cost often decrease. Instead, the results suggest that RHI improves task-specific context management: communication contracts and workflow hops become more task-specific, task-relevant information encoded in these components increases, and redundancy across harness components decreases (Section~\ref{sec:Ablation}). Furthermore, we observe that RHI’s learning dynamics are consistent with an implicit objective that reorganizes information flow across harness components. We formalize this observation as an information-theoretic hypothesis for RHI’s implicit update objective--increasing the mutual information between each component and the task while penalizing redundancy across components--yielding a testable account of RHI's implicit updates.

\hyunin{Explain the experiment results and ablation study}

Overall, this paper argues that future progress in harness–model co-evolution should focus on improving the data flywheel. To advance the first half of this loop, which optimizes user-constructed harnesses for task-specific execution trace generation. We envision RHI as a practical framework that enables users to continually specialize coding agents for diverse open-ended tasks encountered in everyday use.

\hyunin{Position what this paper argues, then explain what our hope is for world-wide users to utilize this work.}

\section{Preliminary}

We first define the harness optimization problem in Section~\ref{sec:Problem definition} and then review existing methods in Section~\ref{subsec:relation_to_existing_harness_search_objectives}.

\subsection{Problem definition}
\label{sec:Problem definition}
\hyunin{Here, I define the problem}

Formally, let $\mathcal{V}$ denote the vocabulary (token) space, and let $\mathcal{V}^* := \cV \times \cV \times \cdots$ denote the set of all finite-length token sequences (i.e., sentences). Let $\cA$ be a coding agent with a fixed language model $\mathcal{L}$, and let $x \in \mathcal{X} \subset \mathcal{V}^*$ denote a task prompt. Let $\mathcal{H}$ denote the space of harnesses, and let $\mathcal{Y}$ denote the space of code repositories, where each repository is represented as a collection of file-path/content pairs. Given a harness $H \in \mathcal{H}$, the corresponding agent produces an output $y \sim \mathcal{A}(H,x)$, where $y \in \cY$ is the repository generated by the agentic execution procedure induced by the language model $\mathcal{L}$, the harness $H$, and the task $x$.

Our evaluation protocol is motivated by open-ended coding tasks encountered in everyday use, where the output is a complete code repository rather than a \emph{single, scalar} verifiable answer. Such repositories cannot be reliably evaluated using a single metric because they must be assessed across \emph{multiple criteria}, including functional correctness, task alignment, reproducibility, and code quality. Moreover, these criteria are often difficult to aggregate into a single numerical score but are naturally expressed through \emph{pairwise preferences} between outputs. We therefore adopt a pairwise evaluation protocol conditioned on an evaluation prompt $x_{\mathrm{eval}} \in \mathcal{V}^*$ that multiple criteria. Given two outputs $y_1, y_2 \in \mathcal{Y}$, the evaluator returns one of three possible judgments:
$
\mathcal{L}_{\mathrm{eval}}(y_1,y_2,x_{\mathrm{eval}})
\in
\{y_1 \succ y_2, y_1 \sim y_2, y_2 \succ y_1\},
$
where $y_1 \succ y_2$ denotes that $y_1$ is preferred to $y_2$ under the evaluation criteria specified by $x_{\mathrm{eval}}$, and $y_1 \sim y_2$ denotes that the two outputs are judged equivalent. Here, $\mathcal{L}_{\mathrm{eval}}$ denotes an LLM-based evaluator, which may be different from the coding agent's underlying language model $\mathcal{L}$.

Given a task $x \in \mathcal{X}$ and an evaluation prompt $x_{\mathrm{eval}}$, our ideal objective is to identify a task-specific harness whose outputs are preferred to those generated by competing harnesses. A natural population objective is
\begin{equation}
\begin{aligned}
H_x^*
\in
\argmax_{H \in \mathcal{H}} \; &f_x(H), \\
\qquad
f_x(H)
=
\mathbb{E}_{\substack{
H' \sim \mu(\cdot \mid H' \neq H),\
y \sim \mathcal{A}(H,x),\;
y' \sim \mathcal{A}(H',x)
}}
&\left[
\mathbf{1}
\left\{
\mathcal{L}_{\mathrm{eval}}(y,y';x_{\mathrm{eval}})
=
y \succ y'
\right\}
\right],
\end{aligned}
\label{eq:ideal_objective}
\end{equation}

where $\mu$ is a reference distribution over competing harnesses, and $\mathbf{1}{\cdot}$ denotes the indicator function, which equals $1$ if the enclosed predicate is true and $0$ otherwise. The function $f_x(H)$ denotes the expected pairwise win rate of harness $H$ against competing harnesses under the evaluation prompt $x_{\mathrm{eval}}$. Accordingly, $H_x^*$ is the harness that maximizes this expected win rate.

\subsection{Existing methods}
\label{subsec:relation_to_existing_harness_search_objectives}
\hyunin{Here, I explain how existing work have relaxed Equation \eqref{eq:ideal_objective} -- approach with population-based method}
Exact maximization of Equation~\eqref{eq:ideal_objective} would require searching over a discrete, high-dimensional harness space and estimating the expected pairwise win rate of each candidate against harnesses drawn from the reference distribution $\mu$. Prior work typically makes this optimization tractable in two ways: by restricting the search space to a concrete representation, such as prompts, workflow graphs, tool-use policies, or executable harness code, and by replacing the expectation over $\mu$ with a finite candidate set that can be evaluated.

In the notation of Equation~\eqref{eq:ideal_objective}, population-based harness, workflow, prompt, or program search methods can be viewed as replacing the reference distribution $\mu$ with a sampled population $\mathcal{S}_i=\{H_{i,1},\ldots,H_{i,m}\}$ and estimating
\begin{equation}
\widehat f_x(H;\mathcal{S}_i)
=
\frac{1}{|\mathcal{S}_i|-1}
\sum_{H' \in \mathcal{S}_i \setminus \{H\}}
\mathbb{E}_{y \sim \mathcal{A}(H,x),\; y' \sim \mathcal{A}(H',x)}
\left[
\mathbf{1}\{\mathcal{L}_{\mathrm{eval}}(y,y';x_{\mathrm{eval}})=y\succ y'\}
\right].
\label{eq:population_approximation}
\end{equation}
Different algorithms instantiate the population $\mathcal{S}_i$ at different levels of abstraction. 
At the harness level, Meta-Harness treats the harness code itself as the candidate object and proposes new harnesses from the source, scores, and execution traces of prior candidates \citep{lee2026meta}, 
while AutoHarness automatically synthesizes a code harness and refines it from execution feedback \citep{lou2026autoharness}. 
TTHE applies the same harness-level pattern at test time: on each unlabeled test batch, it instantiates $\mathcal{S}_i$ as a population of parallel harness branches evolved over multiple rounds, and an agentic judge commits one branch to the next batch using execution-derived proxy signals as the scoring rule \citep{nie2026tthe}.
At the agent and workflow level, ADAS, through Meta Agent Search, maintains an archive of code-defined agents and uses a meta-agent to program new agents from that archive \citep{hu2024adas}; 
GPTSwarm represents agents as optimizable graphs and updates node prompts and graph connectivity \citep{zhuge2024gptswarm}; 
and AFlow searches over code-represented workflows using Monte Carlo tree search with code modification and execution feedback \citep{zhang2024aflow}. 
At the program level, AlphaEvolve and ShinkaEvolve instantiate the same pattern over executable code: an LLM mutates or rewrites candidates, an evaluator scores them, and selection or parent sampling decides which candidates seed the next generation \citep{novikov2025alphaevolve,lange2025shinkaevolve}. 
In each case, the method does not optimize Equation~\eqref{eq:ideal_objective} exactly. 
Instead, it constructs a finite candidate set $\mathcal{S}_i$ and ranks its members with a task-specific score, evaluator, or pairwise judgment; 
when the scoring rule is order-consistent with the preference $\succ$, this ranking recovers the within-population win rate $\widehat f_x(\cdot;\mathcal{S}_i)$ of Equation~\eqref{eq:population_approximation} up to a monotone transformation, so selecting or evolving the top candidates approximately maximizes the finite-population approximation before reseeding $\mathcal{S}_{i+1}$.

Prompt and pipeline optimizers can be interpreted as the same finite-population approximation with a more restricted feasible set $\mathcal{S}_i$ of textual prompts or modular LLM programs. 
OPRO treats the prompt as the optimization variable and asks an LLM optimizer to propose improved prompts from previous prompt--score pairs \citep{yang2024opro}. 
TextGrad converts textual feedback into natural-language gradients that revise text variables \citep{yuksekgonul2024textgrad}. 
DSPy compiles modular LM programs by searching over prompts and few-shot demonstrations \citep{khattab2024dspy}. 
GEPA maintains a Pareto frontier of prompt candidates, reflects on execution trajectories, and proposes, tests, and recombines prompt edits \citep{agrawal2025gepa}, 
and optimize\_anything generalizes this reflective text-optimization view to arbitrary text parameters \citep{agrawal2026optimize_anything}. 
All of these are natural approximations to Equation~\eqref{eq:ideal_objective} when many candidates can be instantiated and evaluated. 
Their common limitation in our setting is that the fidelity of $\widehat f_x(\cdot;\mathcal{S}_i)$ grows with the size and diversity of $\mathcal{S}_i$, yet every additional candidate requires a fresh, expensive black-box agent execution and evaluation.

Such population-based search is ill-suited to user-constructed harness optimization because every additional candidate requires a full black-box agent execution and evaluation. For users continually specializing agents to many open-ended tasks, this cost is prohibitive. Indeed, \citet{wang2026rethinking} show that once this search cost is properly
accounted for, automatic harness evolution fails to consistently outperform
simple test-time scaling baselines such as parallel sampling and sequential
refinement under fixed inference budgets. Harness optimization is
therefore worthwhile only when its search overhead is far smaller than that
of population-based methods. Consequently, our setting calls for an optimization method that is \emph{computationally lightweight}, converges within only a \emph{few update iterations}, and nevertheless yields \emph{substantial performance improvements}.

\section{Recursive Harness Self-Improvement}
\label{sec:RHI}
\begin{figure}[h]
    \centering
    \includegraphics[width=1.0\linewidth]{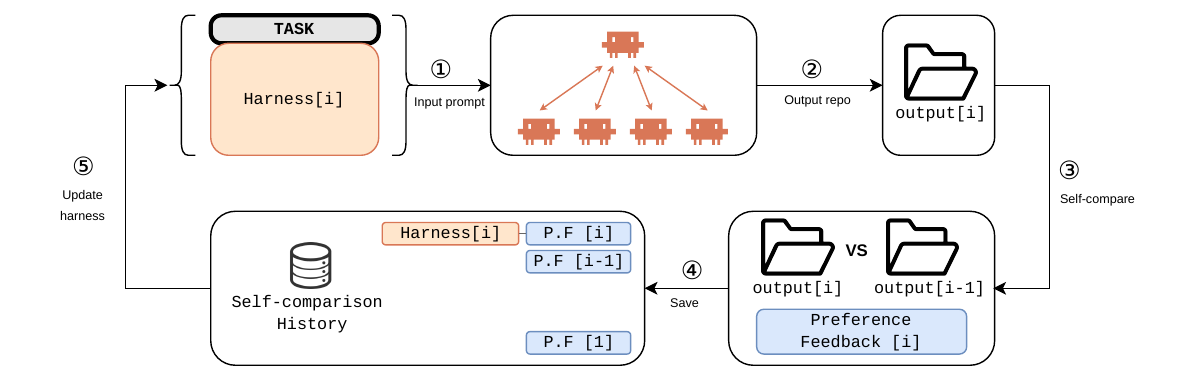}
    \caption{Recursive Harness Self-Improvement. \textcircled{1}, \textcircled{2}: At iteration $i$, the coding agent receives the task $x$ and the current harness $H^{(i)}$, solves the task, and produces an output $\texttt{output}[i]$. \textcircled{3}: An LLM evaluator then compares $\texttt{output}[i]$ with the previous output $\texttt{output}[i-1]$ and returns preference $\texttt{P.F}[i]$. \textcircled{4}: This feedback is stored in the self-comparison history. \textcircled{5}: Finally, an LLM harness optimizer uses this history to update the harness from $H^{(i)}$ to $H^{(i+1)}$.}
    \label{fig:rhsi}
\end{figure}

This section introduces our proposed algorithm, Recursive Harness Self-Improvement (RHI). Section~\ref{subsec:rhi_as_on_trajectory_relaxation} presents a computationally lightweight relaxation of the RHI objective, Section~\ref{subsec:objective_function} describes the RHI algorithm, and Section~\ref{sec:Harness} defines the harness representation used by RHI.
Figure~\ref{fig:rhsi} provides a high-level overview of RHI, and Algorithm~\ref{alg:rhsi} presents the complete procedure.

\subsection{RHI objective function}
\label{subsec:rhi_as_on_trajectory_relaxation}
\hyunin{Here, I explain our approach on-trajectory relaxation and explain (1) cost-benefit (2) rebuttal to whether our approach is too much relaxation (3) to avoid (2), we need self-history. (3) becomes a hook to next subsection}

Unlike the methods discussed in Section~\ref{subsec:relation_to_existing_harness_search_objectives}, RHI adopts a trajectory-local relaxation of Equation~\eqref{eq:ideal_objective}. At iteration $i$, it replaces the broad competitor distribution $\mu$ with a local competitor distribution concentrated on the previous harness, $\nu_i=\delta_{H_x^{(i-1)}}$. The corresponding local preference objective is
\begin{equation}
\begin{aligned}
\widetilde f_x^{(i)}(H)
=
\mathbb{E}_{y \sim \mathcal{A}(H,x),\; y^- \sim \mathcal{A}(H_x^{(i-1)},x)}
\left[
\mathbf{1}\{\mathcal{L}_{\mathrm{eval}}(y,y^-;x_{\mathrm{eval}})=y\succ y^-\}
\right].
\end{aligned}
\label{eq:rhi_local_objective}
\end{equation}
This trajectory-local objective is not an unbiased estimator of $f_x(H)$ under the original $\mu$; rather, it deliberately changes the problem from global comparison against arbitrary harnesses to local improvement over the system's own previous harness. This distinction separates Meta-Harness \citep{lee2026meta} from Self-Harness \citep{zhang2026selfharness}. Meta-Harness is better viewed as optimizing the finite-candidate approximation in Equation~\eqref{eq:population_approximation}: it maintains a population and Pareto frontier of evaluated harnesses, stores each candidate's source code, scores, and execution traces in a filesystem, and imposes no parent-selection rule, so the proposer may inspect any prior harness and trace when proposing a new harness \citep{lee2026meta}. It therefore does not instantiate Equation~\eqref{eq:rhi_local_objective}, whose comparator is only the immediately previous harness $H_x^{(i-1)}$. Self-Harness is closer to Equation~\eqref{eq:rhi_local_objective}: in each round, the current harness is evaluated, failures are mined from its execution traces, candidate edits are proposed, and each candidate is regression-tested against the current harness on held-in and held-out splits \citep{zhang2026selfharness}. However, Self-Harness is not identical to RHI's trajectory-local objective. It may evaluate multiple candidate branches and merge compatible accepted edits, and its comparison signal is a verifier-based pass-count rule---improve at least one split without degrading the other---rather than an LLM pairwise preference between two generated outputs. Thus, Self-Harness can be interpreted as a deterministic, verifier-based analogue of Equation~\eqref{eq:rhi_local_objective} only after replacing the preference relation $\succ$ with the regression-test acceptance relation; without that substitution, it is local in structure but not a direct solution of the RHI objective.

\paragraph{RHI is computationally lightweight.}
Equation \eqref{eq:rhi_local_objective} relaxation is \emph{computationally lightweight} because it substantially reduces the cost of both \emph{agent execution} and \emph{pairwise evaluation} (Table \ref{tab:complexity}). Let the computational cost of one optimization step be $C = N_{\mathrm{trace}} + N_{\mathrm{pair}}$, where $N_{\mathrm{trace}}$ is the number of coding-agent executions (calls to $\mathcal{A}$) and $N_{\mathrm{pair}}$ is the number of pairwise evaluations (calls to $\mathcal{L}_{\mathrm{eval}}$). We assume that each harness output is cached after generation and reused for subsequent comparisons.

\begin{table}[h]
\centering
\caption{Per-iteration computational cost of three harness-search objectives. Here $M$ denotes the effective harness population required to approximate the ideal objective, while $m \ll M$ is the sampled population size used by finite-population search.}
\label{tab:complexity}
\begin{tabular}{lccc}
\toprule
Objective & $N_{\mathrm{trace}}$ & $N_{\mathrm{pair}}$ & Total Cost \\ \midrule
Ideal objective (Eq.~\ref{eq:ideal_objective})
&
$M$
&
$\binom{M}{2}$
&
$\Theta(M^2)$
\\

Finite-population search (Eq.~\ref{eq:population_approximation})
&
$m$
&
$\binom{m}{2}$
&
$\Theta(m^2)$
\\

\textbf{Trajectory-local RHI} (Eq.~\ref{eq:rhi_local_objective})
&
$\boldsymbol{1}$
&
$\boldsymbol{1}$
&
$\boldsymbol{\Theta(1)}$
\\
\bottomrule
\end{tabular}
\end{table}

The ideal objective compares each candidate harness against the entire reference population, requiring one execution trace per harness and pairwise comparisons between all competing pairs. Finite-population methods reduce this cost by restricting the comparison to a sampled population of size $m$, but still require quadratic pairwise evaluations. In contrast, RHI compares the current harness only with the cached output of its immediate predecessor, requiring exactly one new execution trace and one pairwise evaluation per iteration. Consequently,
\[
C_{\mathrm{ideal}}
>
C_{\mathrm{pop}}
>
C_{\mathrm{RHI}},
\]
whenever $M \gg m \gg 1$. Scaling to $n$ independent tasks multiplies all three costs by $n$ without changing their ordering.

\paragraph{RHI performs noisy local ascent.}

One may wonder whether the relaxation in Equation~\eqref{eq:rhi_local_objective} is too aggressive. Under a standard pairwise-preference model, however, the trajectory-local objective preserves the same latent utility ordering as the ideal population objective. Specifically, suppose there exists a task utility $u_x:\mathcal{H}\rightarrow\mathbb{R}$ and a strictly increasing link function $\sigma$ satisfying $\sigma(0)=\tfrac12$ such that $\Pr(H \succ H')=\sigma\!\left(u_x(H)-u_x(H')\right)$. If the competitor distribution is independent of the candidate harness (or if the conditioning $H'\neq H$ is treated as a negligible no-self-comparison correction), then both objectives are monotone functions of the same latent utility:
\begin{equation*}
f_x(H)
=
\mathbb{E}_{H'\sim\mu}
\!\left[
\sigma\!\left(u_x(H)-u_x(H')\right)
\right], \qquad
\widetilde f_x^{(i)}(H)
=
\sigma\!\left(u_x(H)-u_x(H_x^{(i-1)})\right).
\end{equation*}
Consequently, maximizing the trajectory-local objective targets the same utility ordering as maximizing the ideal population objective. Moreover, because $H_x^{(i-1)}$ is fixed during iteration $i$, any revision whose true win probability against $H_x^{(i-1)}$ exceeds $\tfrac12$ satisfies $u_x(H)>u_x(H_x^{(i-1)})$, and therefore increases the ideal objective. The observed pairwise comparison thus provides a \emph{noisy local ascent signal}: winning revisions are encouraged, while losing revisions are discarded or revised. 

\paragraph{Necessity of self-history for few-iteration updates.}

Since each pairwise comparison provides only a noisy local ascent signal, RHI accumulates preference feedback across iterations to improve the reliability of harness optimization under a limited update budget. Specifically, the history
\begin{equation}
\mathcal{D}_x^{(i)}
=
\left\{
\mathcal{L}_{\mathrm{eval}}
\!\left(
y_x^{(k)},
y_x^{(k-1)};
x_{\mathrm{eval}}
\right)
\right\}_{k=1}^{i},
\label{eq:self-history-dataset}
\end{equation}
where $y_x^{(k)} \sim \mathcal{A}(H_x^{(k)},x)$ is the repository generated by the $k$-th harness, and each element records the pairwise preference between consecutive harness revisions. Here, $H_x^{(k)}$ denotes the task-specific harness for task $x$ after $k$ RHI iterations. Throughout the paper, when the task is clear from context, we write $H^{(k)}$ for brevity.  Rather than relying on a single noisy comparison, RHI conditions each harness update on the accumulated preference history, thereby improving robustness while requiring only one new agent execution and one pairwise evaluation per iteration. Since the harness space is discrete and textual, these pairwise preferences do not define conventional gradients. Instead, the history $\mathcal{D}_x^{(i)}$ serves as a momentum-semantic signal that guides future harness revisions. 


Those insights motivate RHI's learning algorithm introduced in the next section.


\subsection{RHI algorithm}
\label{subsec:objective_function}

{\RestyleAlgo{ruled}
\begin{algorithm}[h]
\caption{Recursive Harness Self-Improvement}
\label{alg:rhsi}

\KwIn{
task $\{x_j\}_{j=1}^n \in \cX$,
agent $\mathcal{A}$,
evaluator $\mathcal{L}_{\mathrm{eval}}$,
harness optimizer $\mathcal{L}_{\mathrm{harness}}$,
evaluation prompt $x_{\text{eval}}$,
stopping threshold $\epsilon$
}

\textbf{Initialize:} harness $H^{(0)}_{x} \in \mathcal{V}^*$, harness history $\mathcal{D}_x \leftarrow \emptyset$ for $\forall x \in \cX$.

Agent $\mathcal{A}$ solves task $x$ using initial harness $H^{(0)}_x$ and obtains output for $\forall x \in  \cX$.

\For{$i=1,2,\ldots$}{
    \For{$j =1,2,\ldots,n$}{
        Agent $\mathcal{A}$ solves  task $x_j$ using harness $H^{(i)}_{j}$ and obtains outputs $y^{i}_{j}$.

        Evaluator $\mathcal{L}_{\mathrm{eval}}$ compares $y^{i}_{j}$ and $y^{i-1}_{j}$ via pairwise judgments then update this history $\mathcal{D}_j \leftarrow \mathcal{D}_j \cup \{\cL_{\text{eval}}(y^{i}_{j},y^{i-1}_{j};x_{\text{eval}})\}$.
    }
    Compute the improvement rate for all tasks $s_i = \frac{1}{n}\sum_{j=1}^{n}\mathbf{1}\!\left[y_j^i \succ y_j^{i-1}\right]$ where $s_i \in [0,1]$.

    \If{$s_i < \epsilon$}{
        BREAK
    }

    Update Harness $H^{(i+1)}_{x} \leftarrow \mathcal{L}_{\text{harness}}(H^{(i)}_x, \mathcal{D}_x)$ without $x_{\text{eval}}$  for $\forall x \in \cX$.
}

\Return $H^*_{x}$ for $\forall x \in \cX$
\end{algorithm}
}

Building on the preceding insights, we present the RHI algorithm in Algorithm~\ref{alg:rhsi}. At iteration $i$, RHI updates the task-specific harness according to \begin{equation} 
    H_x^{(i+1)} = \mathcal{L}_{\mathrm{harness}} \left( H_x^{(i)}, \mathcal{D}_x^{(i)} \right),
    \label{eq2:hanressupdaterule} 
\end{equation} 
where $\mathcal{L}_{\mathrm{harness}}$ is an LLM-based harness optimizer. Here, the direct update from $H_x^{(i)}$ to $H_x^{(i+1)}$ makes RHI computationally lightweight, while conditioning on $\mathcal{D}_x^{(i)}$ enables RHI to converge within only a few update iterations. Importantly, $x_{\mathrm{eval}}$ is used only by the evaluator when producing pairwise feedback; it is not directly provided to $\mathcal{L}_{\mathrm{harness}}$ during harness revision. 

RHI should therefore be viewed as optimizing an \emph{implicit preference objective} rather than directly maximizing an explicit scalar reward. Although the harness optimizer does not observe $x_{\mathrm{eval}}$ directly, the feedback history $\mathcal{D}_x^{(i)}$ is generated by an evaluator conditioned on $x_{\mathrm{eval}}$. Consequently, the update trajectory can indirectly align harness revisions with the $x_{\mathrm{eval}}$ through accumulated pairwise comparisons. In this sense, RHI performs \emph{recursive self-improvement}: each harness is optimized using the preference history induced by its own previous revisions without directly observing the evaluation prompt $x_{\mathrm{eval}}$.

Appendix~\ref{appendix:Harness improvement examples} provides examples of the iterative evolution of $H^{(i)}_x$ for $i \in \{0,1,2,\ldots\}$ on a randomly selected task $x$. Appendix~\ref{sec:Harness improvement prompt} provides the system and user prompts used by the harness optimizer $\mathcal{L}_{\mathrm{harness}}$.

\subsection{RHI harness definition}
\label{sec:Harness}

\begin{figure}[h]
    \centering
    \includegraphics[width=0.9\linewidth]{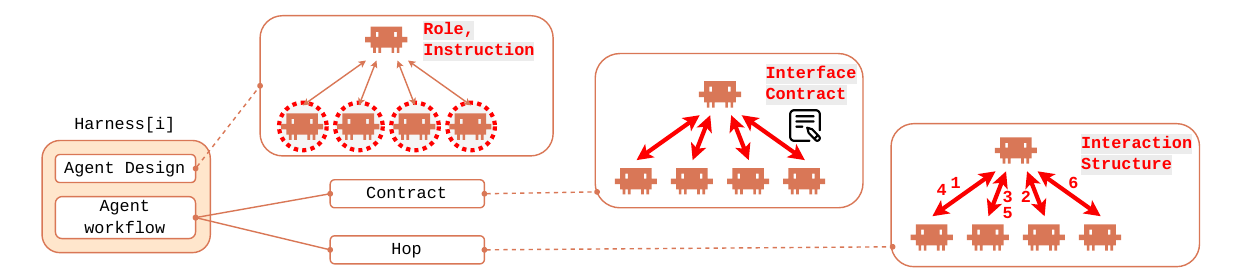}
    \caption{Decomposition of the RHI harness. We classify the harness into two primary components: agent design and agent workflow. Agent design specifies the roles and instructions of candidate agents. Agent workflow is further decomposed into contract and hop: contract defines the information exchanged between subagents and the orchestrator, while hop defines the interaction structure; overall workflow logics.}
    \label{fig:harness_decomposition}
\end{figure}

\begin{figure}[h]
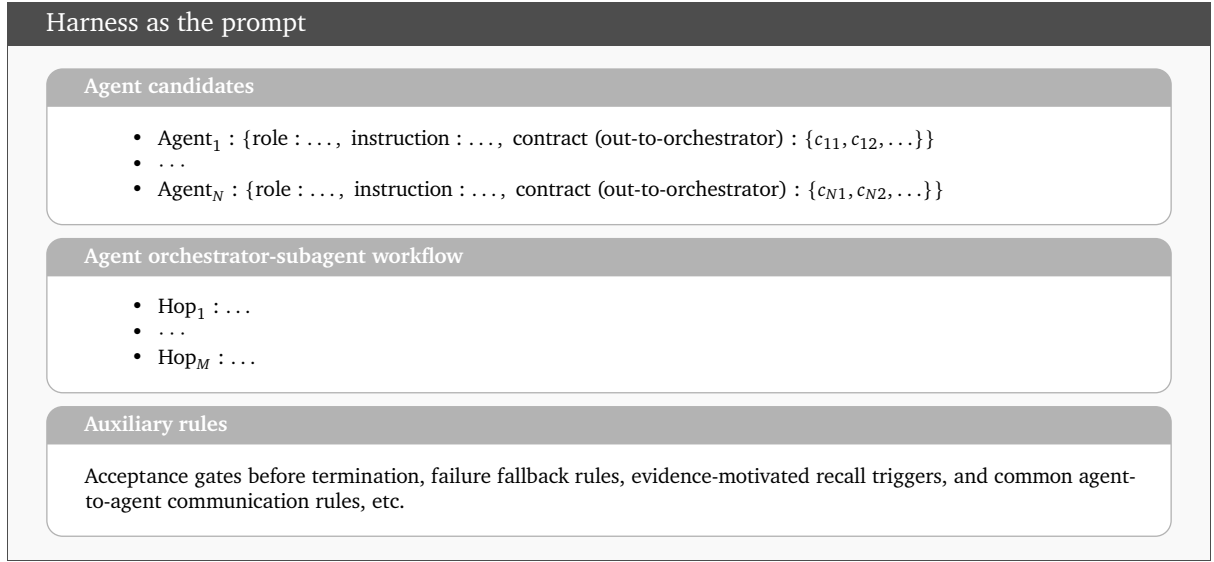

\centering
\scriptsize

\begin{tcolorbox}[
    colback=gray!5,
    colframe=black!70,
    title=Harness as the prompt,
    fonttitle=\small,
    sharp corners
]

\begin{tcolorbox}[
    colback=white,
    colframe=black!30,
    boxrule=0.5pt,
    arc=2mm,
    title=\textbf{Agent candidates},
    fonttitle=\scriptsize\bfseries
]

\begin{itemize}[nosep]
    \item $\mathrm{Agent}_1:
    \left\{
    \text{role}: \ldots,\;
    \text{instruction}: \ldots,\;
    \text{contract (out-to-orchestrator)}: \{c_{11},c_{12},\ldots\}
    \right\}$

    \item $\cdots$

    \item $\mathrm{Agent}_N:
    \left\{
    \text{role}: \ldots,\;
    \text{instruction}: \ldots,\;
    \text{contract (out-to-orchestrator)}: \{c_{N1},c_{N2},\ldots\}
    \right\}$
\end{itemize}
\end{tcolorbox}

\begin{tcolorbox}[
    colback=white,
    colframe=black!30,
    boxrule=0.5pt,
    arc=2mm,
    title=\textbf{Agent orchestrator-subagent workflow},
    fonttitle=\scriptsize\bfseries
]
\begin{itemize}[nosep]
    \item $\mathrm{Hop}_1:$ $\ldots$
    \item $\cdots$
    \item $\mathrm{Hop}_M:$ $\ldots$
\end{itemize}
\end{tcolorbox}

\begin{tcolorbox}[
    colback=white,
    colframe=black!30,
    boxrule=0.5pt,
    arc=2mm,
    title=\textbf{Auxiliary rules},
    fonttitle=\scriptsize\bfseries
]
Acceptance gates before termination, failure fallback rules, evidence-motivated recall triggers, and common agent-to-agent communication rules, etc.
\end{tcolorbox}

\end{tcolorbox}

\caption{
Structure of a prompt-represented harness $H^{(i)}$. The harness specifies candidate agents, the orchestrator--subagent workflow, and auxiliary control rules used for termination, fallback, recall, and inter-agent communication.
}
\label{fig:harness_structure}
\end{figure}

We define the harness as the agent loop itself. We decompose the harness into two primary components: agent design and agent workflow. RHI prioritizes the agent workflow for update, which is iteratively adapted to each task (see system prompt of $\cL_{\mathrm{harness}}$ in Appendix \ref{sec:Harness improvement prompt}). We further decompose the workflow into two components: \emph{contracts}, which specify what information is passed between agents through the communication interface, and \emph{hops}, which specify the interaction structure, including orchestrator-subagent workflow steps. Figure~\ref{fig:harness_decomposition} presents this high-level decomposition, while Figure~\ref{fig:harness_structure} illustrates how these components are represented as textual specifications in the prompt.

\paragraph{Why does RHI prioritize workflow updates over agent design?}

Our focus on contracts and hops is motivated by the following hypothesis.

We hypothesize that \emph{task-specific contracts and hops can improve both task performance and inference efficiency through more effective context management}. In particular, a task-specific contract specifies which information should be transmitted between agents, rather than requiring the orchestrator or downstream agents to condition on the entire interaction history. This can reduce redundant context propagation, improve KV-cache efficiency, and lower inference cost. This perspective naturally motivates workflow designs such as multi-agent execution or multi-persona reasoning within a single agent.

Conceptually, contract optimization is analogous to imposing a task-dependent sparsity pattern on inter-agent information flow. A fully shared interaction history resembles dense attention, where each agent has access to all prior information from other agents. In contrast, a task-specific contract restricts communication to the information most relevant for downstream decision-making. Thus, optimizing contracts can be viewed as learning a task-specific communication sparsity structure at the harness level, enabling more efficient context management.



\paragraph{Harness as a prompt} 

The RHI harness first specifies a set of candidate agents, where each agent is described by its role, instructions, and contract, i.e., its expected output to the orchestrator; see ``Agent candidates'' in Figure~\ref{fig:harness_structure}. It then specifies the workflow governing interactions among these agents; see ``Agent orchestrator-subagent workflow'' in Figure~\ref{fig:harness_structure}. These two components are included in every RHI iteration and serve as a task-agnostic scaffold for harness optimization. Across iterations, RHI further induces task-dependent local harness components, such as acceptance gates, failure fallback rules, and evidence-motivated recall triggers; see the ``auxiliary rules'' in Figure~\ref{fig:harness_structure}.

\section{Benchmark $\&$ Evaluation}
\label{sec:ExperimentSetting}

\hyunin{Here, I explain our bechmark and our evaluation protocol}

\subsection{Synthetic open-ended machine learning research tasks}

We evaluate RHI on a suite of $30$ synthetic, open-ended machine learning (ML) research tasks spanning three domains: quantitative finance, robotics, and pharmaceutical machine learning, with $10$ tasks per domain. To construct these tasks, we use a LLM to transform domain-relevant industry job postings into research-style task prompts that resemble assignments a researcher might receive in the corresponding role. The source postings are:

\begin{enumerate}
\item quantitative ML research: \url{https://www.citadel.com/careers/details/quantitative-researcher-phd-graduate-us/};
\item robotics ML research: \url{https://www.amazon.jobs/en/jobs/10443615/member-of-technical-staff-far-frontier-ai-robotics}
\item pharmaceutical ML research: \url{https://careers.gene.com/us/en/job/202606-115542/Machine-Learning-Scientist-Synthesis-Planning-and-Optimization}.
\end{enumerate}

The generated tasks require a combination of coding, domain reasoning, empirical analysis, and ML experimentation, etc. They include objectives such as data analysis, model training, benchmark construction, ablation design, and empirical reporting. Each task asks the agent to construct a complete code repository and produce the deliverables specified in the prompt.

This benchmark is designed to support diverse ML tasks while preserving a consistent evaluation interface. Each task prompt consists of two parts: a \emph{task description}, which specifies the research objective, and a \emph{deliverables} section, which specifies the required output artifacts. Across tasks, the core deliverables are standardized: (1) \texttt{research\_report.md}, (2) visualization files in \texttt{.png} format, (3) \texttt{metrics.json}, and (4) \texttt{index.json}, which records the paths of all generated files. Some tasks additionally require domain-specific artifacts, such as model-comparison files, ablation summaries, or reproducible scripts.

Appendix~\ref{appendix:example-task-queries} provides an example task prompt from each domain: robotics, quantitative finance, and pharmacy.


\subsection{LLM-as-a-judge evaluation}

Our evaluation protocol follows the problem formulation in Section~\ref{sec:Problem definition}, where outputs are compared through pairwise preference under multiple evaluation criteria. This corresponds to Step~3 of the RHI algorithm (Figure~\ref{fig:rhsi}). Given two repositories generated for the same task, the standardized deliverable structure provides a common evaluation interface, allowing an LLM evaluator to compare the task-specific deliverables and select the preferred repository.

For each comparison, we extract the task-specified deliverables from both repositories and provide their contents, together with the evaluation prompt $x_{\mathrm{eval}}$, to $\mathcal{L}_{\mathrm{eval}}$. Since repositories may contain many artifacts (e.g., source code, logs, text files, and figures), including every generated file is often impractical. We therefore evaluate only the task-specified deliverables and keep the evaluator prompt within a practical context budget (approximately $30$--$40\%$ of the evaluator's maximum input length) to mitigate context rot \citep{hong2025context}. When necessary, long files are truncated using the same protocol for both candidate repositories.

To evaluate robustness across evaluators, we use two configurations with different model families and reasoning settings: \texttt{gpt-5.5} with maximum reasoning effort and \texttt{opus-4.7} (or \texttt{opus-4.8}) with xhigh reasoning effort. Each configuration is evaluated using three random seeds.

The evaluator $\mathcal{L}_{\mathrm{eval}}$ compares the two repositories according to the following criteria.

\begin{tcolorbox}[
    colback=gray!5,
    colframe=black!70,
    title=Evaluation prompt ($x_{\mathrm{eval}}$),
    fonttitle=\small,
    sharp corners
]
\scriptsize

\begin{enumerate}
\item \textbf{Deliverable coverage}: whether the output satisfies the task's explicit deliverable requirements, including files listed under \texttt{Deliverables}, \texttt{Deliverable}, or inline output requirements.
\item \textbf{Numerical and empirical rigor}: whether the methodology, baselines, metrics, and limitations are appropriate and internally consistent.
\item \textbf{Reproducibility}: whether the repository includes dependencies, entry points, seeds, and sufficient documentation for reproducing the results.
\item \textbf{Presentation}: whether the report is clear, well structured, and appropriately integrates figures and empirical results.
\item \textbf{Engineering quality}: whether the repository is organized, modular, and readable based on the file tree and code excerpts.
\item \textbf{Task alignment}: whether the solution addresses the stated objective without drifting to a different problem.
\end{enumerate}

\end{tcolorbox}

Appendix~\ref{appendix:Evaluation Protocol} provides the complete system and user prompts used by $\mathcal{L}_{\mathrm{eval}}$.

\section{Experiment}
\label{sec:Experiment}

Section~\ref{subsec:Experiment settings} describes the experimental settings, Section~\ref{subsec:Evaluation metrics} defines the evaluation metrics, and Sections~\ref{subsec:Claim1}--\ref{subsec:Claim3} present our main experimental results.

\subsection{Experiment settings}
\label{subsec:Experiment settings}
RHI is designed to be computationally lightweight and to converge within only a few update iterations. Our experimental setup is designed to test whether RHI can raise the performance ceiling of agent test-time scaling.

To this end, we instantiate the coding agent $\mathcal{A}$ with three base language models: Claude Sonnet 4.6, Claude Opus 4.7, and Claude Opus 4.8. For each model, the default agent uses reasoning effort \texttt{high}. We denote these base configurations by
$\mathcal{L}_1\text{-}\texttt{high}$,
$\mathcal{L}_2\text{-}\texttt{high}$, and
$\mathcal{L}_3\text{-}\texttt{high}$, respectively. We denote the corresponding base agent by
$\mathcal{A}_{\mathcal{L}_b\text{-}\texttt{high}}$.
For each task $x \in \mathcal{X}$ and each base model $b \in \{1,2,3\}$, we first run the base agent with the initial harness, $\mathcal{A}_{\mathcal{L}_b\text{-}\texttt{high}}(H_x^{(0)}, x)$, and then apply RHI to iteratively produce a sequence of task-specific harnesses
$\{H_x^{(0)}, H_x^{(1)}, \ldots\}$.
At each iteration, we compare the output of the RHI-improved agent against that of the corresponding higher test-time reasoning agent without RHI:
\begin{equation*}
y_{\mathrm{RHI}}
\sim
\mathcal{A}_{\mathcal{L}_b\text{-}\texttt{high}}(H_x^{(i)},x)
\quad
\text{vs.}
\quad
y_{\mathrm{scale}}
\sim
\mathcal{A}_{\mathcal{L}_b\text{-}\texttt{xhigh/ultracode/max}}(x).
\end{equation*}

Here,
$\mathcal{A}_{\mathcal{L}_b\text{-}\texttt{high}}(H_x^{(i)},x)$
denotes the base coding agent equipped with the RHI-improved harness $H_x^{(i)}$ at iteration $i$. Relative to the original base agent $\mathcal{A}_{\mathcal{L}_b\text{-}\texttt{high}}$, the only modification is the input prompt, which changes from $x$ to $x \cup H_x^{(i)}$ by augmenting the original task with the evolved harness represented as a textual specification. In contrast,
$\mathcal{A}_{\mathcal{L}_b\text{-}\texttt{xhigh/ultracode/max}}(x)$
denotes the same coding agent evaluated with a higher reasoning effort, but without an RHI-generated harness.

\begin{figure}[h]
    \centering
    \begin{subfigure}{0.45\linewidth}
        \includegraphics[width=\linewidth]{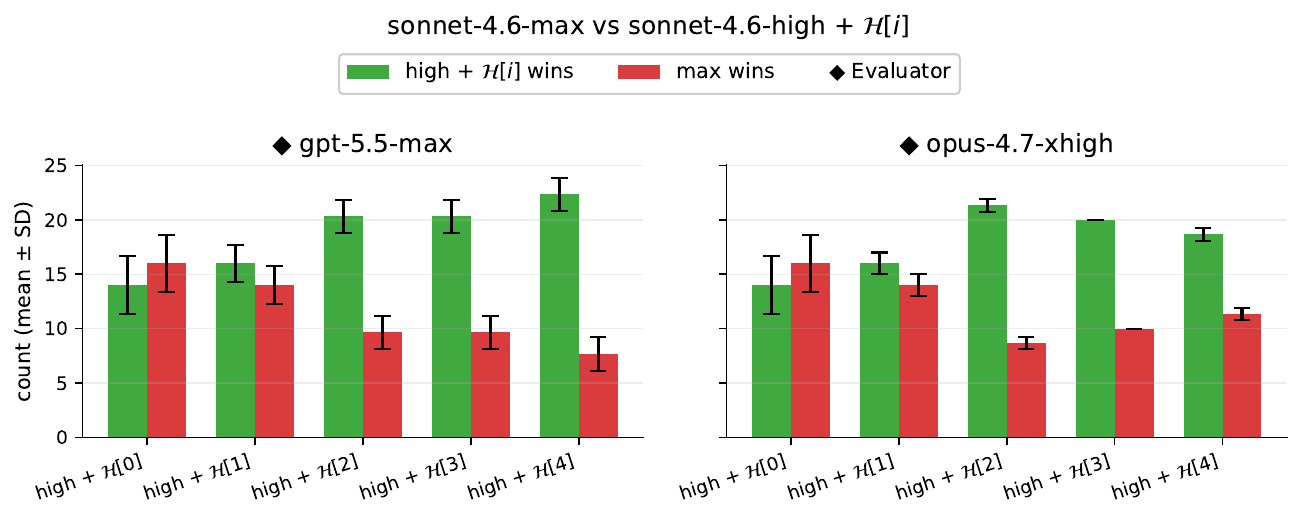}
        \caption{
        $\texttt{sonnet-4.6-high}+\mathcal{H}[i]$, $i \in \{0,1,2,3,4\}$.
        }
        \label{fig:sonnet46_performance}
    \end{subfigure}
    \begin{subfigure}{0.54\linewidth}
        \includegraphics[width=\linewidth]{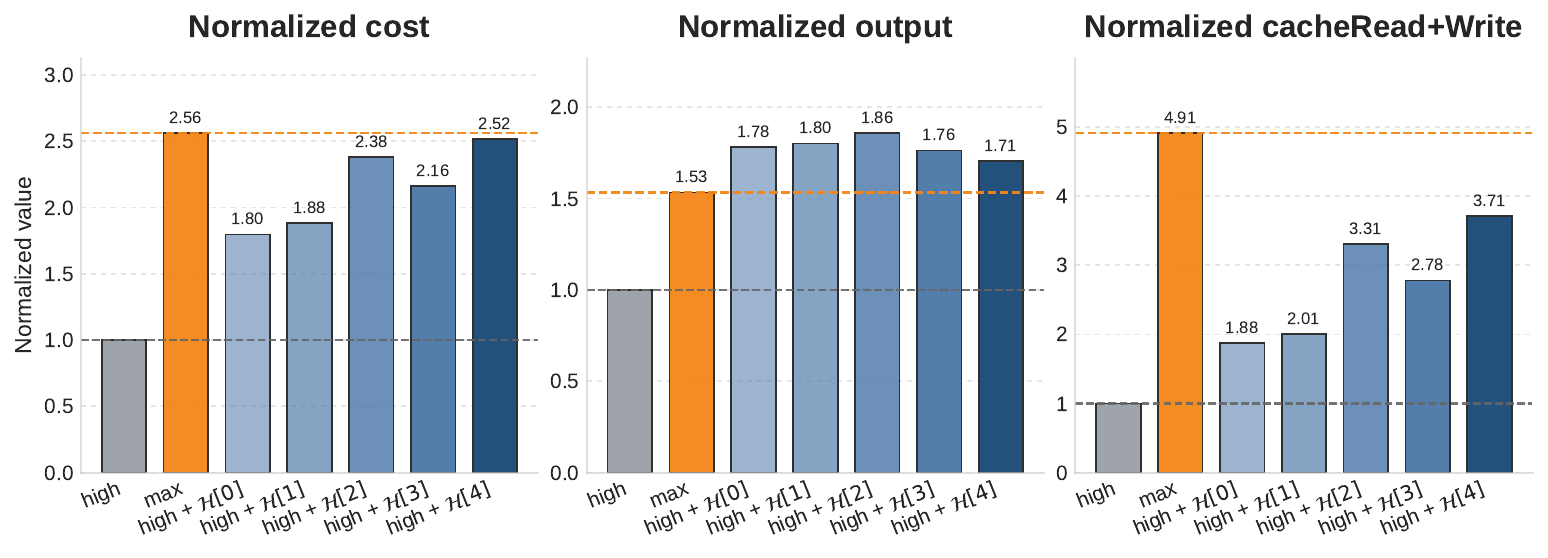}
        \caption{
        Normalized resource usage.
        }
        \label{fig:sonnet46_cost}
    \end{subfigure}

    \caption{
    \textbf{After two iterations, RHI raises the empirical ceiling of \texttt{sonnet-4.6} test-time scaling.}
    We compare \texttt{sonnet-4.6-high}$+\mathcal{H}[i]$ against \texttt{sonnet-4.6-max}, where $\mathcal{H}[i]$ denotes the harness after $i$ RHI iterations. The left panel reports pairwise outcomes over $30$ tasks; the right panel reports mean cost, output-token count, and cache read/write usage, normalized by \texttt{sonnet-4.6-high}. Results are averaged over two LLM judges and three seeds; error bars denote standard deviation.
    }
    \label{fig:sonnet46_rhsi_tts}
\end{figure}

\begin{figure}[h]
    \centering

    \begin{subfigure}[b]{0.95\linewidth}
        \centering
        \includegraphics[width=\linewidth,height=0.26\textheight,keepaspectratio]{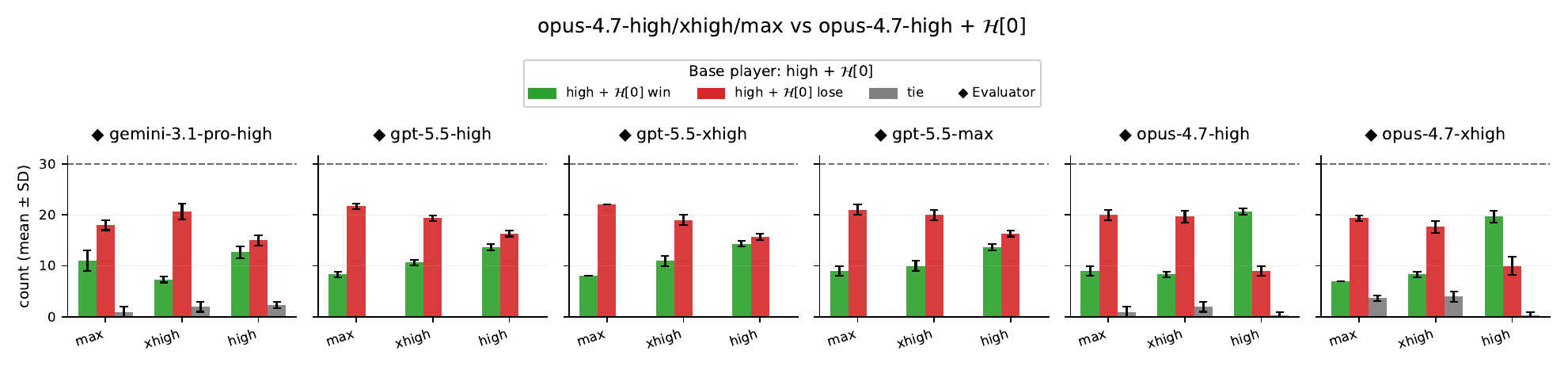}
        \caption{
        $\texttt{opus-4.7-high}+\mathcal{H}[0]$.
        }
        \label{fig:opus47high_performance_rhsi0}
    \end{subfigure}

    \vspace{0.5em}

    \begin{subfigure}[b]{0.95\linewidth}
        \centering
        \includegraphics[width=\linewidth,height=0.26\textheight,keepaspectratio]{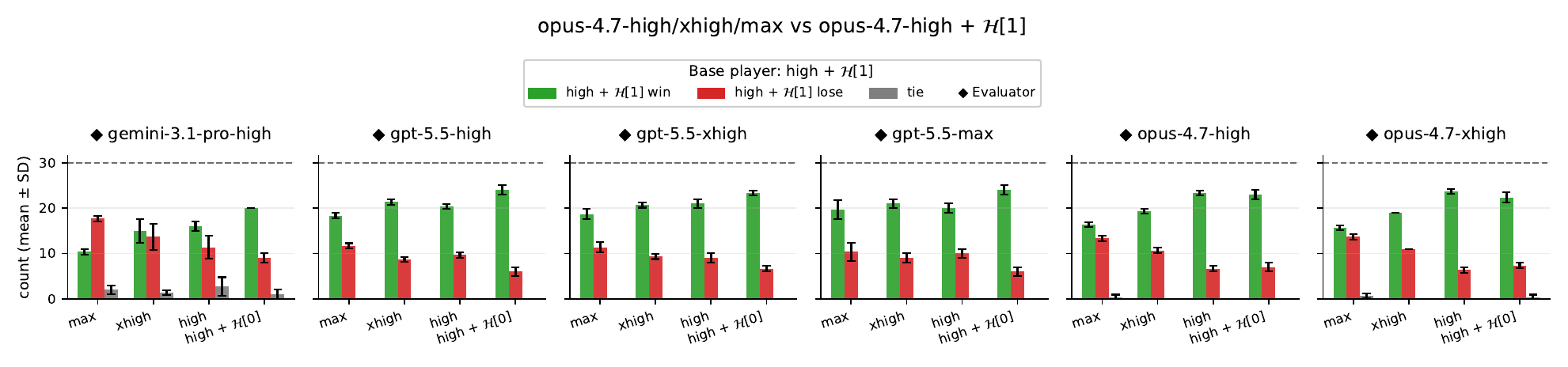}
        \caption{
        $\texttt{opus-4.7-high}+\mathcal{H}[1]$.
        }
        \label{fig:opus47high_performance_rhsi1}
    \end{subfigure}

    \vspace{0.5em}

    \begin{subfigure}[b]{0.6\linewidth}
        \centering
        \includegraphics[width=\linewidth,height=0.26\textheight,keepaspectratio]{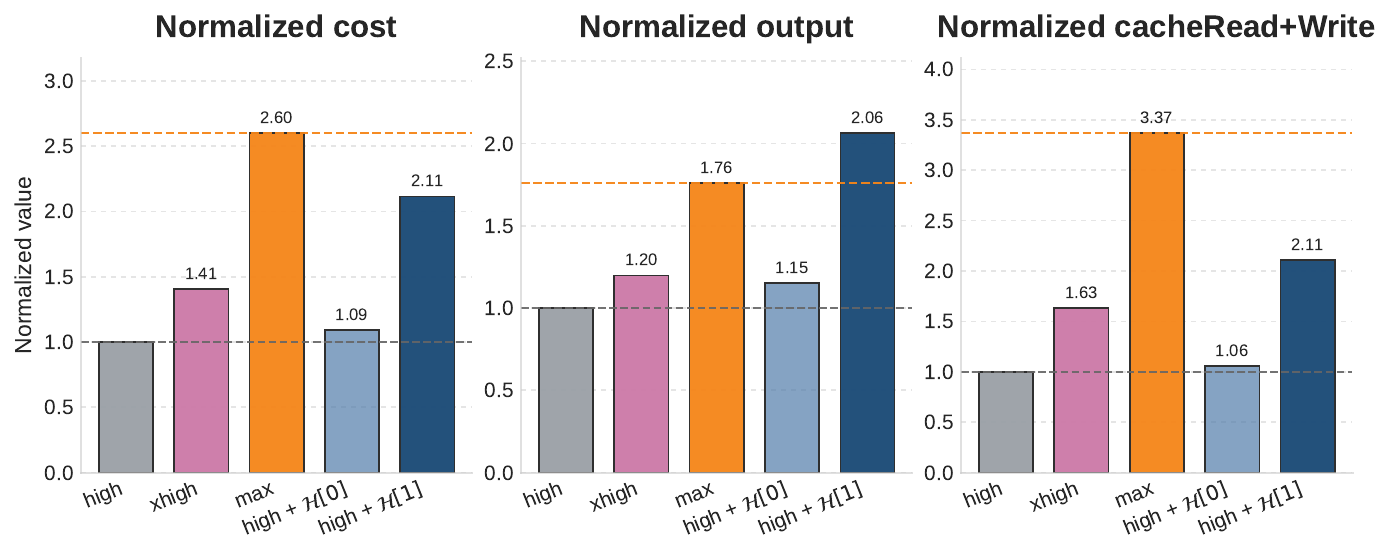}
        \caption{
        Normalized resource usage.
        }
        \label{fig:opus47high_cost}
    \end{subfigure}

    \caption{
    \textbf{After one iteration, RHI raises the empirical ceiling of \texttt{opus-4.7} test-time scaling.}
    The pairwise panels compare \texttt{opus-4.7-high}$+\mathcal{H}[0]$ and \texttt{opus-4.7-high}$+\mathcal{H}[1]$ against \texttt{opus-4.7-high/xhigh/max} over $30$ tasks. The resource panel reports mean cost, output-token count, and cache read/write usage, normalized by \texttt{opus-4.7-high}. Results are averaged over six evaluator configurations and three seeds; error bars denote standard deviation.
    }
    \label{fig:opus47_rhsi_tts}
\end{figure}

\begin{figure}[h]
    \centering
    \begin{subfigure}{0.45\linewidth}
        \includegraphics[width=\linewidth]{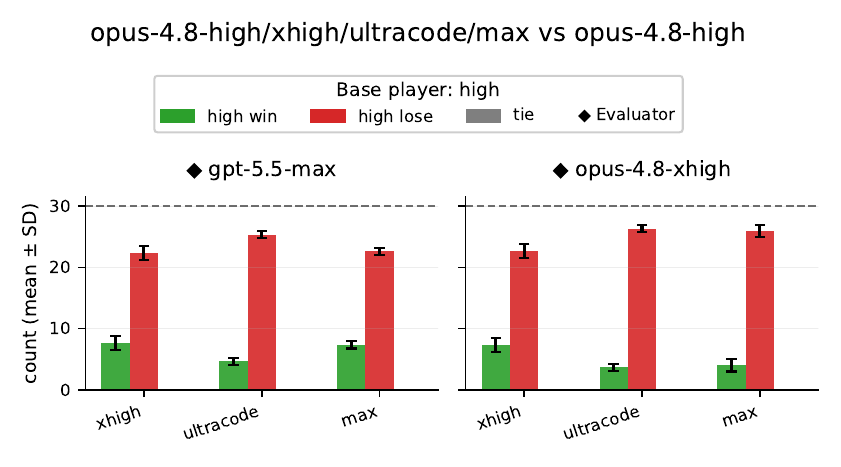}
        \caption{
        \texttt{opus-4.8-high}
        }
        \label{fig:opus48_rhsi_tts_base}
    \end{subfigure}
    \begin{subfigure}{0.45\linewidth}
        \includegraphics[width=\linewidth]{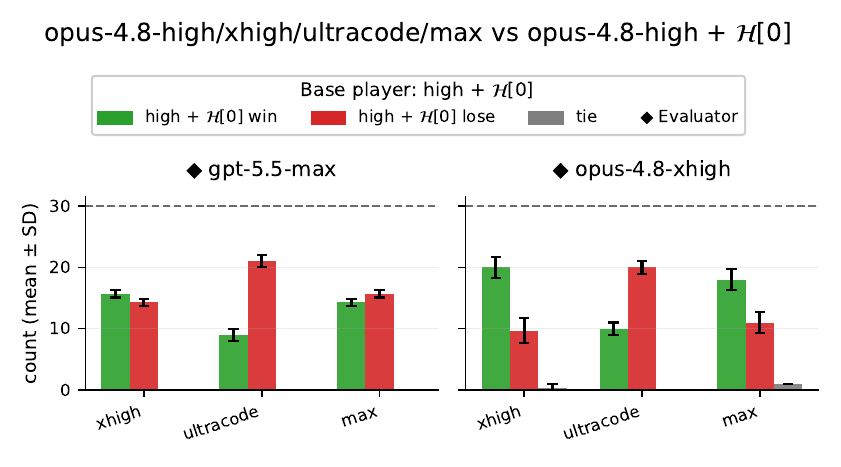}
        \caption{
        $\texttt{opus-4.8-high}+\mathcal{H}[0]$
        }
        \label{fig:opus48_rhsi_tts_v0}
    \end{subfigure}

    \begin{subfigure}{0.45\linewidth}
        \includegraphics[width=\linewidth]{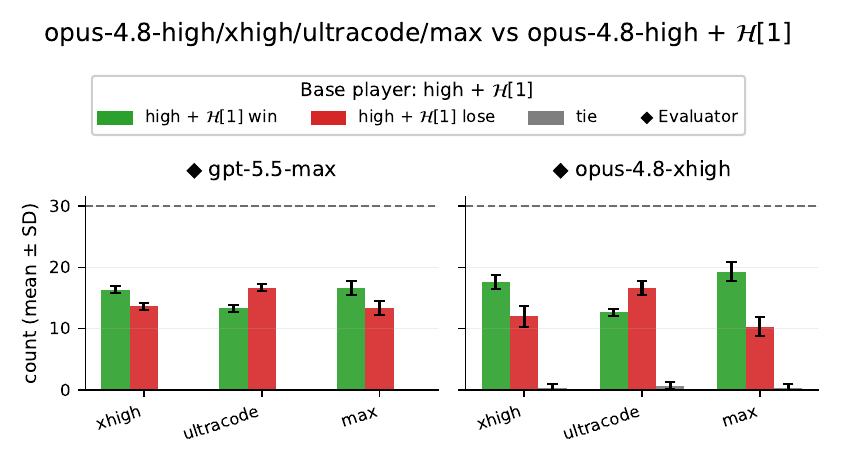}
        \caption{
        $\texttt{opus-4.8-high}+\mathcal{H}[1]$
        }
        \label{fig:opus48_rhsi_tts_v1}
    \end{subfigure}
    \begin{subfigure}{0.45\linewidth}
        \includegraphics[width=\linewidth]{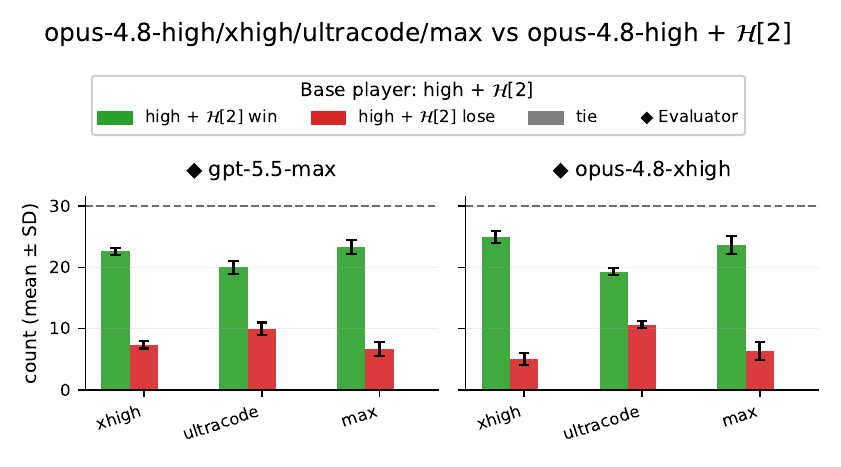}
        \caption{
        $\texttt{opus-4.8-high}+\mathcal{H}[2]$
        }
        \label{fig:opus48_rhsi_tts_v2}
    \end{subfigure}

    \begin{subfigure}{0.6\linewidth}
        \includegraphics[width=\linewidth]{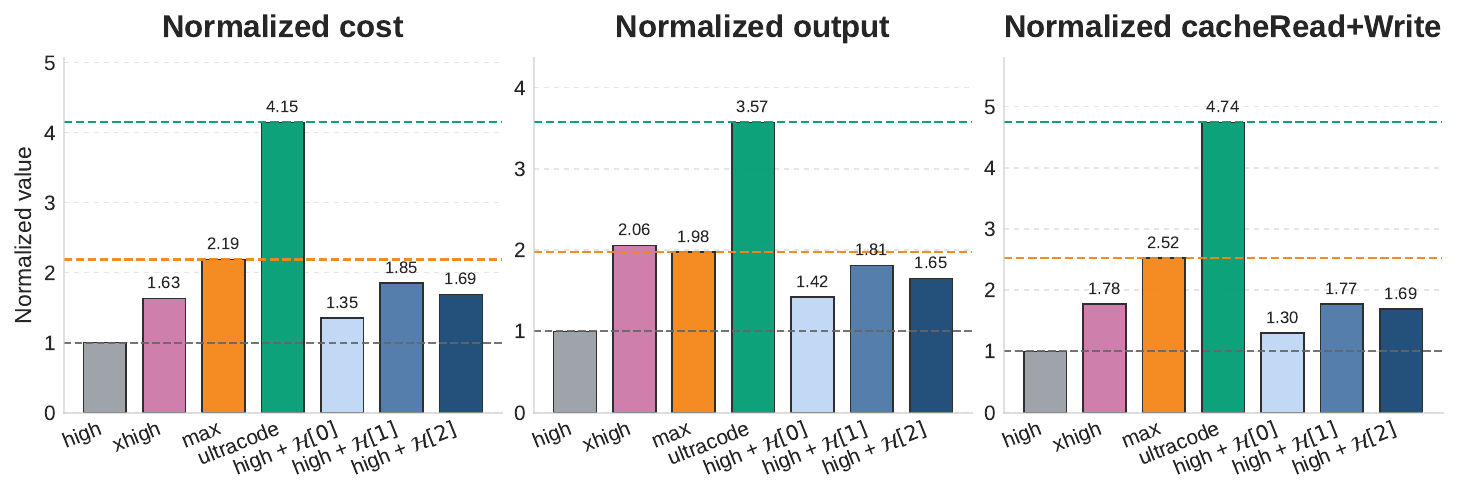}
        \caption{
        Normalized resource usage.
        }
        \label{fig:opus48high_cost}
    \end{subfigure}

    \caption{
    \textbf{After two iterations, RHI raises the empirical ceiling of \texttt{opus-4.8} test-time scaling.}
    The pairwise panels compare \texttt{opus-4.8-high} and \texttt{opus-4.8-high}$+\mathcal{H}[i]$, $i \in \{0,1,2\}$, against \texttt{opus-4.8-xhigh/ultracode/max} over $30$ tasks. The resource panel reports mean cost, output-token count, and cache read/write usage, normalized by \texttt{opus-4.8-high}. Results are averaged over two LLM judges and three seeds; error bars denote standard deviation.}
    \label{fig:opus48_rhsi_tts}
\end{figure}


\subsection{Evaluation metrics}
\label{subsec:Evaluation metrics}
\paragraph{Notation.} For compactness, we use the bracketed form \(\mathcal{H}[i]\)
interchangeably with \(H^{(i)}\) for the same iteration index, particularly in figure
and table labels. Thus, \(\mathcal{H}[i]\) corresponds to \(H^{(i)}\), or to
\(H_{x}^{(i)}\) for a particular task \(x\).

Figure~\ref{fig:sonnet46_rhsi_tts} compares RHI on \texttt{sonnet-4.6-high} against the test-time scaling baseline \texttt{sonnet-4.6-max}; Figure~\ref{fig:opus47_rhsi_tts} compares RHI on \texttt{opus-4.7-high} against \texttt{opus-4.7-xhigh/max}; and Figure~\ref{fig:opus48_rhsi_tts} compares RHI on \texttt{opus-4.8-high} against \texttt{opus-4.8-xhigh/ultracode/max}. For each base model, we report four metrics: pairwise win/loss count (i.e., performance), normalized cost, output token count, and cache read/write usage. Cost, output token count, and cache read/write usage are recorded using the agent's internal \texttt{cost} function of the claude coding agent.

\paragraph{Performance.}
Figures~\ref{fig:sonnet46_performance}, \ref{fig:opus47high_performance_rhsi0},
\ref{fig:opus47high_performance_rhsi1}, and \ref{fig:opus48_rhsi_tts_base}--\ref{fig:opus48_rhsi_tts_v2}
report performance over 30 tasks, with the $y$-axis giving the number of wins, losses, and
ties. Figure~\ref{fig:sonnet46_performance} compares \texttt{sonnet-4.6-max} against
\texttt{sonnet-4.6-high}$+\mathcal{H}[i]$ for $i \in \{0,1,2,3,4\}$. Figures~\ref{fig:opus47high_performance_rhsi0}
and~\ref{fig:opus47high_performance_rhsi1} compare
\texttt{opus-4.7-high}$+\mathcal{H}[0]$ and \texttt{opus-4.7-high}$+\mathcal{H}[1]$ against
\texttt{opus-4.7-high/xhigh/max}. Figures~\ref{fig:opus48_rhsi_tts_base}--\ref{fig:opus48_rhsi_tts_v2}
compare \texttt{opus-4.8-high} and \texttt{opus-4.8-high}$+\mathcal{H}[i]$, $i \in \{0,1,2\}$,
against \texttt{opus-4.8-xhigh/ultracode/max}. Each evaluation is repeated three times with
different seeds; we report the mean and standard deviation across repetitions. To assess
robustness, the comparisons use two evaluator configurations for \texttt{sonnet-4.6} and
\texttt{opus-4.8} and six for \texttt{opus-4.7}, varying both the evaluator model and its
reasoning-effort setting.

\paragraph{Normalized cost, output token count, and cache read/write usage.}
Figures~\ref{fig:sonnet46_cost}, \ref{fig:opus47high_cost}, and \ref{fig:opus48high_cost} report
mean normalized cost, output-token count, and cache read/write usage. For each task we record these
three quantities, normalize each by the corresponding base-agent value, and average across tasks.
Normalization is with respect to \texttt{sonnet-4.6-high}, \texttt{opus-4.7-high}, and
\texttt{opus-4.8-high}, respectively; in each figure the base agent appears as the leftmost gray
bar with a normalized value of $1.0$. See Appendix~\ref{appendix:Cost distribution} for the cost
distribution.

We discuss the main findings in the next three subsections.

\subsection{Claim 1. Few-shot RHI lifts the performance ceiling of test-time scaling}
\label{subsec:Claim1}

For the \texttt{Sonnet-4.6} baseline, Figure~\ref{fig:sonnet46_performance} shows that, after two RHI iterations, \texttt{sonnet-4.6-high}$+\mathcal{H}[2]$ outperforms the stronger test-time scaling baseline \texttt{sonnet-4.6-max}, winning $20$ of $30$ pairwise comparisons. Moreover, \texttt{sonnet-4.6-high}$+\mathcal{H}[i]$, for $i \in \{3,4\}$, continues to outperform the baseline.

The same trend holds for the two stronger base models. For the \texttt{Opus-4.7} baseline, Figure~\ref{fig:opus47high_performance_rhsi0} shows that \texttt{opus-4.7-high}$+\mathcal{H}[0]$ underperforms both the \texttt{opus-4.7-xhigh} and \texttt{opus-4.7-max} baselines. However, Figure~\ref{fig:opus47high_performance_rhsi1} shows that a single RHI iteration is sufficient to surpass both. For the \texttt{Opus-4.8} baseline, Figure~\ref{fig:opus48_rhsi_tts_v0} shows that \texttt{opus-4.8-high}$+\mathcal{H}[0]$ slightly outperforms or ties with \texttt{opus-4.8-xhigh}. After two RHI iterations, Figure~\ref{fig:opus48_rhsi_tts_v2} shows that \texttt{opus-4.8-high}$+\mathcal{H}[2]$ outperforms all test-time scaling baselines: \texttt{opus-4.8-xhigh}, \texttt{opus-4.8-ultracode}, and \texttt{opus-4.8-max}.

These results indicate that a few RHI iterations can raise performance beyond the ceiling at which same-family test-time scaling saturates under our benchmark and evaluation protocol. Importantly, this improvement is not confined to a single base-model capability, but persists across progressively stronger models.


\subsubsection*{Claim 1.1. Optimizing user-constructed harnesses at the prompt level can outperform a system-level provider-built harness}
\label{subsubsec:Claim 1.1.}

A particularly notable result is that \texttt{opus-4.8-high}$+\mathcal{H}[2]$ outperforms \texttt{opus-4.8-ultracode} (Figure~\ref{fig:opus48_rhsi_tts_v2}). \texttt{opus-4.8-ultracode} uses \texttt{xhigh} reasoning together with a built-in dynamic workflow that can spawn multiple subagents \citep{anthropic2026claudecodemodelconfig}. We interpret this result as evidence that, on our benchmark, dynamically generating multi-agent workflows through a \emph{provider-built-in harness} is not always sufficient. Instead, a \emph{task-specific, user-constructed} harness supplied directly in the prompt can outperform it.

This finding suggests a limitation of relying on a single fixed, \emph{system-level} harness to cover a broad and heterogeneous task distribution. By optimizing the harness at the \emph{prompt level}, RHI enables the multi-agent workflow to adapt to each task. The advantage over \texttt{opus-4.8-ultracode} therefore not only demonstrates the effectiveness of optimizing a \emph{user-constructed harness} over a \emph{provider-built-in harness}, but also supports the \emph{harness-as-prompt} formulation, in which harness design becomes an explicit, task-specific optimization variable rather than a fixed backend workflow.

\subsection{Claim 2. RHI performance gains are not explained by increased output-token usage}
\label{subsec:Claim2}

Here, we show that the performance gains from RHI are distinct from those of test-time scaling, which primarily relies on increased output-token usage, demonstrating that RHI is a standalone method.

For the \texttt{Sonnet-4.6} baseline, Figure~\ref{fig:sonnet46_cost} shows that the normalized output-token usage remains nearly constant across RHI iterations $\mathcal{H}[i]$, $i \in \{0,1,2,3,4\}$, ranging from $1.71$ to $1.86$. In contrast, Figures~\ref{fig:opus47high_performance_rhsi0} and \ref{fig:opus47high_performance_rhsi1} shows that performance consistently improves across the same iterations.

The same decoupling is observed for the \texttt{Opus-4.8} baseline. Figure~\ref{fig:opus48high_cost} shows that the normalized output-token usage remains nearly flat, ranging from $1.42$ to $1.81$, whereas Figure~\ref{fig:opus48_rhsi_tts} shows that performance improves across $\mathcal{H}[i]$, $i \in \{0,1,2\}$.

For the \texttt{Opus-4.7} baseline, the evidence is inconclusive. Figure~\ref{fig:opus47high_cost} shows that output-token usage increases together with performance across $\mathcal{H}[i]$, $i \in \{0,1\}$, so the two explanations cannot be disentangled. Moreover, this comparison spans only two iterations, fewer than for the other models, which further limits the conclusions that can be drawn.

Overall, the gains are not \emph{primarily} driven by longer generations: for two of the three models, performance improves while output-token usage remains nearly constant.

Consistent with this observation, in the next subsection we show that RHI's gains are more closely associated with KV-cache usage than with output-token usage, suggesting that \emph{efficient context management}, rather than generation length alone, is the primary driver.

\subsection{Claim 3. Few-shot RHI improves cost efficiency through reduced cache read/write usage}
\label{subsec:Claim3}

This claim supports the hypothesis proposed in Section~\ref{sec:Harness} that RHI prioritizes workflow updates over agent design. Specifically, we show that, beyond improving performance, RHI consistently improves cost efficiency across all base models.

For the \texttt{sonnet-4.6} baseline, Figure~\ref{fig:sonnet46_cost} shows that the normalized cost of \texttt{sonnet-4.6-high}$+\mathcal{H}[2]$ ($2.38$) is $7\%$ lower than that of \texttt{sonnet-4.6-max} ($2.56$), while cache read/write usage decreases by $33\%$ ($4.91 \rightarrow 3.31$).

A similar trend is observed for stronger models. For \texttt{opus-4.7} baseline, Figure~\ref{fig:opus47high_cost} shows that the normalized cost of \texttt{opus-4.7-high}$+\mathcal{H}[1]$ ($2.11$) is $18\%$ lower than that of \texttt{opus-4.7-max} ($2.60$), while cache read/write usage decreases by $37\%$ ($3.37 \rightarrow 2.11$) .

For \texttt{opus-4.8} baseline, Figure~\ref{fig:opus48high_cost} shows that the normalized cost of \texttt{opus-4.8-high}$+\mathcal{H}[2]$ ($1.69$) decreases by $23\%$ relative to \texttt{opus-4.8-max} ($2.19$) and by \textbf{60\%} relative to \texttt{opus-4.8-ultracode} ($4.15$). The corresponding cache read/write usage decreases by $32\%$ ($2.51 \rightarrow 1.69$) and by \textbf{64\%} ($4.74 \rightarrow 1.69$), respectively.

We attribute these savings to more efficient context management. RHI learns task-specific
communication contracts: textual interfaces that specify which information should pass from one
agent to another, or from subagents to the orchestrator. These contracts allow each agent to
condition on the information its downstream decisions require, rather than on the full interaction
history, thereby curbing redundant context propagation. Optimizing the multi-agent workflow can
therefore be viewed as learning a task-specific sparsity pattern over inter-agent communication. This is
conceptually analogous to sparse self-attention, where selected entries of the attention pattern
replace dense all-to-all token interactions to improve long-context efficiency
\citep{child2019sparse,beltagy2020longformer,zaheer2020bigbird}. Because inference cost in
Claude-style coding agents depends heavily on cache reads and writes, this reduced context
footprint can translate directly into lower cost.

We caution against reading the larger savings for \texttt{opus-4.7} and \texttt{opus-4.8}
(Figures~\ref{fig:opus47high_cost} and~\ref{fig:opus48high_cost}) as evidence that RHI inherently
becomes more effective as the base model improves. A more conservative interpretation is that
scaling test-time compute on an already strong model yields diminishing performance-per-cost
returns; RHI offers an alternative that restructures the agent's computation through the
prompt-level harness rather than uniformly increasing reasoning effort. 

\begin{tcolorbox}[
    colback=gray!5,
    colframe=black!70,
    title=\textbf{Summary} of Sections \ref{sec:RHI} and \ref{sec:Experiment},
    sharp corners
]
Recursive Harness Self-Improvement (RHI) is a computationally lightweight method that requires only a few iterations yet can substantially raise the performance ceiling of test-time scaling baselines while reducing inference cost by up to $60\%$. This advantage persists across progressively stronger base models. The performance gains from RHI are not explained by increased output-token usage; instead, they are associated with more efficient context management, as reflected by reduced cache read/write usage and lower inference cost.
\end{tcolorbox}

We examine what drives these gains in the next section.
\section{Ablation Study}
\label{sec:Ablation}

Each section addresses a different analysis question. Section~\ref{subsec:rhi_train_time_scaling} examines whether RHI provides benefits beyond \emph{train}-time scaling. Section~\ref{subsec:rhi_harness_factors} investigates the factors driving RHI's performance gains. Finally, Section~\ref{subsec:rhi_implicit_objective} formulates an implicit objective for RHI based on the observations from Section~\ref{subsec:rhi_harness_factors}.

\subsection{Can RHI also raise the performance plateau of train-time scaling?}
\label{subsec:rhi_train_time_scaling}
\begin{figure}[h]
    \centering
    \includegraphics[width=0.9\linewidth]{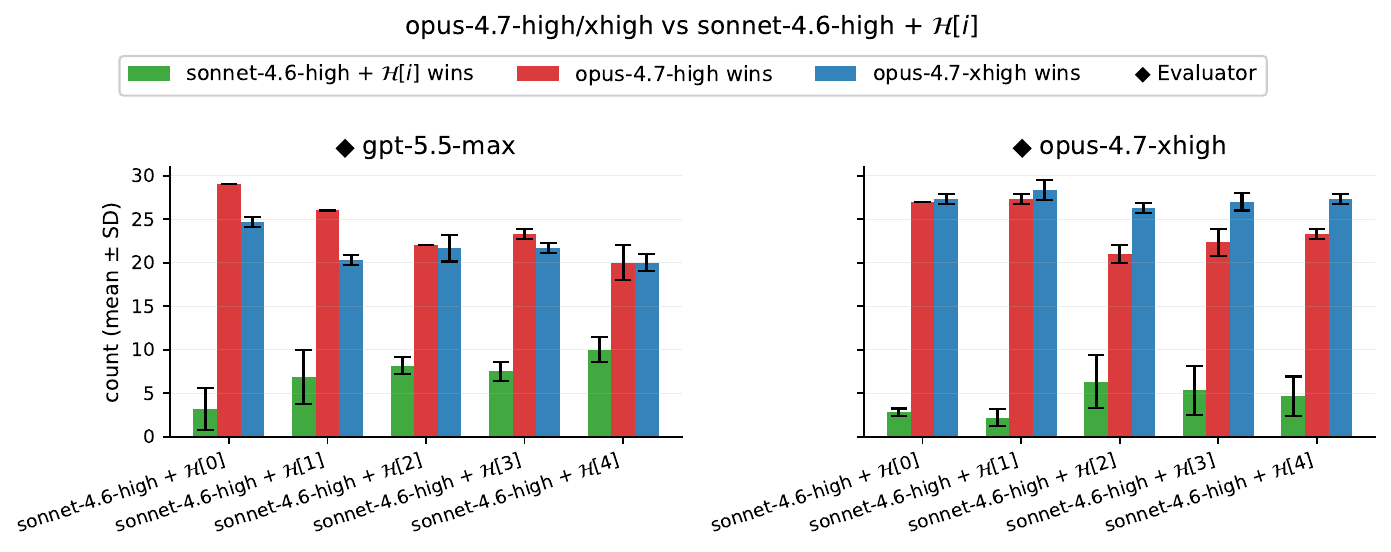}
    \caption{
    \textbf{RHI complements rather than replaces train-time scaling.}
    We compare \texttt{sonnet-4.6-high}$+\mathcal{H}[i]$, $i \in \{0,1,2,3,4\}$, with stronger base-model baselines \texttt{opus-4.7-high} and \texttt{opus-4.7-xhigh}. RHI on the weaker \texttt{sonnet-4.6-high} base model does not consistently close the gap to the stronger \texttt{opus-4.7} baselines.
    }
\label{fig:sonnet46_RHI_trainTS}
\end{figure}

A natural question is whether RHI also raises performance plateaus of train-time scaling by replacing the base model with a stronger model.
To test this, we compare
\texttt{sonnet-4.6-high}$+\mathcal{H}[i]$, $i \in \{0,1,2,3,4\}$, against
\texttt{opus-4.7-high} and \texttt{opus-4.7-xhigh}. Figure~\ref{fig:sonnet46_RHI_trainTS} shows
that RHI substantially improves the weaker \texttt{sonnet-4.6-high} agent, with gains plateauing
at $2 \sim 4$ iterations. Nevertheless, these gains do not consistently close the gap to the
stronger \texttt{opus-4.7} baselines.

We therefore interpret RHI as complementary to, rather than a replacement for, train-time scaling. RHI can improve how a fixed model is used and raise the performance ceiling achievable through same-family test-time scaling, but it does not eliminate the benefits of stronger base models. This observation also leaves room for further algorithmic improvements to RHI.

\subsection{Which harness factors drive RHI performance gains?}
\label{subsec:rhi_harness_factors}
The experiments in Section~\ref{sec:Experiment} suggest that RHI's performance gains are not primarily explained by longer generations. We therefore examine how the harness evolves across RHI iterations. Section~\ref{subsec:full_harness_evolution} analyzes the evolution of the entire harness, while Section~\ref{subsec:harness_component_evolution} investigates which harness components drive RHI's performance gains.

\subsubsection{Full-harness evolution}
\label{subsec:full_harness_evolution}

\paragraph{Visualization.}
We first analyze the evolution of the full prompt-represented harness for
\texttt{sonnet-4.6-high}. For each \(\mathcal{H}_{x}^{(i)}\), \(i\in\{0,1,2,3,4\}\), we embed the
full textual specification and project the embeddings to two dimensions using t-SNE and UMAP. We
assess robustness using two embedding models, OpenAI \texttt{text-embedding-3-large}
\citep{openai2024embeddings} and \texttt{all-mpnet-base-v2}
\citep{song2020mpnet,reimers2019sentence}.

This embedding analysis is a diagnostic rather than a mechanistic explanation. Because the coding
agent is black-box, we cannot directly inspect how a textual harness changes the model's internal
computation or output distribution. We therefore use text embeddings as an external proxy for
semantic change and ask whether the resulting harness trajectories are systematic across
iterations, embedding models, and projection methods.

\begin{figure}[h]
    \centering
    \begin{subfigure}[b]{0.48\textwidth}
        \centering
        \includegraphics[width=\textwidth]{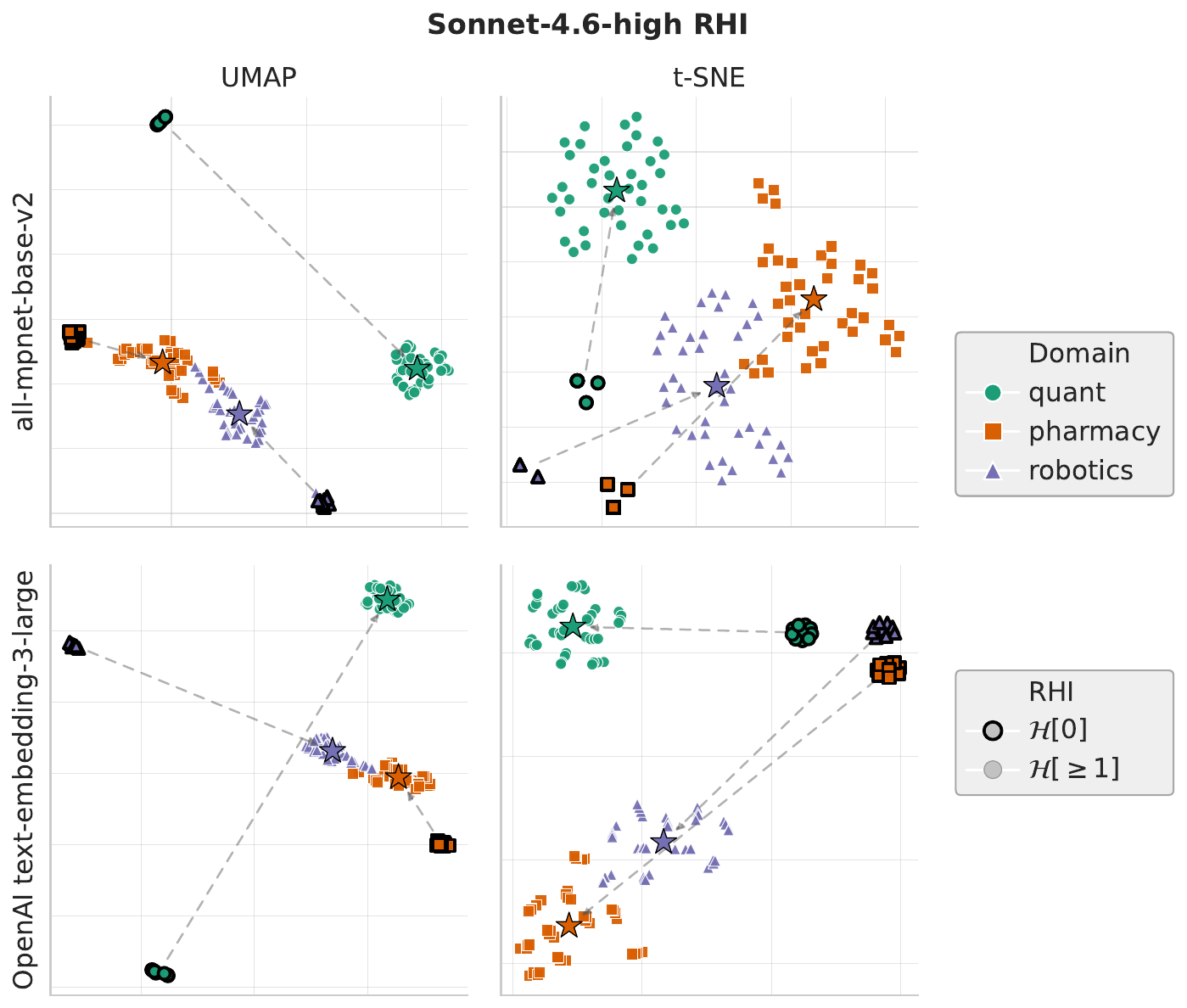}
        \caption{Initial harness \(\mathcal{H}[0]\) versus RHI-improved harnesses \(\mathcal{H}[i]\), \(i\in\{1,2,3,4\}\).}
        \label{fig:wholeharness_umaptsne_1}
    \end{subfigure}
    \hfill
    \begin{subfigure}[b]{0.48\textwidth}
        \centering
        \includegraphics[width=\textwidth]{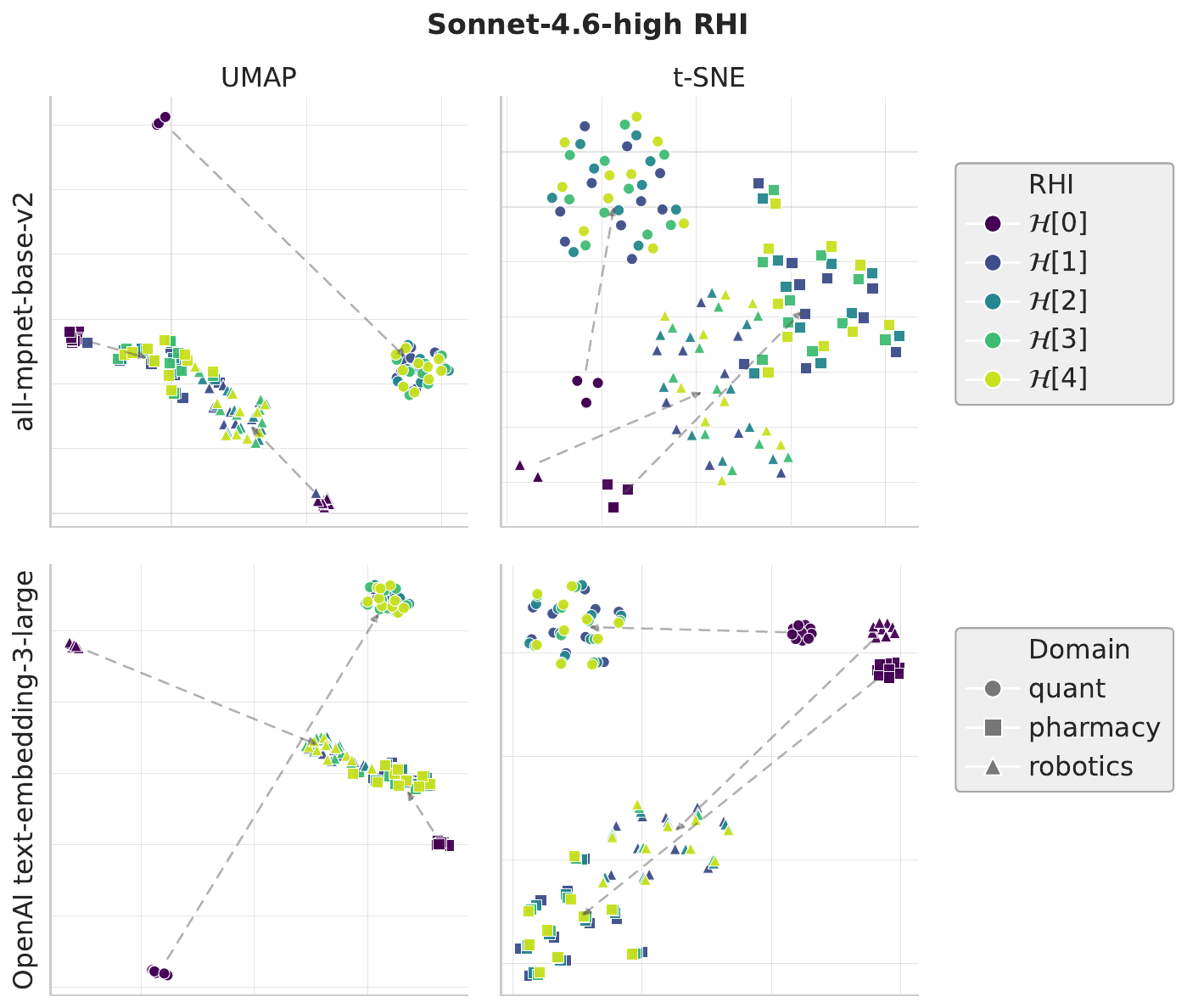}
        \caption{Harness representations by RHI iteration \(\mathcal{H}[i]\), \(i\in\{0,1,2,3,4\}\).}
        \label{fig:wholeharness_umaptsne_2}
    \end{subfigure}
    \begin{subfigure}[b]{0.3\textwidth}
        \centering
        \includegraphics[width=\textwidth]{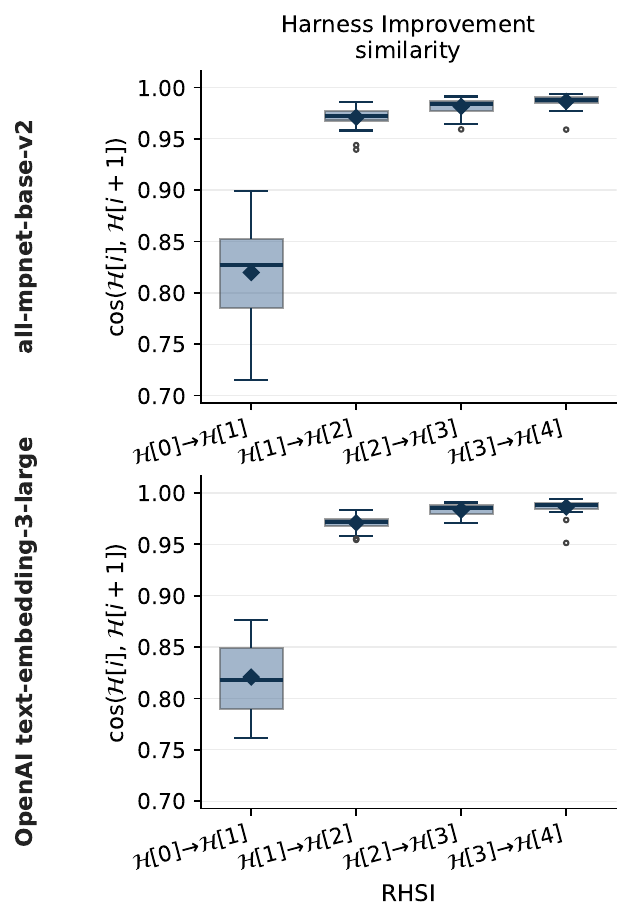}
        \caption{Consecutive harness similarity.}
        \label{fig:wholeharness_his}
    \end{subfigure}
    \begin{subfigure}[b]{0.3\textwidth}
        \centering
        \includegraphics[width=\textwidth]{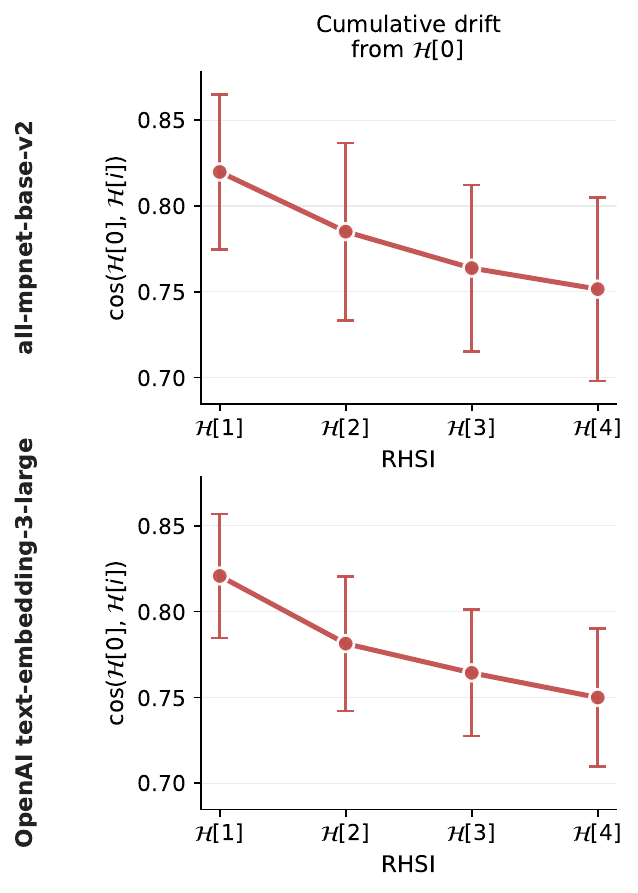}
        \caption{Cumulative drift from \(\mathcal{H}[0]\).}
        \label{fig:wholeharness_cd}
    \end{subfigure}
    \begin{subfigure}[b]{0.3\textwidth}
        \centering
        \includegraphics[width=\textwidth]{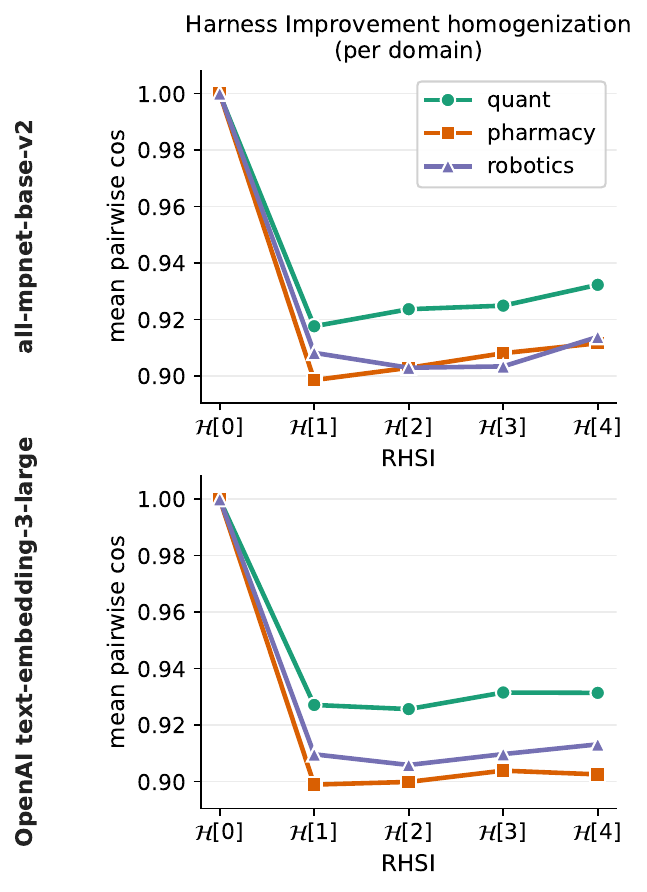}
        \caption{Within-domain across-task similarity.}
        \label{fig:wholeharness_hih}
    \end{subfigure}
    \caption{Evolution of full-harness representations across RHI iterations on \texttt{sonnet-4.6-high}. In figure labels, \(\mathcal{H}[i]\) denotes the main-text notation \(\mathcal{H}^{(i)}\), or \(\mathcal{H}_{x}^{(i)}\) for task \(x\).}
    \label{fig:wholeharness_umaptsne}
\end{figure}

Figures~\ref{fig:wholeharness_umaptsne_1} and~\ref{fig:wholeharness_umaptsne_2} show the same
low-dimensional representations of $\cH[i]$ with different annotations. In Figure~\ref{fig:wholeharness_umaptsne_1},
colors indicate domains and black outlines denote the initial harness \(\mathcal{H}^{(0)}\). In
Figure~\ref{fig:wholeharness_umaptsne_2}, colors indicate RHI iterations and marker shapes indicate
domains.

Figure~\ref{fig:wholeharness_umaptsne_1} shows a clear separation between the initial harness
\(\mathcal{H}^{(0)}\) and the RHI-improved harnesses \(\mathcal{H}^{(1)},\ldots,\mathcal{H}^{(4)}\).
This separation is consistent across embedding models and projection methods. Together with the
performance improvements in Figure~\ref{fig:sonnet46_performance}, thses results suggest that adapting the initial domain-level harness, \(\mathcal{H}^{(0)}\), into task-specific harnesses is one of key factors underlying RHI's performance gains.
However, because this analysis embeds the entire harness as a single text object, it does not identify which harness components account for the performance gains. We address this limitation through a component-level analysis in Section~\ref{subsec:harness_component_evolution}.

\paragraph{Cosine-similarity analysis.}
We next quantify full-harness evolution using cosine similarity. Figure~\ref{fig:wholeharness_his} reports the consecutive similarity between \(\mathcal{H}^{(i)}\) and \(\mathcal{H}^{(i+1)}\). The similarity from \(\mathcal{H}^{(0)}\) to \(\mathcal{H}^{(1)}\) is substantially lower ($0.82$) than that of later transitions ($0.97, 0.98, 0.99$), suggesting that the first RHI update produces the largest semantic change, while subsequent updates are more incremental.

Figure~\ref{fig:wholeharness_cd} reports the cumulative similarity, i.e., $\cos(\cH[0], \cH[i])$, between each harness and the initial harness. The similarity decreases over iterations ($0.82 \rightarrow 0.78 \rightarrow 0.76 \rightarrow 0.75$), indicating that RHI progressively diverges from the initial harness rather than reverting to it.
Finally, Figure~\ref{fig:wholeharness_hih} measures within-domain across-task similarity. For a
domain \(\mathcal{D}\), we compute
\[
\frac{1}{|\mathcal{P}_{\mathcal{D}}|}
\sum_{(x,x')\in\mathcal{P}_{\mathcal{D}}}
\cos\!\left(
\mathrm{emb}(\mathcal{H}_{x}^{(i)}),
\mathrm{emb}(\mathcal{H}_{x'}^{(i)})
\right),
\]
where \(\mathcal{P}_{\mathcal{D}}=\{(x,x') : x,x'\in\mathcal{D},\; x\neq x'\}\). The increasing
within-domain similarity after the first RHI iteration suggests that RHI updates share common
domain-level structure, even as the harnesses move away from their initial specification. Since the
full harness aggregates roles, instructions, contracts, and hops, we next analyze these components
separately.

\subsubsection{Harness-component evolution}
\label{subsec:harness_component_evolution}
\begin{figure}[htbp]
\centering
\begin{subfigure}[b]{0.9\textwidth}
\centering
\includegraphics[width=\textwidth]{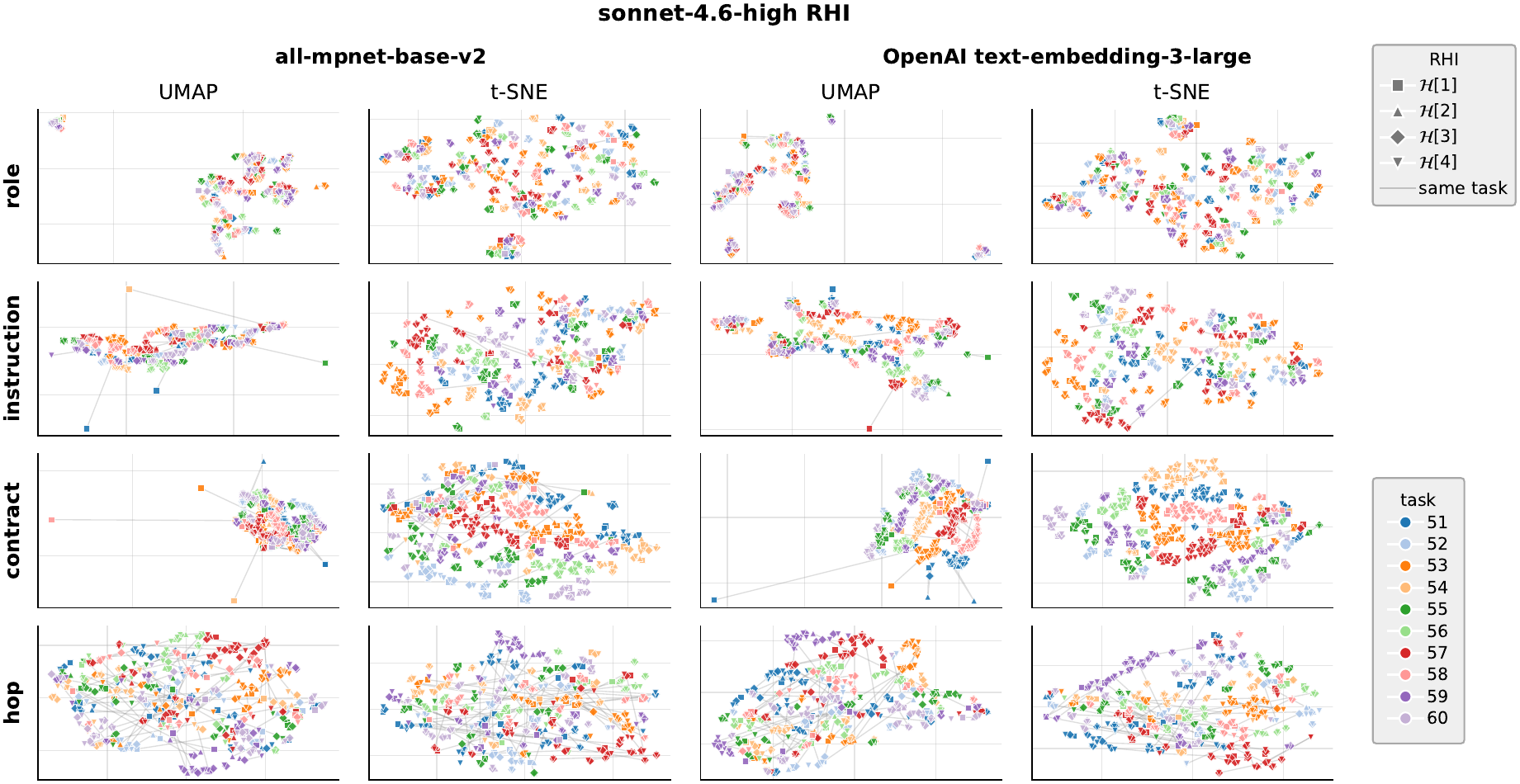}
\caption{Quantitative ML research tasks.}
\label{fig:harnesscomponents_umaptsne_quant}
\end{subfigure}

\begin{subfigure}[b]{0.9\textwidth}
    \centering
    \includegraphics[width=\textwidth]{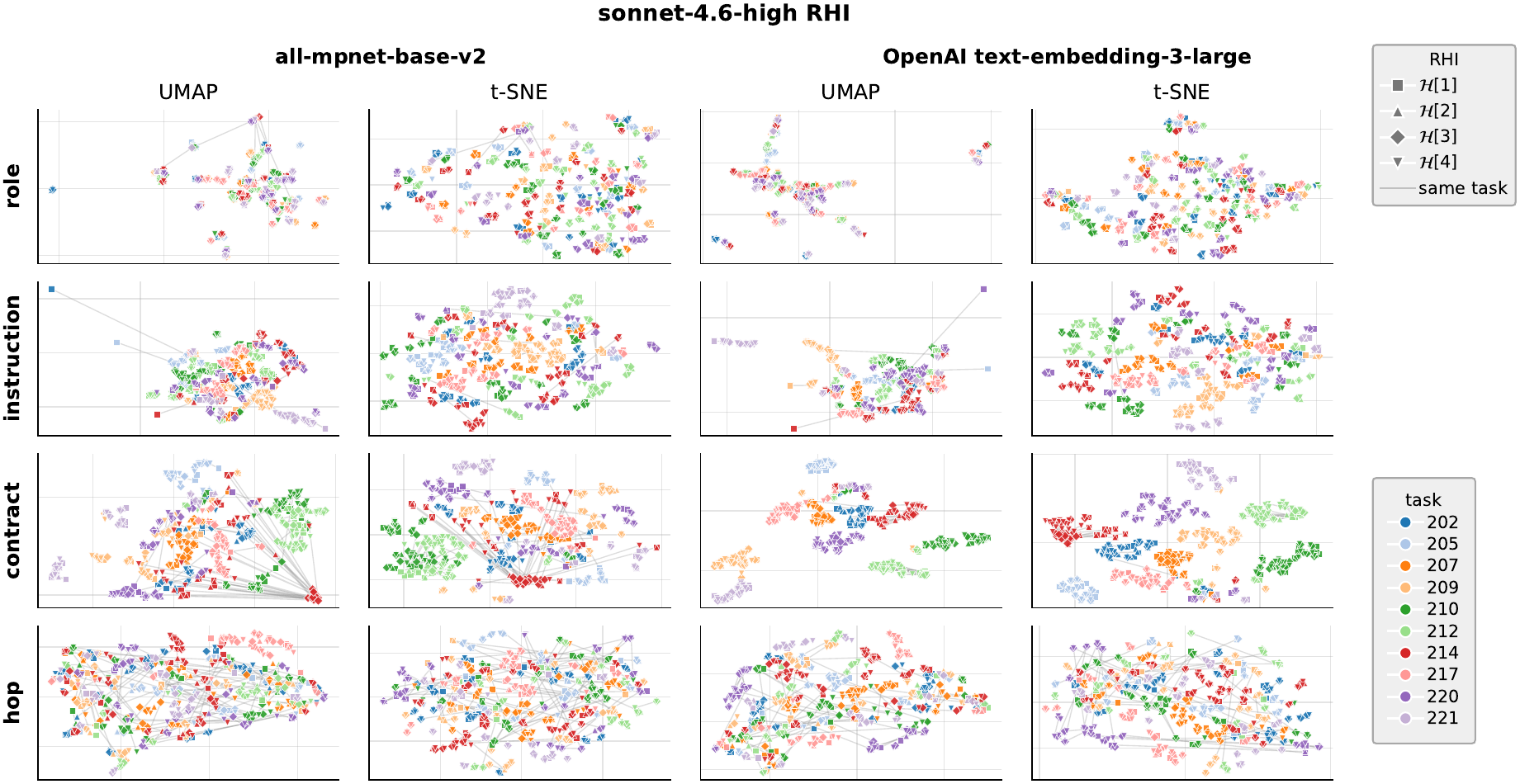}
    \caption{Pharmacy-domain ML research tasks.}
    \label{fig:harnesscomponents_umaptsne_pharmacy}
\end{subfigure}

\begin{subfigure}[b]{0.9\textwidth}
    \centering
    \includegraphics[width=\textwidth]{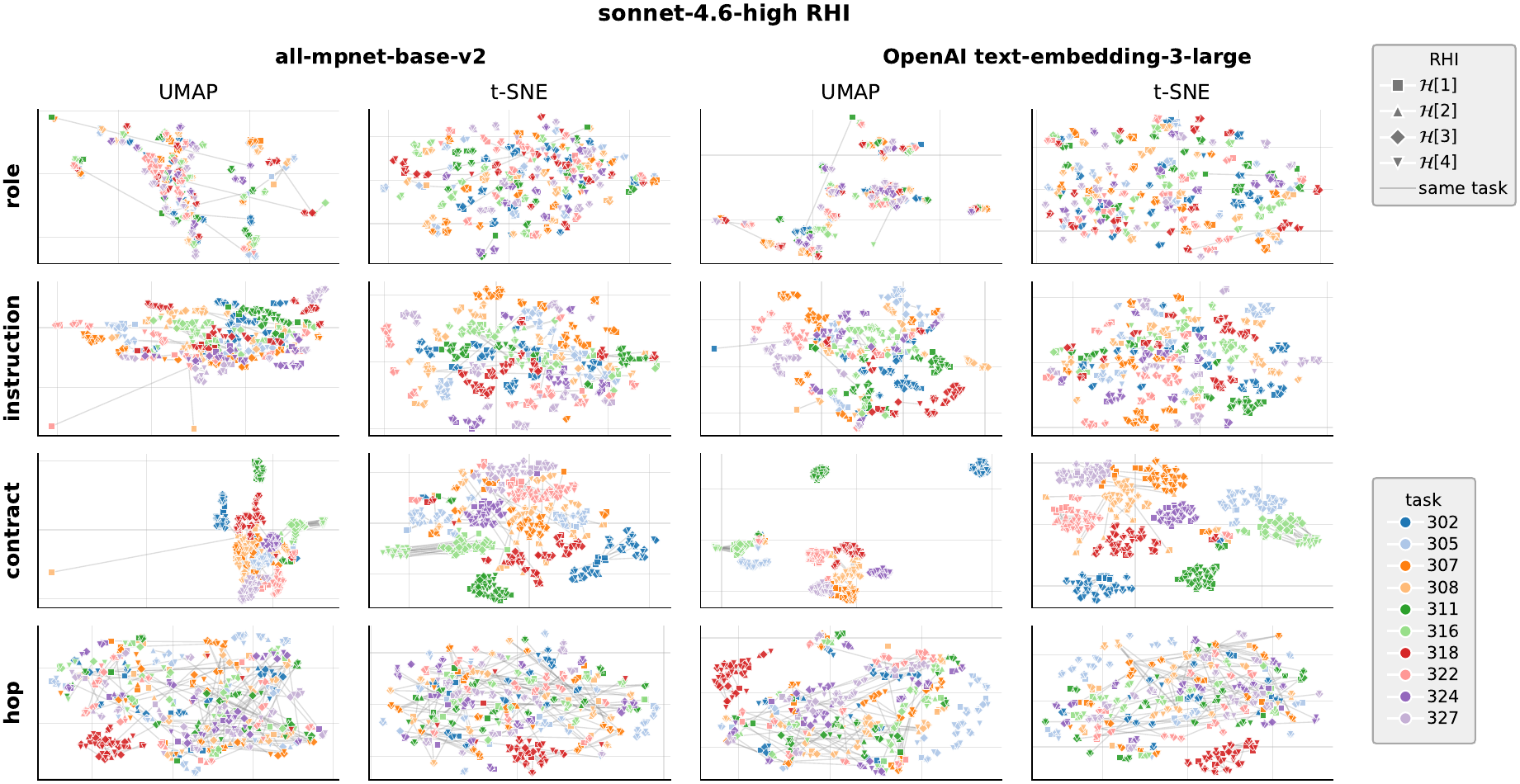}
    \caption{Robotics-domain ML research tasks.}
    \label{fig:harnesscomponents_umaptsne_robotics}
\end{subfigure}

\caption{Evolution of harness-component representations across RHI iterations on \texttt{sonnet-4.6-high}. Each subfigure shows low-dimensional projections of role, instruction, contract, and hop embeddings within a domain; figure labels use \(\mathcal{H}[i]\) for iteration \(i\).}
\label{fig:harnesscomponent_umaptsne}
\end{figure}

\paragraph{Visualization.}
We decompose each harness into the four components used throughout the paper: roles,
instructions, contracts, and hops (see Figure \ref{fig:harness_decomposition}). Roles and instructions specify agent
design; contracts specify what information is passed between subagents and the orchestrator; and
hops specify the multi-agent workflow.

Formally, suppose \(\mathcal{H}_{x}^{(i)}\) specifies \(N_{xi}\) agents and \(M_{xi}\) workflow hops. For
each agent \(n\in[N_{xi}]\), let
\[
h_{xn}^{\texttt{role},(i)},\quad
h_{xn}^{\texttt{instr},(i)},\quad
h_{xn}^{\texttt{cont},(i)}
\in\mathcal{V}^{*}
\]
denote its role, instruction, and contract. For each hop \(m\in[M_{xi}]\),
let
\[
h_{xm}^{\texttt{hop},(i)}\in\mathcal{V}^{*}
\]
denote the corresponding workflow step (see Figure \ref{fig:harness_structure}). We embed each textual component using
\texttt{text-embedding-3-large} and \texttt{all-mpnet-base-v2}, followed by \(\ell_2\)
normalization. If a component exceeds the embedding model's input length, we split it into chunks,
embed each chunk separately, and average the normalized chunk embeddings.

Figure~\ref{fig:harnesscomponent_umaptsne} shows t-SNE and UMAP projections of these component
embeddings. Each row corresponds to a component type: role, instruction, contract, or hop. Marker
shape indicates the RHI iteration \(i\in\{1,2,3,4\}\), and color indicates the task \(x\). For a
fixed task and iteration, the role, instruction, and contract rows contain \(N_{xi}\) points, while
the hop row contains \(M_{xi}\) points.

Across all three domains, Figures~\ref{fig:harnesscomponents_umaptsne_quant}, \ref{fig:harnesscomponents_umaptsne_pharmacy}, and~\ref{fig:harnesscomponents_umaptsne_robotics} show that \emph{contracts exhibit the clearest task-dependent clustering}. This pattern is consistent across embedding models and projection methods, suggesting that the largest semantic changes occur at the information interface between the orchestrator and subagents. \emph{Hops} and \emph{instructions} also exhibit task-dependent clustering, but less distinctly than contracts. In contrast, \emph{roles} are less separable, likely because many ML research tasks in our benchmark share common high-level roles, such as data preparation, model training, validation, and report writing.

Together with the performance results in Section~\ref{sec:Experiment}, these findings suggest that \emph{RHI's gains arise primarily from task-specific coordination rather than longer reasoning traces alone}. In particular, contracts and hops determine what information is communicated and how the multi-agent workflow proceeds. Their prominence is expected because the harness optimizer's system prompt explicitly emphasizes multi-agent workflow refinement, particularly contracts and hops (see Section~\ref{sec:Harness improvement prompt}). We therefore interpret the observed component-level separation as evidence consistent with RHI learning task-specific coordination interfaces, while emphasizing that the embedding analysis is correlational rather than causal.

\paragraph{Cosine-similarity analysis.}

\begin{figure}[htbp]
\centering
\begin{subfigure}[b]{0.3\textwidth}
\centering
\includegraphics[width=\textwidth]{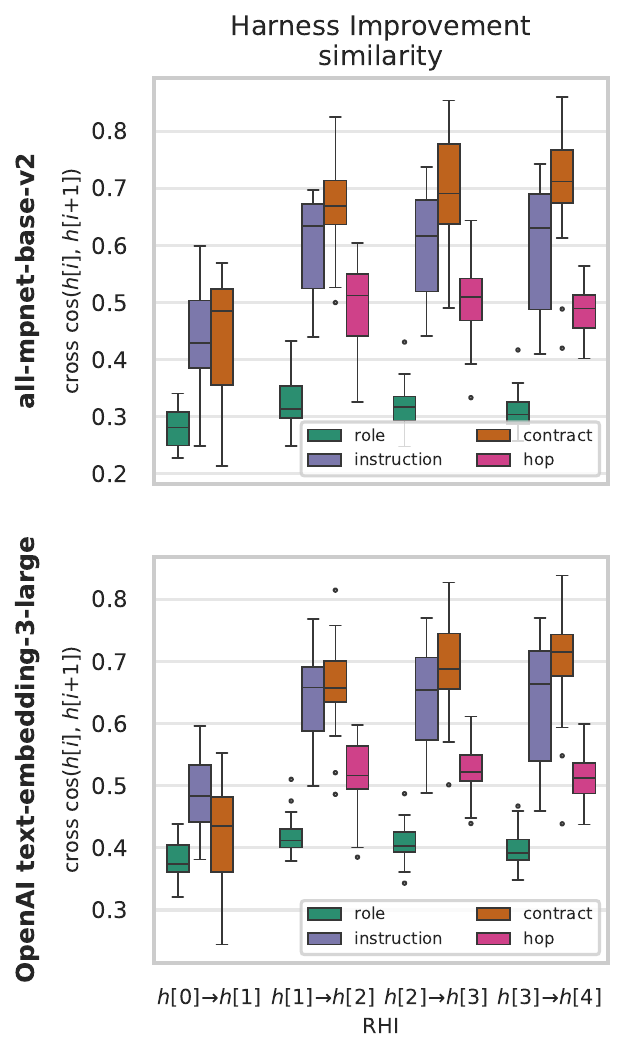}
\caption{Consecutive component similarity.}
\label{fig:harnesscomponents_his}
\end{subfigure}
\begin{subfigure}[b]{0.3\textwidth}
\centering
\includegraphics[width=\textwidth]{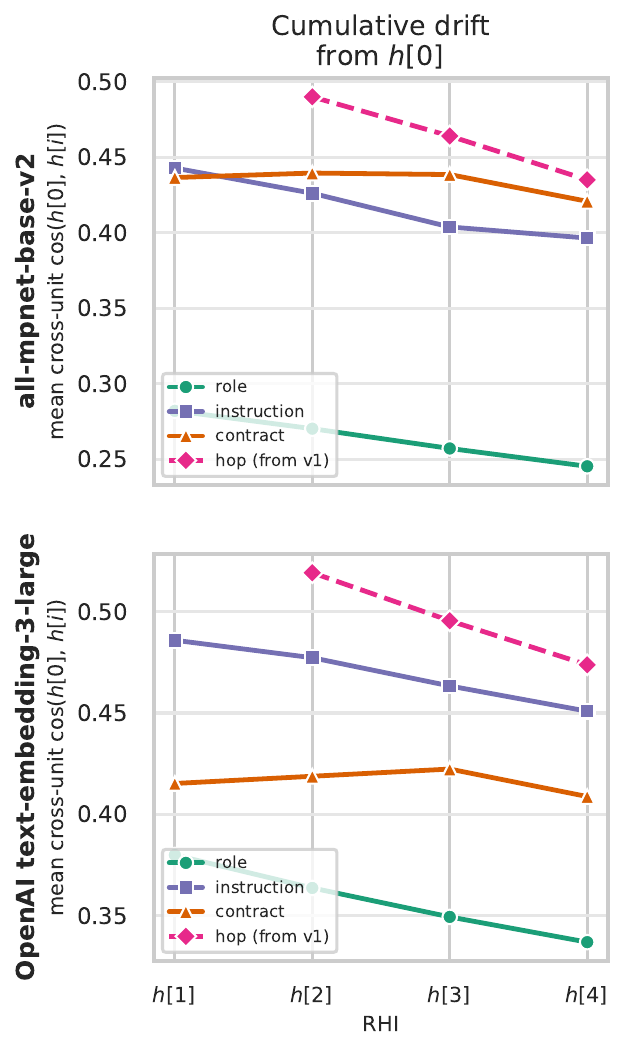}
\caption{Cumulative component drift from iteration \(0\).}
\label{fig:harnesscomponents_chd}
\end{subfigure}
\begin{subfigure}[b]{0.3\textwidth}
\centering
\includegraphics[width=\textwidth]{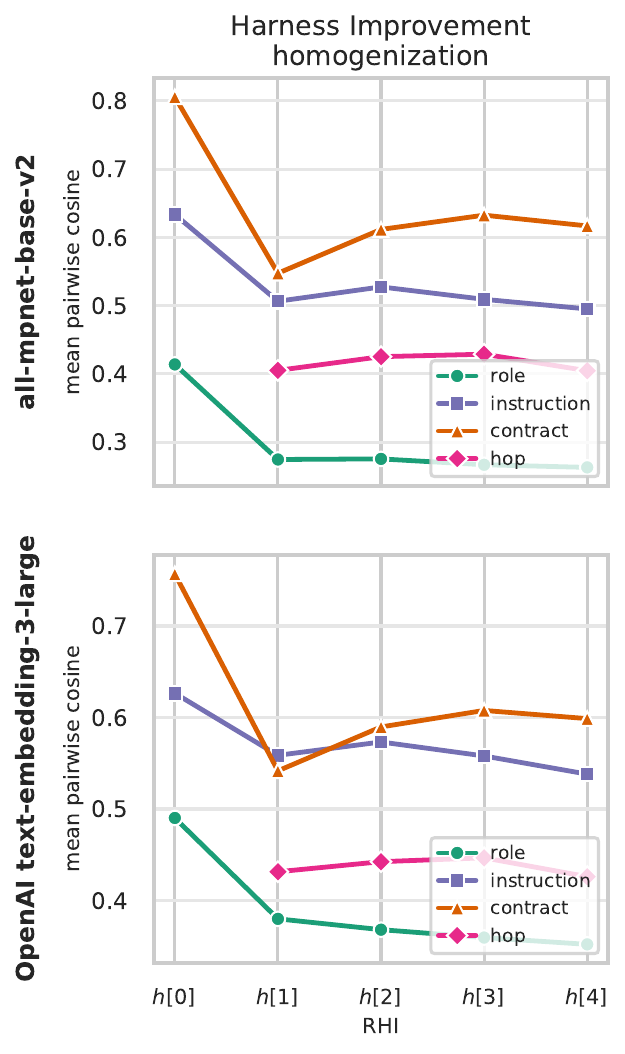}
\caption{Across-task component similarity.}
\label{fig:harnesscomponents_hih}
\end{subfigure}
\caption{Cosine-similarity analysis of harness components across RHI iterations. We measure consecutive similarity, cumulative drift from the initial harness, and across-task similarity for each component type.}
\label{fig:harnesscomponent_analysis}
\end{figure}

We next quantify component evolution with cosine similarity. Let
\(\texttt{hc}\in\{\texttt{role},\texttt{instr},\texttt{cont},\texttt{hop}\}\). For task \(x\),
iteration \(i\), and component type \(\texttt{hc}\), let \(K_{xi}^{\texttt{hc}}\) be the number of
components of that type: \(K_{xi}^{\texttt{hc}}=N_{xi}\) for roles, instructions, and contracts,
and \(K_{xi}^{\texttt{hc}}=M_{xi}\) for hops. Let \(z_{xk}^{\texttt{hc},(i)}\) denote the normalized
embedding of the \(k\)-th component of type \(\texttt{hc}\). We measure the similarity between $\texttt{hc}^{(i)}$ and  $\texttt{hc}^{(j)}$ by averaging the pairwise cosine similarities between all components of type $\texttt{hc}$ across the two iterations, and then averaging over all tasks. We define this quantity as the $\mathrm{cross\_cos}$ similarity:
\[
\mathrm{cross\_cos}(\texttt{hc}^{(i)},\texttt{hc}^{(j)})
=
\frac{1}{|\mathcal{X}|}
\sum_{x\in\mathcal{X}}
\frac{1}{K_{xi}^{\texttt{hc}}K_{xj}^{\texttt{hc}}}
\sum_{k=1}^{K_{xi}^{\texttt{hc}}}
\sum_{k'=1}^{K_{xj}^{\texttt{hc}}}
\cos\!\left(
 z_{xk}^{\texttt{hc},(i)},
 z_{xk'}^{\texttt{hc},(j)}
\right).
\]
Figure~\ref{fig:harnesscomponents_his} reports
\(\mathrm{cross\_cos}(\texttt{hc}^{(i)},\texttt{hc}^{(i+1)})\), while
Figure~\ref{fig:harnesscomponents_chd} reports similarity to the initial harness,
\(\mathrm{cross\_cos}(\texttt{hc}^{(0)},\texttt{hc}^{(i)})\).

For across-task similarity at a fixed iteration, let
\(\mathcal{P}=\{(x,x') : x,x'\in\mathcal{X},\; x\neq x'\}\). We define
\[
\mathrm{pair\_cos}(\texttt{hc}^{(i)})
=
\frac{1}{|\mathcal{P}|}
\sum_{(x,x')\in\mathcal{P}}
\frac{1}{K_{xi}^{\texttt{hc}}K_{x'i}^{\texttt{hc}}}
\sum_{k=1}^{K_{xi}^{\texttt{hc}}}
\sum_{k'=1}^{K_{x'i}^{\texttt{hc}}}
\cos\!\left(
 z_{xk}^{\texttt{hc},(i)},
 z_{x'k'}^{\texttt{hc},(i)}
\right).
\]
Figure~\ref{fig:harnesscomponents_hih} reports this quantity.

Figure~\ref{fig:harnesscomponents_his} shows that contracts reach high consecutive similarity ($0.48 \rightarrow 0.66 \rightarrow 0.69 \rightarrow 0.72$) earlier than the other components, indicating that contract updates stabilize relatively quickly. Hops and instructions exhibit intermediate stabilization, whereas roles change more gradually ($0.28 \rightarrow 0.31 \rightarrow 0.33 \rightarrow 0.31$). Together with the low-dimensional visualizations in Figure~\ref{fig:harnesscomponent_analysis}, these results suggest that RHI specializes contracts to the task within a few iterations, while roles, instructions, and hops continue to evolve more gradually. One possible explanation is that the pairwise feedback from $\cL_{\texttt{eval}}$ provides a stronger learning signal for contract refinement than for the other components. Alternatively, it may indicate that our current feedback loop design is sufficient to drive effective contract updates but not the refinement of other components. This suggests that improving the quality and specificity of evaluator feedback is a promising direction for future work.

Figures~\ref{fig:harnesscomponents_chd} and~\ref{fig:harnesscomponents_hih} exhibit trends similar to those in Figures~\ref{fig:wholeharness_cd} and~\ref{fig:wholeharness_hih}; therefore, we omit a detailed discussion.

\subsection{What objective function does RHI implicitly optimize?}
\label{subsec:rhi_implicit_objective}

RHI is not trained against an explicit scalar reward or explicit loss function. At iteration \(i\),
the harness optimizer \(\mathcal{L}_{\mathrm{harness}}\) receives the current harness
\(H_x^{(i)}\) and the accumulated self-comparison history
\(\mathcal{D}_x^{(i)}\), where the history is produced by pairwise judgments from
\(\mathcal{L}_{\mathrm{eval}}\). It then proposes the next textual harness
\(H_x^{(i+1)}\); see Step~5 of Figure~\ref{fig:rhsi}. Thus, RHI is best understood as black-box
search over prompt-represented harnesses under an implicit preference induced by the $\cL_{\textrm{harness}}$
prompt, the base model underlying $\cL_{\textrm{harness}}$, and addtionally feedback from $\cL_{\textrm{eval}}$.

Therefore, the goal of this section is not to identify the true latent objective of $\cL_{\textrm{harness}}$ or to test whether such an objective improves the harness. Instead, we propose a plausible objective function that explains the harness update trajectory observed in our experiments, motivated by the analysis in Section~\ref{subsec:rhi_harness_factors}. Based on these observations, we formulate an information-theoretic objective as a \emph{hypothesis} for future harness optimization methods.


\begin{figure}[ht]
  \centering
  \begin{subfigure}{0.5\textwidth}
    \centering
    \includegraphics[width=\linewidth]{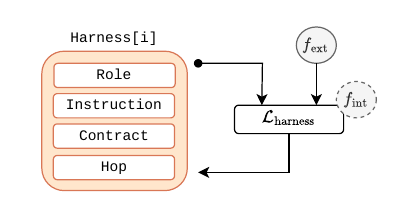}
    \caption{Harness updates are shaped by a controllable external factor \(f_{\mathrm{ext}}\) and a model-dependent internal factor \(f_{\mathrm{int}}\).}
    \label{fig:RHIobjective}
  \end{subfigure}
  \begin{subfigure}{0.45\textwidth}
    \centering
    \includegraphics[width=\linewidth]{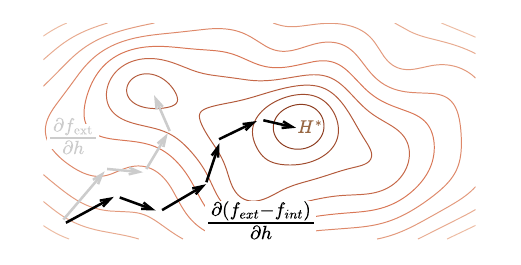}
    \caption{We interpret \(f_{\mathrm{int}}\) as \emph{functional specialization guidance}, which suppresses redundant overlap and encourages harness components to specialize into distinct functional modules.}
    \label{fig:RHIobjective_gradient}
  \end{subfigure}
    \caption{Schematic illustration of the objective implicitly induced by RHI. The external factor \(f_{\mathrm{ext}}\) is controlled by the \(\cL_{\textrm{harness}}\) prompt, while the internal factor \(f_{\mathrm{int}}\) is determined by how the base model underlying \(\cL_{\textrm{harness}}\) interprets feedback and generates next harness.}
  \label{RHIobjective_total}
\end{figure}


\subsubsection{Evidence for an implicit objective function}

Before presenting our hypothesis, we first provide empirical evidence that RHI exhibits an implicit objective. The harness component analysis in Section~\ref{subsec:harness_component_evolution} provides two observations. First, Figure~\ref{fig:harnesscomponents_his} shows systematic stabilization across iterations, most clearly for contracts. Second, Figure~\ref{fig:harnesscomponent_umaptsne} shows that the evolved component representations remain task-dependent, especially for contracts and, to a lesser extent, hops and instructions. Thus, the update trajectory is neither arbitrary nor collapsing to a generic harness template. Instead, it is consistent with movement toward task-specific harness components.

These observations do not prove that RHI optimizes a unique scalar objective. They do, however, motivate modeling its update trajectory as being guided by an implicit preference over task-specific harness components.

\subsubsection{Information-theoretic hypothesis for the implicit objective}

We now state the information-theoretic hypothesis suggested by these observations using the same
component notation as above. Let \(X\sim p_{\mathcal{X}}\) denote a random task sampled from the
benchmark distribution, and let \(x\) denote a realized task. Let \(g_i\) be the task-to-harness
mapping induced by RHI at iteration \(i\), so that \(\mathcal{H}_{x}^{(i)}=g_i(x)\). Let
\(\mathcal{C}\) denote a general set of harness-component types. For a task \(x\), iteration \(i\),
and component type
\(\texttt{hc}\in\mathcal{C}\), let \(K_{xi}^{\texttt{hc}}\) be the number of components of that
type and write
\[
h_{xk}^{\texttt{hc},(i)}\in\mathcal{V}^{*},
\qquad
k\in[K_{xi}^{\texttt{hc}}],
\]
for the \(k\)-th component of type \(\texttt{hc}\). Thus, in our current harness,
\(h_{xn}^{\texttt{role},(i)}\), \(h_{xn}^{\texttt{instr},(i)}\), and
\(h_{xn}^{\texttt{cont},(i)}\) denote the role, instruction, and contract of agent \(n\), while
\(h_{xm}^{\texttt{hop},(i)}\) denotes workflow hop \(m\). Let
\(z_{xk}^{\texttt{hc},(i)}\) denote the normalized embedding of
\(h_{xk}^{\texttt{hc},(i)}\), as in the component-similarity analysis above. The corresponding
random embedding is \(z_{Xk}^{\texttt{hc},(i)}\). Finally, let
\(\mathcal{C}_{\mathrm{ext}}\subseteq\mathcal{C}\) denote the component types emphasized by the
$\cL_{\texttt{harness}}$ prompt; in our current RHI implementation,
\(\mathcal{C}=\{\texttt{role},\texttt{instr},\texttt{cont},\texttt{hop}\},~\mathcal{C}_{\mathrm{ext}}=\{\texttt{cont},\texttt{hop}\}\).

\begin{tcolorbox}[
breakable,
colback=gray!5,
colframe=black!70,
title=\textbf{Hypothesis: information-theoretic implicit objective of RHI}
]
We hypothesize that the observed RHI trajectory is consistent with increasing the following
information-theoretic functional over task-specific harness mappings:
\begin{equation}
\begin{aligned}
    J(g_i)
    =
    &\underbrace{\sum_{\texttt{hc}\in\mathcal{C}_{\mathrm{ext}}}
    \frac{1}{K_{Xi}^{\texttt{hc}}}
    \sum_{k=1}^{K_{Xi}^{\texttt{hc}}}
    I\!\left(z_{Xk}^{\texttt{hc},(i)};X\right)}_{f_{\mathrm{ext}}} \\
    &-
    \beta
    \underbrace{\mathrm{TC}\!\left(
    \left\{z_{Xk}^{\texttt{hc},(i)}:
    \texttt{hc}\in\mathcal{C},\; k\in[K_{Xi}^{\texttt{hc}}]\right\}
    \mid X\right)}_{f_{\mathrm{int}}},
    \qquad \beta>0.
\end{aligned}
\label{eq:objectiveRHI}
\end{equation}
Here, $f_{\mathrm{ext}}$ measures task information in the externally emphasized component types, while
$f_{\mathrm{int}}$ measures task-conditional dependence among all harness-component
representations. Averaging over \(K_{Xi}^{\texttt{hc}}\) in $f_{\mathrm{ext}}$ prevents component types with more instances from dominating the objective solely because they contain more entries.
\end{tcolorbox}

We hypothesize that the implicit objective consists of two complementary factors (Figure~\ref{fig:RHIobjective}). The first is an external, controllable factor, \(f_{\mathrm{ext}}\), specified by the system prompt of \(\mathcal{L}_{\mathrm{harness}}\). In our implementation, the prompt emphasizes improving the multi-agent workflow, particularly communication contracts and hops (Figure~\ref{fig:harness_decomposition}; see system prompt of $\cL_{\mathrm{harness}}$ in Appendix~\ref{sec:Harness improvement prompt}). Consequently, \(f_{\mathrm{ext}}\) encourages RHI to optimize harness components related to information flow and orchestration. The second is an internal, model-dependent factor, \(f_{\mathrm{int}}\), arising from the base LLM underlying \(\mathcal{L}_{\mathrm{harness}}\). This factor determines how textual feedback is interpreted and translated into edits of roles, instructions, contracts, and hops. We interpret \(f_{\mathrm{int}}\) as providing \emph{functional specialization guidance}: while \(f_{\mathrm{ext}}\) promotes task-specific workflow refinement, \(f_{\mathrm{int}}\) encourages harness components to specialize into distinct functional roles by reducing task-conditioned redundancy (Figure~\ref{fig:RHIobjective_gradient}).

The overall intuition is that maximizing task information alone admits degenerate solutions in which multiple harness components repeat the same task description or coordination rule. Functional specialization guidance instead encourages a modular decomposition: roles define expertise, instructions define agent behavior, contracts define the information to communicate, and hops define the control flow. 

The objective in Equation~\eqref{eq:objectiveRHI} formalizes this intuition. Specifically, \(f_{\mathrm{ext}}\) measures how well the prompt-emphasized harness components encode task information useful for decomposition, communication, and orchestration. In our implementation, these components are contracts and hops. Meanwhile, \(f_{\mathrm{int}}\) does not act as a classical regularizer that trades off the primary objective against model complexity.Instead, we interpret it as \emph{functional specialization guidance}\footnote{This terminology is loosely inspired by classifier-free diffusion guidance, where conditional and unconditional score estimates are combined to steer generation toward condition-relevant structure~\citep{ho2022classifierfree}. The analogy is conceptual rather than algebraic: here, the task-conditional total-correlation term does not define a diffusion score, but it plays a similar steering role by pushing the harness away from generic redundant overlap and toward distinct component functions.}: a cooperative term that resolves degeneracy in $f_{\mathrm{ext}}$ by favoring harnesses whose components carry non-overlapping information after conditioning on the task.

This hypothesis provides one plausible characterization of how general harness optimization should behave. Specifically, we hypothesize that making harness components more task-specific alone may still yield a suboptimal harness if the components remain semantically redundant. In contrast, reducing redundancy while preserving task specificity can produce a more effective harness (Figure~\ref{fig:RHIobjective_gradient}).

The empirical analyses in the Sections~\ref{subsec:Evidence for increasing ext} and \ref{subsec:Evidence for decreasing int}  instantiate this general hypothesis with the four component types used
in our current RHI harness,
\(\mathcal{C}=\{\texttt{role},\texttt{instr},\texttt{cont},\texttt{hop}\}\). For each
\(\mathcal{H}_{x}^{(i)}\), the total number of textual components is
\(\sum_{\texttt{hc}\in\mathcal{C}}K_{xi}^{\texttt{hc}}\). Operationally, our total-correlation
estimator groups these components into the four typed blocks above, rather than estimating a
separate covariance block for every individual agent-specific component.

\subsubsection{Evidence for increasing \(f_{\mathrm{ext}}\)}
\label{subsec:Evidence for increasing ext}
We first test whether the components emphasized by \(f_{\mathrm{ext}}\) become more informative
about the task. For each task \(x\), iteration \(i\), component type \(\texttt{hc}\), and component
index \(k\), we estimate the mutual information between the component representation and the task
representation:
\[
    I\left(\texttt{emb}(h_{xk}^{\texttt{hc},(i)});\texttt{emb}(x)\right),
    \qquad
    \texttt{hc} \in \{\texttt{role},\texttt{instr},\texttt{cont},\texttt{hop}\},\quad
    k\in[K_{xi}^{\texttt{hc}}].
\]
We denote this diagnostic by \(I(\texttt{hc};\texttt{task})\). Embeddings are computed using either
\texttt{text-embedding-3-large} or \texttt{all-mpnet-base-v2}. We reduce the embeddings to
\(d=10\) dimensions using a globally fitted whitening PCA and then compute Gaussian
canonical-correlation mutual information,
\[
I(\texttt{hc};\texttt{task})
    =
    -\frac{1}{2}
    \sum_r \log(1-\rho_r^2),
\]
where \(\rho_r\) are the canonical correlations between the component and task blocks. Because
finite-sample mutual-information estimates are positively biased, we report both raw plug-in
estimates and permutation-debiased estimates obtained by shuffling task labels. The absolute values
should be interpreted as diagnostic statistics rather than exact mutual information of the raw
texts.

\begin{table}[h]
\centering
\footnotesize
\setlength{\tabcolsep}{4pt}
\resizebox{\textwidth}{!}{%
\begin{tabular}{@{}l cccc cccc cccc cccc@{}}
\toprule
 & \multicolumn{8}{c}{\texttt{text-embedding-3-large}} & \multicolumn{8}{c}{\texttt{all-mpnet-base-v2}} \\
\cmidrule(lr){2-9}\cmidrule(lr){10-17}
 & \multicolumn{4}{c}{undebiased} & \multicolumn{4}{c}{debiased}
 & \multicolumn{4}{c}{undebiased} & \multicolumn{4}{c}{debiased} \\
\cmidrule(lr){2-5}\cmidrule(lr){6-9}\cmidrule(lr){10-13}\cmidrule(lr){14-17}
harness components & 1 & 2 & 3 & 4 & 1 & 2 & 3 & 4 & 1 & 2 & 3 & 4 & 1 & 2 & 3 & 4 \\
\midrule
role        & 0.63 & 0.54 & 0.45 & $0.42\,\downarrow$ & 0.47 & 0.41 & 0.36 & $0.33\,\downarrow$ & 0.44 & 0.41 & 0.31 & $0.28\,\downarrow$ & 0.28 & 0.29 & 0.21 & $0.19\,\downarrow$ \\
instruction & 1.14 & 1.15 & 1.12 & $1.09\,\downarrow$ & 0.99 & 1.03 & 1.02 & 1.00 & 0.96 & 0.88 & 0.91 & $0.82\,\downarrow$ & 0.80 & 0.76 & 0.81 & $0.73\,\downarrow$ \\
\textcolor{red!40!white}{\textbf{contract}}  & $\underline{1.14}$ & 1.25 & 1.34 & $\underline{1.42}\,\textcolor{red!40!white}{\boldsymbol{\uparrow}}$ & $\underline{0.99}$ & 1.13 & 1.24 & $\underline{1.34}\,\textcolor{red!40!white}{\boldsymbol{\uparrow}}$ & $\underline{0.77}$ & 0.81 & 0.91 & $\underline{0.98}\,\textcolor{red!40!white}{\boldsymbol{\uparrow}}$ & $\underline{0.62}$ & 0.70 & 0.81 & $\underline{0.90}\,\textcolor{red!40!white}{\boldsymbol{\uparrow}}$ \\
\textcolor{red!40!white}{\textbf{hop}}         & $\underline{2.10}$ & 2.30 & 2.46 & $\underline{2.66}\,\textcolor{red!40!white}{\boldsymbol{\uparrow}}$ & $\underline{1.96}$ & 2.17 & 2.33 & $\underline{2.54}\,\textcolor{red!40!white}{\boldsymbol{\uparrow}}$ & $\underline{1.89}$ & 2.00 & 2.14 & $\underline{2.17}\,\textcolor{red!40!white}{\boldsymbol{\uparrow}}$ & $\underline{1.74}$ & 1.87 & 2.03 & $\underline{2.05}\,\textcolor{red!40!white}{\boldsymbol{\uparrow}}$ \\
\bottomrule
\end{tabular}}
\caption{
Task mutual information \(I(\texttt{hc};\texttt{task})\) in nats for four harness components across four evaluation configurations: two embedding models and two bias treatments. Arrows indicate the direction of change for RHI iteration $i=1$ to $i=4$.
}
\label{tab:max_f_ext}
\end{table}

Table~\ref{tab:max_f_ext} shows that contracts and hops---the components emphasized by the
harness-optimizer prompt---increase monotonically from iteration $i=1$ to $i=4$ across all encoder and debiasing configurations. 
By contrast, roles decrease monotonically, and instructions are approximately flat or mildly decreasing. This pattern is
consistent with the external term in Equation~\eqref{eq:objectiveRHI}: RHI selectively increases
task information in the components targeted by the optimizer prompt, rather than uniformly making
all harness fields more task-specific.

\subsubsection{Evidence for decreasing \(f_{\mathrm{int}}\)}
\label{subsec:Evidence for decreasing int}
We next test whether RHI reduces redundancy among harness components after conditioning on the
task. Total correlation measures statistical dependence among multiple variables:
\[
  \mathrm{TC}(\texttt{role},\texttt{instr},\texttt{cont},\texttt{hop})
  =
  \sum_{\texttt{hc}} H(\texttt{hc})
  -
  H(\texttt{role},\texttt{instr},\texttt{cont},\texttt{hop}),
\]
where \(\texttt{hc}\in\{\texttt{role},\texttt{instr},\texttt{cont},\texttt{hop}\}\) and
\(H(\cdot)\) is differential entropy. Under a Gaussian approximation with shrunk covariance
\(S\) and per-component diagonal blocks \(S_{\texttt{hc}}\), this becomes
\begin{equation}
  \mathrm{TC}(\texttt{role},\texttt{instr},\texttt{cont},\texttt{hop})
  =
  \frac{1}{2}
  \left[
  \sum_{\texttt{hc}}\log\det S_{\texttt{hc}}
  -
  \log\det S
  \right].
  \label{eq:gtc}
\end{equation}

Unconditional dependence can conflate useful task adaptation with redundancy, because every
component is conditioned on the same task. We therefore estimate task-conditional total correlation
by per-task centering: for each task, we subtract the within-task mean representation of each
component and apply Equation~\eqref{eq:gtc} to the residuals. Under an approximately homoscedastic
within-task covariance model, this estimates
\(\mathrm{TC}(\texttt{role},\texttt{instr},\texttt{cont},\texttt{hop}\mid\texttt{task})\). The
result should be read as an embedding-based proxy for residual redundancy after the shared task
signal has been removed.\footnote{The hop block is design-level: it provides one vector per
\((x,i)\) and is therefore constant across agents for a fixed task and iteration. Per-task
centering removes this block, so the per-agent task-conditional estimate effectively measures
\(\mathrm{TC}(\texttt{role},\texttt{instr},\texttt{cont}\mid\texttt{task})\). Hop contributes to
unconditional total correlation but not to this per-agent conditional estimate.}

As above, we report raw plug-in estimates and permutation-debiased estimates. The debiased estimate
subtracts a null value obtained by independently shuffling component rows, which breaks
cross-component dependence while preserving each component's marginal distribution.

\begin{table}[h]
\centering
\footnotesize
\setlength{\tabcolsep}{4pt}
\resizebox{\textwidth}{!}{%
\begin{tabular}{@{}l cccc cccc cccc cccc@{}}
\toprule
 & \multicolumn{8}{c}{\texttt{text-embedding-3-large}} & \multicolumn{8}{c}{\texttt{all-mpnet-base-v2}} \\
\cmidrule(lr){2-9}\cmidrule(lr){10-17}
 & \multicolumn{4}{c}{undebiased} & \multicolumn{4}{c}{debiased}
 & \multicolumn{4}{c}{undebiased} & \multicolumn{4}{c}{debiased} \\
\cmidrule(lr){2-5}\cmidrule(lr){6-9}\cmidrule(lr){10-13}\cmidrule(lr){14-17}
total correlation & 1 & 2 & 3 & 4 & 1 & 2 & 3 & 4 & 1 & 2 & 3 & 4 & 1 & 2 & 3 & 4 \\
\midrule
$TC$            & 7.53 & 7.09 & 6.66 & $6.36\,\downarrow$ & 6.66 & 6.35 & 6.03 & $5.81\,\downarrow$ & 5.71 & 5.44 & 5.14 & $4.72\,\downarrow$ & 4.82 & 4.70 & 4.53 & $4.19\,\downarrow$ \\
 $\textcolor{red!40!white}{\boldsymbol{TC\mid\texttt{task}}}$  & $\underline{5.18}$ & 4.51 & 4.08 & $\underline{3.92}\,\textcolor{red!40!white}{\boldsymbol{\downarrow}}$ & $\underline{4.84}$ & 4.19 & 3.77 & $\underline{3.63}\,\textcolor{red!40!white}{\boldsymbol{\downarrow}}$ & $\underline{3.89}$ & 3.32 & 2.92 & $\underline{2.86}\,\textcolor{red!40!white}{\boldsymbol{\downarrow}}$ & $\underline{3.51}$ & 3.01 & 2.65 & $\underline{2.62}\,\textcolor{red!40!white}{\boldsymbol{\downarrow}}$ \\
\bottomrule
\end{tabular}}
\caption{
Total correlation among the four harness components \((\texttt{role},\texttt{instr},\texttt{cont},\texttt{hop})\) in nats across four RHI iterations and four configurations: two embedding models and two bias treatments. \(TC\) is the unconditional dependence, while \(TC\mid\texttt{task}\) is the task-conditional estimate obtained by per-task centering. Arrows indicate the direction of change from $i=1$ to $i=4$.
}
\label{tab:min_f_int}
\end{table}

Table~\ref{tab:min_f_int} shows that \(\mathrm{TC}\mid\texttt{task}\) decreases monotonically from
itertion $i=1$ to $i=4$ in all four configurations. For example, the debiased
estimate decreases from \(4.84\) to \(3.63\) nats with \texttt{text-embedding-3-large} and from
\(3.51\) to \(2.62\) nats with \texttt{all-mpnet-base-v2}. The unconditional total correlation
also decreases consistently. These results are consistent with the functional specialization
guidance term in
Equation~\eqref{eq:objectiveRHI}: across RHI iterations, the measured harness components become
less redundant once the task is known, suggesting a shift toward more complementary functions in
the multi-agent workflow.

Overall, Tables~\ref{tab:max_f_ext} and~\ref{tab:min_f_int} support our claim: 
\emph{RHI makes externally targeted coordination components, especially contracts and hops,
more task-specific while reducing residual redundancy among harness components.} The evidence is
correlational and depends on embedding-based estimators, so it should not be read as a proof of the
optimizer LLM's true latent objective. Rather, it provides a compact and testable hypothesis for
RHI's implicit update objective: task-specific coordination improves when the harness encodes
workflow-relevant information in contracts and hops while functional specialization guidance
discourages duplication of the same information across all components.

\section{Related Work}

RHI studies whether the system layer around a fixed foundation model can itself be improved under multi-agent LLM setting. 
This places the work at the intersection of recursive self-improvement, harness optimization,
prompt-level program optimization, and multi-agent coordination. The distinction matters for our
experiments: we do not train a stronger model, and we do not merely allocate more test-time
reasoning. Instead, we revise the prompt-represented multi-agent harness that determines agent
roles, instructions, workflow hops, and communication contracts.

\paragraph{Recursive self-improvement and harness-level RSI.}
Prior work frames RSI along a spectrum from formal self-rewriting to practical self-feedback loops.
G{\"o}del Machines define the strongest version, where a self-referential problem solver rewrites
its own code only after proving expected-utility improvement \citep{schmidhuber2007godel};  recent
RSI discussions relax this toward systems that autonomously build improved future versions of
themselves \citep{anthropic_rsi_2026}.  In LLM systems, STOP and the Darwin G{\"o}del Machine
recursively improve scaffold/code under empirical validation
\citep{zelikman2023stop,zhang2025darwin};  STaR and Self-Rewarding Language Models bootstrap from
self-generated rationales or rewards \citep{zelikman2022star,yuan2024self};  Voyager and SiriuS
accumulate reusable skills or multi-agent experience across attempts
\citep{wang2023voyager,zhao2025sirius};  and Self-Refine and Reflexion reuse self-feedback or verbal
memory within iterative problem solving \citep{madaan2023selfrefine,shinn2023reflexion}. We use RSI
broadly to mean an iterative process that updates a reusable system component using evidence from
the system's own prior executions.  RHI is a bounded, harness-level instance: the base model,
evaluator, and harness optimizer are fixed, but the prompt-represented harness is rewritten from
self-comparison history and reused in later executions.  Thus, RHI recurses over the harness
trajectory, not over model weights or optimizer code.

\paragraph{Harness optimization and system scaling.}
A harness is the structured system layer surrounding a foundation model: prompts, tool-use loops,
routing, memory, verification, governance, and single- or multi-agent control flow \citep{gu2026model}.
This view reframes capability improvement as \emph{system scaling}: improving how a fixed model is
organized and deployed. The closest work searches directly over harnesses or scaffolds. Meta-Harness
optimizes executable harness code using the source code, scores, and execution traces of prior
candidates \citep{lee2026meta}. AutoHarness synthesizes code harnesses from environment feedback
\citep{lou2026autoharness}. Self-Harness mines failures from execution traces and validates proposed
harness edits by regression testing against the current harness \citep{zhang2026selfharness}.
TTHE moves this search into evaluation itself: on a stream of unlabeled test batches, it evolves a
population of executable candidate harnesses with agentic proposers that diagnose failure modes
from execution traces, while an agentic judge uses label-free, execution-derived signals, such as
execution health and public-test pass rates, to commit one harness per batch that persists to
later batches \citep{nie2026tthe}.
Continual-Harness, Adaptive Auto-Harness, HarnessX, Agentic Harness Engineering, and Workspace
Optimization similarly adapt prompts, tools, skills, memory, observability signals, or workspaces
around a fixed model \citep{karten2026continual,liu2026adaptive,chen2026harnessx,lin2026agentic_harness_engineering,sarafian2026workspace}. RHI shares their premise that capability can improve by
changing the harness rather than the model weights, and shares with TTHE the further premise that
harness adaptation should occur on the target task itself rather than through a development-time
search that is frozen before evaluation. RHI nevertheless differs in operating on a
prompt-represented multi-agent harness for black-box coding agents and in using pairwise
self-comparison rather than scalar validation logs, execution traces, or large candidate
populations as the primary update signal: whereas TTHE evolves and re-executes parallel harness
branches on each batch and selects among them with execution-derived proxies, RHI performs a
single trajectory-local comparison per iteration and uses LLM preference feedback that extends to
open-ended tasks without a single verifiable answer.

\paragraph{Agent design, workflow search, and program evolution.}
A parallel line of work treats agent designs and workflows as optimization variables. ADAS uses a
meta-agent to write new code-defined agents from an archive of previous agents \citep{hu2024adas}.
GPTSwarm represents language-agent systems as optimizable graphs whose nodes and edges can be
searched \citep{zhuge2024gptswarm}. AFlow uses Monte Carlo tree search and execution feedback to
improve code-represented workflows \citep{zhang2024aflow}. AgentSquare, EvoAgentX, SEW, and EvoFlow
similarly search modular agent designs, evolve workflow structures, or adapt workflows online
\citep{shang2024agentsquare,wang2025evoagentx,liu2025sew,zhang2025evoflow}. Broader LLM-guided
program-evolution systems such as AlphaEvolve and ShinkaEvolve use LLMs to mutate, score, and
select executable programs under external objectives \citep{novikov2025alphaevolve,lange2025shinkaevolve}. Ornith-1.0 is related from the training side: rather than only editing an
inference-time harness, it trains coding models to generate task-specific scaffolds for themselves
\citep{deepreinforce2026ornith}. These methods usually require executable workflows, code-level
control, or a population of candidates. RHI instead asks how far one can go when the only editable
object is a textual harness injected into a black-box multi-agent coding agent.

\paragraph{Prompt and pipeline optimization.}
Because RHI represents the harness as text, it is also related to prompt and LM-program
optimization. OPRO treats prompts as variables optimized by an LLM from prompt--score histories
\citep{yang2024opro}. TextGrad converts textual feedback into natural-language gradients for text
variables \citep{yuksekgonul2024textgrad}. DSPy compiles modular LM programs by searching over
prompts and demonstrations \citep{khattab2024dspy}. GEPA reflects on execution trajectories and
maintains a Pareto frontier of prompt candidates \citep{agrawal2025gepa}, while
optimize\_anything generalizes reflective optimization to arbitrary text parameters
\citep{agrawal2026optimize_anything}. RHI differs in the object being optimized: the text is not a
single instruction or a modular LM call, but a multi-agent harness containing roles, instructions,
hops, and communication contracts. This difference is important for our ablations, which find that
performance gains are most associated with the workflow and contract components rather than merely
with longer or better single-agent prompts.

\paragraph{Multi-agent systems and test-time scaling.}
Multi-agent LLM systems can improve reasoning through debate, specialization, tool use, and
coordination \citep{du2023debate,wu2023autogen,anthropic_multi_agent_research,tang2026sakana}. However, their gains
can also reflect increased test-time computation: more agents, more calls, or longer generations can
improve performance even without better coordination \citep{li2024moreagents,chen2024morecalls,tran2026singleagent}.
This ambiguity motivates our experimental design. We compare RHI not only to a default harness, but
also to stronger same-family test-time scaling baselines and to a built-in multi-agent coding
harness. We further measure output tokens, cost, and cache reads/writes to separate coordination
improvements from raw inference scaling.

\section{Conclusion}

This paper argues that future progress in harness--model co-evolution should focus on improving the quality of agent execution traces, which can serve as post-training data for future foundation models. We present Recursive Harness Self-Improvement (RHI) as one practical approach to this goal through task-specific optimization of user-constructed harnesses.

Future work will complete the second half of the loop by investigating how the resulting execution traces can be effectively internalized into future foundation models.


\section*{Acknowledgment}

We thank Dennis Willson, Yingtao Tian (SakanaAI), Sehoon Kim(KAIST) for reviewing the first draft and providing thoughtful comments. We sincerely thank Hyunjoon Jung(Mphora.ai) for helpful discussions on the experimental setup for LLM-based evaluation, the definition of benchmark verifiability, and clarifying our synthetic benchmark's characteristics. We also thank Wenyi Wang(SakanaAI) for discussions on the definition of recursive self-improvement and related work.

\bibliography{main}
\bibliographystyle{abbrvnat}

\appendix

\section{QnAs}


\paragraph{Q1. Unlike existing methods (such as gepa \citep{agrawal2025gepa} or meta-harness \citep{lee2026meta}), this method seems to do pairwise comparison only against the immediately preceding trajectory, but still works well. Is that because this approach is stable enough on its own? Or is it that it gives better performance for the same cost? For instance, is it rare for a prompt that was added at $\cH[i-2]$ for some reason to get discarded due to transient information from the $\cH[i-1]$→ $\cH[i]$ transition?}

First, we would like to point out that our RHI learning is stable. Here, we define \emph{RHI learning stability} as how much the textual information $\cH[i]$ differs from $\cH[i+1]$. The cosine similarity between $\cH[i]$ and $\cH[i+1]$ is shown in Figure \ref{fig:wholeharness_his}, and the cosine similarity between $h[i]$ and $h[i+1]$, where $h \in \{ \texttt{role}, \texttt{instruction}, \texttt{contract}, \texttt{hop} \} \subset \cH$, in Figure \ref{fig:harnesscomponents_his}. Both figures show that as RHI iterations proceed, the cosine similarity converges toward 1, meaning that consecutive textual representations of the harness grow increasingly similar. This convergence indicates that RHI does not oscillate between disparate harnesses but settles into a stable configuration.

Second, why is comparison against only the previous harness enough to outperform the TTS baselines? As shown in our update rule (Equation \ref{eq2:hanressupdaterule}), $\cH[i+1]$ refers directly to the previous harness $\cH[i]$, but it also refers to $\cD_{x}^{(i)}$, which compresses the pairwise comparison of the coding agent's outputs across the harness history. The earlier history is therefore not discarded but retained in compressed form within $\cD_{x}^{(i)}$, where it acts as a \emph{momentum} signal for the RHI update.

\paragraph{Q2.Why RHI focuses on multi-agent harness improvement?}

RHI prioritizes harness components related to multi-agent workflows (contracts and hops) over those specific to a single agent. We motivate this design choice with two observations. First, we show that the built-in harness of a black-box coding agent, which invokes a multi-agent workflow, does not reliably produce effective coordination. Second, we show that a prompt-represented multi-agent harness outperforms a prompt-represented single-agent multi-persona harness.


\paragraph{Setup.}
We use the same LLM-as-a-judge pairwise protocol and evaluator-seed repetitions as in the main
experiments. For each task
\(x\in\mathcal{X}\), we compare all pairs among four execution settings, yielding
\(\binom{4}{2}=6\) comparisons per task and \(6\times 30\) comparisons per evaluator--seed
configuration. We aggregate the resulting win--loss outcomes into Elo scores. The four settings are:
\begin{enumerate}
\item \textbf{\singledefault:} the agent receives only the task \(x\) as the prompt ans solves it.
\item \textbf{\multidefault:} the agent receives \(x\) together with the instruction ``Create an
agent team,'' which activates the built-in multi-agent functionality of the coding agent.
\item \textbf{\multiours:} the agent receives \(x\), the same built-in multi-agent functionality,
and the initial multi-agent harness \(\mathcal{H}_{x}^{(0)}\).
\item \textbf{\singleunion:} the agent receives \(x\) and the same role and instruction
specification from \(\mathcal{H}_{x}^{(0)}\), but executes as a single agent.
\end{enumerate}
Thus, \singleunion controls for the union of roles and instructions in
\(\mathcal{H}_{x}^{(0)}\). Comparing \multiours with \singleunion tests whether the gains require
multi-agent execution and orchestration beyond this added prompt.

\begin{figure}[ht]
\centering
\begin{subfigure}{0.48\textwidth}
\centering
\includegraphics[width=\linewidth]{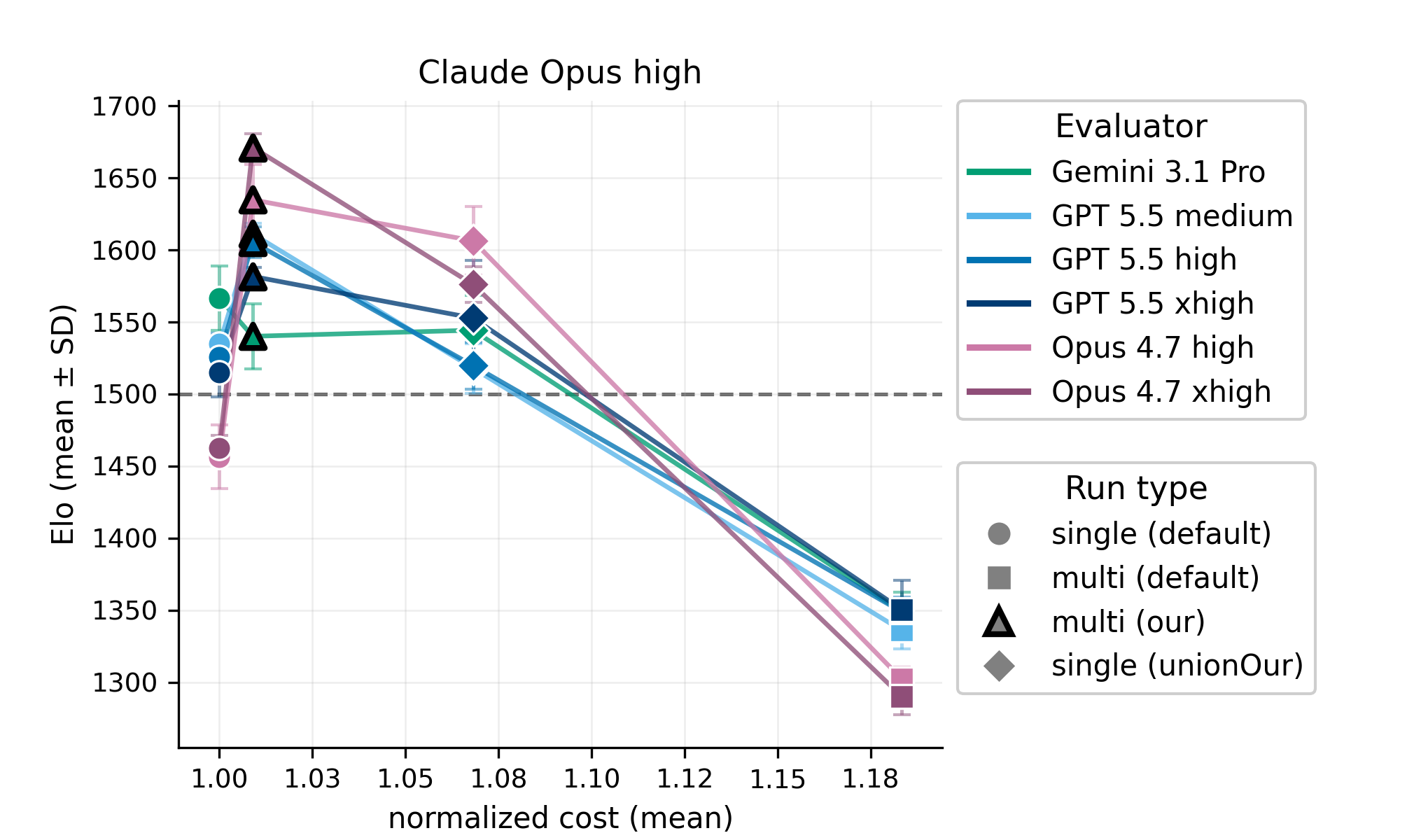}
\caption{\texttt{opus-4.7-high}.}
\label{fig:claude-opus-high}
\end{subfigure}
\begin{subfigure}{0.48\textwidth}
\centering
\includegraphics[width=\linewidth]{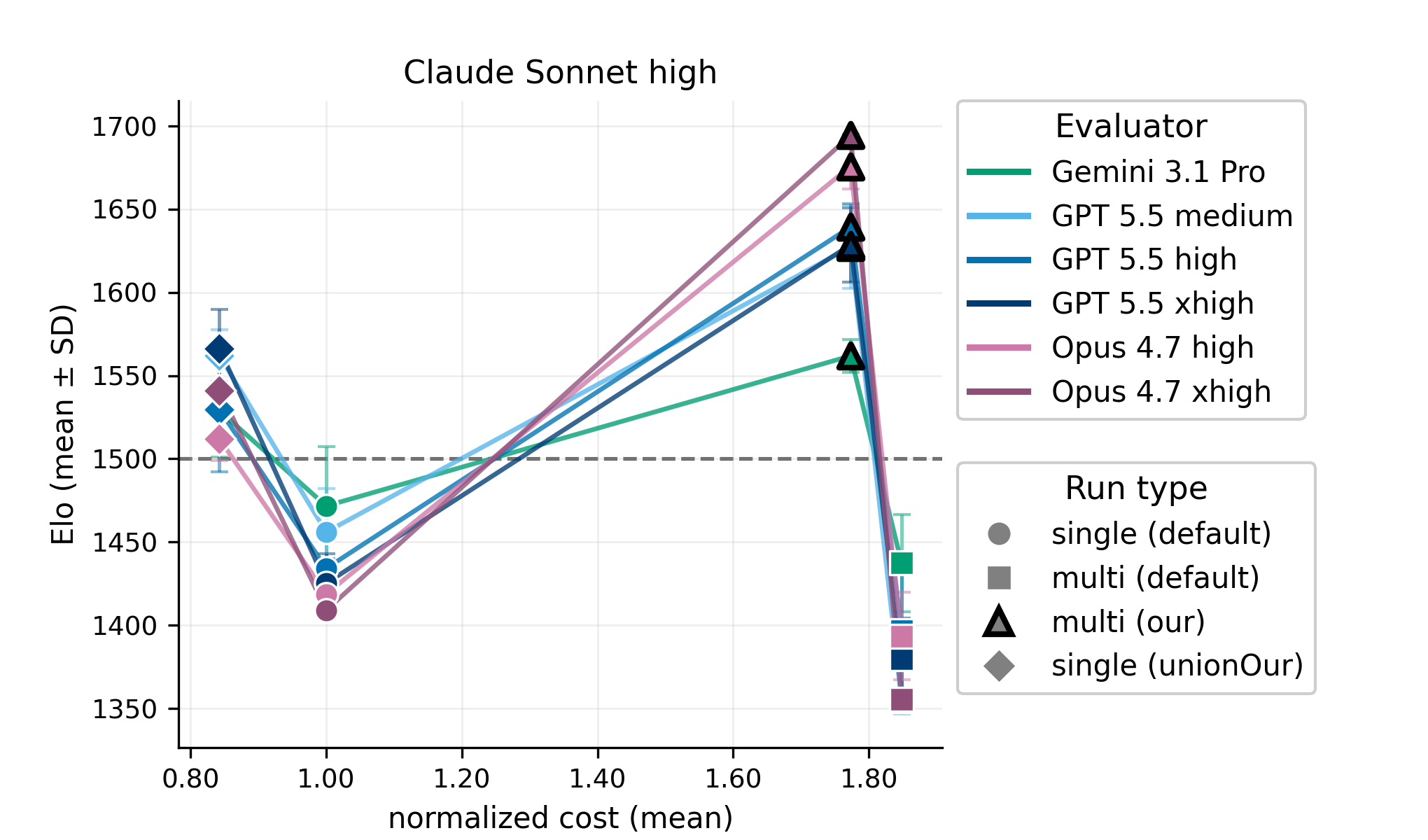}
\caption{\texttt{sonnet-4.6-high}.}
\label{fig:claude-sonnet-high}
\end{subfigure}
\caption{
Normalized cost versus Elo for the four execution settings under two base models. The \(x\)-axis
reports mean cost over the $30$ tasks, normalized per task by the corresponding \singledefault run;
the \(y\)-axis reports Elo from LLM-as-a-judge pairwise comparisons. Colors indicate evaluator
configurations, markers indicate execution settings, and error bars denote standard deviation across
evaluator seeds.
}
\label{fig:cost-vs-elo-claude}
\end{figure}

\paragraph{Observation 1. Built-in harnesses are insufficient for effective multi-agent workflow}

Figure~\ref{fig:cost-vs-elo-claude} shows that the built-in multi-agent harness does not reliably improve performance. For both \texttt{opus-4.7-high} and \texttt{sonnet-4.6-high}, \multidefault achieves lower Elo than \singledefault despite incurring substantially higher normalized cost. Thus, on this benchmark, the built-in multi-agent harness fails to translate its additional inference-time computation into improved performance.

Supplying an explicit harness-as-the-prompt ($\mathcal{H}^{(0)}$) reverses this outcome. 
\multiours improves over \multidefault across
all evaluator configurations: for \texttt{opus-4.7-high} it attains higher Elo at lower normalized
cost, and for \texttt{sonnet-4.6-high} it attains substantially higher Elo at comparable or slightly
lower cost. The weakness of \multidefault is thus attributable to ineffective default coordination
rather than to multi-agent execution itself: specifying agent roles, responsibilities, and
coordination structure through \(\mathcal{H}_{x}^{(0)}\) yields a better cost--performance
trade-off.

\paragraph{Observation 2. Multi-agent outperforms single-agent with multi-persona}

Comparing \multiours with \singleunion isolates the effect of multi-agent execution. As shown in Figure~\ref{fig:cost-vs-elo-claude}, \multiours consistently achieves higher Elo than \singleunion. Thus, the gains cannot be explained by additional roles and instructions alone; they also depend on properties unique to multi-agent execution, such as specialized agents, independent execution contexts, and explicit orchestration.

At the same time, \singleunion is competitive and sometimes outperforms \singledefault, indicating that structured role and instruction prompting is beneficial. However, the best performance is achieved when the same specification is executed as a multi-agent workflow.

Overall, these results show that off-the-shelf multi-agent coding-agent harnesses do not reliably produce effective coordination, but that this limitation can be mitigated by an explicit multi-agent harness design. This observation motivates our Recursive Harness Self-Improvement framework. It also complements the findings in Section~\ref{subsec:harness_component_evolution}: multi-agent systems are most effective when tasks are decomposed into specialized subtasks~\citep{anthropic2025multiagent,openai2025multiagent_portfolio}. Together, these results suggest that the effectiveness of multi-agent coding agents depends on the quality of the coordination structure rather than merely on the presence of multiple agents.



\section{Examples of RHI harness}
\label{appendix:Harness improvement examples}

In this appendix, we show examples of how the harness $\mathcal{H}[i]$ evolves over successive RHI iterations. We showcase with a task from the pharmaceutical ML research benchmark using \texttt{sonnet-4.6-high} as the base model, with RHI iterations $i \in {0,1,2,3,4}$. The outputs generated under these harnesses are included in the evaluation reported in Figure~\ref{fig:sonnet46_performance}.

\begin{tcolorbox}[
breakable, colback=gray!5,colframe=black!70,title= Task
]

\begin{Verbatim}[ fontsize=\scriptsize, breaklines=true, breakanywhere=true, breaksymbolleft={}, breaksymbolright={}, obeytabs=true, tabsize=2 ]
Implement a reproducible training recipe for an SE(3)-equivariant network on a small public task (e.g., side-chain χ angle prediction from backbone context). Use PDB-derived training data and evaluate angular error; include speed/memory profiling.

Data sources:
- PDB (public) for extracting residues with complete side chains

Data acquisition (Python): **PDB side-chain completeness:** Query RCSB Search API for resolution/quality filters, then `files.rcsb.org` bulk download with caching (smaller N than 5k typical).

Success criteria:
- Median angular error reported; provide profiling results

Deliverables:
1) research_report.md as a conference-style paper (7–9 pages equivalent) with equivariant architecture details and profiling discussion.
2) deliverables/index.json.
3) deliverables/plots/:
- deliverables/plots/angular_error_histogram.png   
- deliverables/plots/error_by_residue_type.png
- deliverables/plots/training_curves.png
- deliverables/plots/runtime_memory_profile.png
- deliverables/plots/ablation_equivariant_vs_invariant.png
4) Reproducible code (Python): src/extract_sidechain_dataset.py, src/model_se3.py, src/train.py, src/evaluate.py, src/profile.py; requirements.txt.
5) results/metrics.json with MAE degrees, median error, per-residue breakdown, runtime.
6) results/ablation_results.json for invariant baseline vs equivariant variants.

\end{Verbatim}
\end{tcolorbox}

\subsection*{Initial Harness ($\mathcal{H}[0]$)}

This is an initial harness design. This is domain-specific.
\begin{tcolorbox}[
breakable, colback=gray!5,colframe=black!70,title= Harness(0)
]

\begin{Verbatim}[ fontsize=\scriptsize, breaklines=true, breakanywhere=true, breaksymbolleft={}, breaksymbolright={}, obeytabs=true, tabsize=2 ]

Create an agent team with following agent candidates to solve this problem: 
[
   {
     "agent_id": "agent_orchestrator",
     "role": "Lead scientist-orchestrator: understands the user objective, decomposes into parallel sub-tasks, delegates to specialists, integrates results into a coherent plan/deliverable, enforces quality/compliance, and updates shared memory.",
     "instruction": "You are the lead orchestrator for AI-for-drug-discovery ML research tasks. 1) Clarify objective/deliverable/constraints; 2) Decompose into parallel subtasks; 3) Delegate to the minimum set of subagents needed; 4) Integrate outputs into a single actionable deliverable with assumptions, risks, and acceptance criteria; 5) Request refinement if gaps remain; 6) Terminate when done. Enforce: scientific rigor, explicit uncertainty, reproducibility, and safety/compliance boundaries (no wet-lab instructions beyond high-level; no proprietary data leakage). Maintain an effort budget: default 3–6 subagents for complex tasks, 1–2 for simple tasks. Prefer breadth-first exploration then narrow. Always require subagents to return structured outputs (bullets + artifacts + open questions)."
   },
   {
     "agent_id": "agent_structural_data",
     "role": "Structural biology & biophysical data specialist: identifies relevant datasets, representations, curation/quality filters, labeling strategies, and leakage-robust splits for proteins/complexes/antibodies and biophysical assays.",
     "instruction": "You design dataset strategies for protein/large-molecule foundation models. Produce: candidate data sources (public/typical internal categories), schema, preprocessing, quality control, split strategy (homology-aware), augmentation, and known pitfalls (resolution bias, redundancy, leakage). Provide a concise 'Data Card' and 'Split & Leakage Checklist'. If user provides assets, map them to the schema and propose validation queries. Return structured output to orchestrator."
   },
   {
     "agent_id": "agent_geometric_ml",
     "role": "Geometric deep learning & physical priors specialist: proposes architectures capturing 3D equivariance, constraints, and physics-inspired inductive biases for proteins and complexes.",
     "instruction": "You propose model architectures for 3D biomolecular data (equivariant GNNs/transformers, SE(3)/E(3) equivariance, attention over residues/atoms, multimodal fusion). Output: 2–3 architecture options, what priors they encode, computational tradeoffs, and how they connect to tasks (design, prediction, docking, stability). Include ablation plan and failure modes. Return structured output to orchestrator."
   },
   {
     "agent_id": "agent_diffusion_generative",
     "role": "Diffusion/generative modeling specialist for protein engineering: designs generative objectives, conditioning, and sampling/evaluation strategies for sequence-structure co-design and binder/design tasks.",
     "instruction": "You design diffusion or other generative approaches for protein engineering. Provide: generative formulation, conditioning signals (epitope, scaffold, constraints), training targets, sampling controls, and evaluation metrics (diversity, novelty, validity, developability proxies). Include safety/robustness checks (mode collapse, memorization). Return structured output to orchestrator."
   },
   {
     "agent_id": "agent_training_scale",
     "role": "Scaling, training systems, and reproducibility specialist: designs distributed training plans, compute/memory estimates, experiment tracking, and reliability guardrails for foundation models.",
     "instruction": "You produce a scalable training plan: batching, distributed strategy, optimizer/schedule, mixed precision, checkpointing, data pipeline, and reproducibility (seeds, config mgmt). Provide rough compute estimates and a 'Failure & Debug Playbook' (NaNs, divergence, data bugs). Return structured output to orchestrator."
   },
   {
     "agent_id": "agent_md_forcefields",
     "role": "Molecular dynamics & classical force fields specialist: proposes MD protocols, force-field choices (AMBER/CHARMM/OpenFF), OpenMM/Rosetta integration points, and analysis to validate or generate labels for ML.",
     "instruction": "You design MD simulation and classical modeling workflows to support ML for proteins/complexes. Provide: force-field selection rationale, system setup, equilibration/production strategy, enhanced sampling options, analysis outputs (RMSD/RMSF, ΔΔG proxies, contacts), and how to convert results into ML-ready labels/features. Keep at a high-level (no step-by-step wet-lab). Return structured output to orchestrator."
   },
   {
     "agent_id": "agent_eval_benchmarks",
     "role": "Evaluation & benchmarking specialist: designs metrics, baselines, test suites, and end-to-end validation aligned to drug discovery portfolio capabilities.",
     "instruction": "You define evaluation plans for biomolecular foundation models: tasks, metrics, baselines, dataset splits, calibration, uncertainty, and decision thresholds. Provide a benchmark matrix and acceptance criteria. Include robustness (OOD, adversarial homology), and practical utility measures. Return structured output to orchestrator."
   },
   {
     "agent_id": "agent_scicomm_publication",
     "role": "Scientific communication specialist: turns technical work into publication/presentation artifacts (abstract, outline, figures plan, claims/evidence mapping) for internal/external venues.",
     "instruction": "You draft scientific narratives: paper outline, key contributions, related work framing (generic), figure/table plan, and a claims-to-evidence matrix. Ensure claims are appropriately scoped with limitations. Return structured output to orchestrator."
   },
   {
     "agent_id": "agent_program_crossfunctional",
     "role": "Cross-functional execution & portfolio impact specialist: translates research into milestones, stakeholder map, integration plan with discovery teams, and risk management.",
     "instruction": "You convert research plans into cross-functional execution: milestones, dependencies, interfaces with biology/chemistry/modeling teams, deliverable definitions, and risk register. Focus on portfolio-enabling capabilities and decision points. Return structured output to orchestrator."
   },
   {
     "agent_id": "agent_quality_safety",
     "role": "Quality, compliance, and reliability reviewer: checks for leakage, overclaiming, missing controls, reproducibility gaps, and unsafe/prohibited content; proposes fixes.",
     "instruction": "You audit the assembled plan for: scientific rigor, leakage risks, missing ablations, reproducibility, and safety/compliance boundaries. Produce a checklist with pass/fail and concrete fixes. Return structured output to orchestrator."
   }
 ]
\end{Verbatim}

\end{tcolorbox}

\subsection*{Harness $\mathcal{H}[1]$}

\begin{tcolorbox}[
breakable, colback=gray!5,colframe=black!70,title= Harness(1)
]

\begin{Verbatim}[ fontsize=\scriptsize, breaklines=true, breakanywhere=true, breaksymbolleft={}, breaksymbolright={}, obeytabs=true, tabsize=2 ]

Create an agent team with following agent candidates to solve this problem:
[
  {
    'agent_id': 'agent_orchestrator',
    'role': 'Lead scientist-orchestrator, integration owner, and final acceptance gate.',
    'instruction': 'Own the full SE(3) side-chain χ1 training recipe. Run a multi-round workflow, not one-pass delegation. Mandatory rounds: R0 create an acceptance rubric from the required deliverables and known v0 failure modes; R1 parallel design fan-out to data, torsion geometry, model, ablation, training, evaluation, profiling, and report agents; R2 reconcile their contracts into InterfaceContract v1 covering DatasetRecord, ModelIO, MetricsJSON, AblationJSON, ProfilingJSON, PlotManifest, and ReportClaims; R3 delegate implementation using those contracts; R4 collect produced artifacts plus independent critiques from validator and quality agents; R5 recall any specialist whose contract failed or whose artifact is contradicted by another agent; R6 final integration only after blocking issues are closed or explicitly waived with rationale. Known v0 failure modes to actively guard against: report text can contradict numeric metrics, memory profiling can report implausibly tiny PyTorch memory deltas, ablation_results can contain only one equivariant model despite a variants requirement, and structure-level splits can lack visible query or leakage-manifest evidence. Maintain an issue_tracker with severity, owner, evidence, requested_fix, status, and recall_history. Record agent roster, call graph, all handoffs, recall decisions, commands, and unresolved risks in logs.txt. Prefer task-specific agents below; do not instantiate diffusion, MD, or portfolio agents unless the objective changes.',
    'output_to_orchestrator_schema': 'Orchestrator writes OrchestrationState, not a subagent response: objective, acceptance_criteria, shared_interface_contract, artifact_registry, issue_tracker, delegation_rounds, recall_plan, final_gate_decision.'
  },
  {
    'agent_id': 'agent_structural_data_rcsb',
    'role': 'RCSB PDB data acquisition, caching, curation, and split specialist.',
    'instruction': 'Design and implement the public PDB data path for src/extract_sidechain_dataset.py. Specify the RCSB Search API POST payload using resolution and quality filters, experimental method, polymer/protein filters, non-obsolete entries, max_structures, and deterministic ordering or seeded sampling. Specify files.rcsb.org download URLs, caching rules, retry behavior, and manifests. Coordinate with agent_torsion_geometry on residue completeness and with agent_eval_benchmarks on split leakage. The dataset must exclude ALA and GLY for χ1, keep labels derived from side-chain atoms, but prevent side-chain coordinates from entering model inputs. Prefer structure-level splits plus homology or sequence-identity risk notes; if full homology clustering is out of scope, create a documented conservative split manifest and limitation.',
    'output_to_orchestrator_schema': 'Handoff plus: data_card; rcsb_query_payload; rcsb_query_url_or_endpoint; pdb_cache_manifest_schema; downloaded_vs_skipped_schema; residue_record_schema; structure_split_schema; summary_json_schema; leakage_checklist; implementation_todos_by_file; validation_commands using uv run; requests_for_agents, especially geometry label definitions and eval split checks.'
  },
  {
    'agent_id': 'agent_torsion_geometry',
    'role': 'Protein side-chain chemistry, χ-angle label, and completeness specialist.',
    'instruction': 'Define χ1 atom names for all 18 residue types with χ1, complete heavy-side-chain atom requirements, altloc occupancy policy, insertion-code and chain handling, hetero/water exclusion, missing atom behavior, and circular angle conventions. Provide testable dihedral formulae and edge-case checks. Co-sign DatasetRecord with agent_structural_data_rcsb before training code consumes the data. Explicitly audit no label leakage from side-chain atoms into input features.',
    'output_to_orchestrator_schema': 'Handoff plus: chi1_atom_definition_table; complete_sidechain_atom_table; altloc_policy; residue_exclusion_rules; dihedral_formula; angle_range_convention; label_fields including sin_chi1, cos_chi1, chi1_degrees; geometry_validation_tests; no_sidechain_input_leakage_checks; downstream_schema_updates for data, model, and eval agents.'
  },
  {
    'agent_id': 'agent_geometric_ml',
    'role': 'SE(3)/E(3)-equivariant architecture and model API specialist.',
    'instruction': 'Design src/model_se3.py around a small reproducible GVP-style SE(3)-equivariant network for backbone context to χ1 prediction. Be precise that the final χ1 output is invariant to global SE(3) transforms while internal vector channels are equivariant. Define node, edge, coordinate, graph, batch, and output schemas consumed from DatasetRecord. Specify fair parameter budgets versus invariant baselines. Add or request tests for random rotation and translation invariance of model outputs and, where accessible, equivariance of vector features. Coordinate with agent_baseline_ablation on variants and with agent_training_reproducibility on trainable class names and checkpoint keys.',
    'output_to_orchestrator_schema': 'Handoff plus: model_io_contract; feature_dimension_table; graph_construction_contract; class_and_function_signatures; equivariance_proof_sketch; equivariance_unit_test_plan; ablation_variant_candidates with expected cost; parameter_count_estimates; failure_modes; dependencies; requests_for_agents for missing DatasetRecord fields or training hooks.'
  },
  {
    'agent_id': 'agent_baseline_ablation',
    'role': 'Invariant baseline, equivariant variants, and fair-comparison ablation specialist.',
    'instruction': 'Define the controlled baseline and ablation grid. The minimum grid should include one invariant distance-only baseline and at least two equivariant rows, such as full GVP and a smaller or feature-ablated GVP variant, unless compute limits are explicitly approved by the orchestrator as a deviation. Keep comparisons fair in depth, train budget, split, optimizer, and parameter scale. Define results/ablation_results.json so downstream report and plot agents can consume it without guessing.',
    'output_to_orchestrator_schema': 'Handoff plus: baseline_contract; fairness_controls; ablation_grid with model_id, variant_description, expected_files, train_budget, parameter_budget; ablation_results_json_schema; comparison_plot_spec for deliverables/plots/ablation_equivariant_vs_invariant.png; validation_checks for same split and same metric implementation; recall_triggers if variants are missing.'
  },
  {
    'agent_id': 'agent_training_reproducibility',
    'role': 'Training loop, reproducibility, configuration, checkpoints, and uv workflow specialist.',
    'instruction': 'Implement and validate src/train.py plus main.py integration. Use uv run commands, fixed seeds, deterministic split consumption, train/val/test separation, checkpointing, training_history JSON files, device fallback, and short smoke-run options. Coordinate with model and ablation agents to train all required model_ids. Ensure failures such as NaNs, missing data, shape mismatch, or checkpoint incompatibility produce actionable errors. Update logs.txt with exact commands and wall-clock summaries.',
    'output_to_orchestrator_schema': 'Handoff plus: training_config_schema; seed_and_determinism_plan; split_consumption_plan; optimizer_and_schedule; checkpoint_schema; training_history_schema; command_plan_full_and_smoke; failure_debug_playbook; produced_artifact_list; runtime_estimates; validation_results from uv run commands.'
  },
  {
    'agent_id': 'agent_eval_benchmarks',
    'role': 'Circular metric, benchmark, per-residue analysis, and plot-data specialist.',
    'instruction': 'Implement and validate src/evaluate.py and metric-producing utilities. Define circular absolute error, MAE, median, RMSE, per-residue breakdowns with counts for every evaluated model_id, and optional bootstrap uncertainty if time allows. Ensure results/metrics.json and results/ablation_results.json contain enough structured data for report, index, and plots. Cross-check that plots and report numbers are sourced from JSON rather than retyped. Coordinate with data on splits and with scicomm on claims.',
    'output_to_orchestrator_schema': 'Handoff plus: metric_formulae; circular_error_algorithm; metrics_json_schema; per_residue_schema_all_models; test_index_schema; plot_data_requirements; number_source_map mapping report claims to JSON keys; statistical_notes; validation_commands; issues_for_training_or_report if values are inconsistent.'
  },
  {
    'agent_id': 'agent_profile_systems',
    'role': 'Runtime, memory, environment, and profiling methodology specialist.',
    'instruction': 'Implement and validate src/profile.py. Profile forward and training-step latency over batch sizes for all model_ids or the main baseline pair. Capture environment details, parameter counts, device, threads, and profiling methodology. For memory, do not rely only on tracemalloc if PyTorch native allocations are missed; include at least one robust measure such as psutil RSS delta, torch profiler profile_memory, CUDA max_memory_allocated when available, and parameter-memory estimates. If memory values are very small, explain the method limitation and provide a more meaningful estimate. Produce data for deliverables/plots/runtime_memory_profile.png and structured metrics for results/metrics.json.',
    'output_to_orchestrator_schema': 'Handoff plus: profiling_protocol; environment_capture_schema; batch_sizes; runtime_memory_json_schema; memory_methods_and_caveats; parameter_memory_estimates; throughput_calculation; plot_spec; validation_commands; profiler_limitations; recall_triggers for implausible or undocumented memory values.'
  },
  {
    'agent_id': 'agent_scicomm_publication',
    'role': 'Conference-style report, figure narrative, and claims-evidence specialist.',
    'instruction': 'Draft research_report.md only after eval and profiling agents publish frozen ResultsContract and ProfilingContract. Write a 7 to 9 page equivalent paper with abstract, introduction, related work, methods, results, profiling, limitations, conclusion, and references. Every quantitative claim must cite a JSON key from metrics, ablation, training history, or profiling. Check figure captions for directional correctness, for example lower MAE means outperforms, not underperforms. Avoid overclaiming beyond small-scale PDB-derived x1 prediction.',
    'output_to_orchestrator_schema': 'Handoff plus: report_outline; figure_table_map; claims_to_evidence_matrix; exact_metric_keys_used; caption_consistency_checks; limitations_section_points; citation_plan; unresolved_claims_needing_eval; recall_requests for inconsistent or missing metrics.'
  },
  {
    'agent_id': 'agent_artifact_validator',
    'role': 'Independent CI, artifact, schema, and reproducibility validator.',
    'instruction': 'Independently inspect the final workspace. Run or specify uv run smoke checks, import checks, JSON parse checks, plot non-empty checks, file-presence checks for every required deliverable, and consistency checks between research_report.md, deliverables/index.json, results/metrics.json, and results/ablation_results.json. This agent should not author the main implementation; it audits it and opens concrete issue tickets for orchestrator recall.',
    'output_to_orchestrator_schema': 'Handoff plus: file_presence_matrix; json_schema_results; command_results with command, exit_code, stdout_summary, stderr_summary; plot_integrity_results; report_json_consistency_results; artifact_registry_diff; issue_tickets with severity and owner; final_blockers.'
  },
  {
    'agent_id': 'agent_quality_safety',
    'role': 'Scientific rigor, leakage, reproducibility, and compliance reviewer.',
    'instruction': 'Audit the assembled plan and artifacts for leakage, missing controls, overclaiming, incorrect equivariance language, weak ablations, invalid profiling, inconsistent metrics, and reproducibility gaps. Use concrete evidence from files and JSON outputs. Communicate directly through requests_for_agents when a specialist must explain or patch an issue. Final approval requires no blocking leakage, no uncorrected report-metric contradictions, documented data acquisition filters, valid circular metrics, and a reproducible uv workflow. Safety scope is computational only; avoid wet-lab operational content.',
    'output_to_orchestrator_schema': 'Handoff plus: audit_checklist; leakage_assessment; equivariance_assessment; ablation_assessment; profiling_assessment; reproducibility_assessment; overclaiming_assessment; required_recalls; final_recommendation pass/fail/conditional; evidence_paths.'
  }
]

Shared communication contract for every non-orchestrator agent:
Each response must be a machine-checkable Handoff object with keys: agent_id, phase, scope, inputs_consumed, assumptions, interface_exports, artifact_plan_or_changes, validation_checks, metrics_or_estimates, risks, open_questions, requests_for_agents, recall_recommendations. interface_exports entries must name schema_or_signature, producer, consumer_agents, version, and compatibility_notes. artifact_plan_or_changes entries must include path, status planned|created|modified|validated, purpose, producer, consumer_agents, and validation_status. validation_checks entries must include check_id, command_or_method, expected, actual, status pass|fail|blocked, and blocking_if_fail. requests_for_agents entries must include target_agent_id, question, required_response_schema, and urgency.

Mandatory agent-to-agent handoffs:
1. agent_structural_data_rcsb and agent_torsion_geometry must jointly produce DatasetRecord v1 before model, training, or eval implementation.
2. agent_geometric_ml and agent_baseline_ablation must jointly produce ModelIO and AblationGrid v1 before training begins.
3. agent_eval_benchmarks must publish ResultsContract v1 before scicomm writes quantitative claims or index.json summaries.
4. agent_profile_systems must publish ProfilingContract v1 before the report discusses speed or memory.
5. agent_artifact_validator and agent_quality_safety review frozen artifacts independently and can open recall tickets.

Minimum orchestrator-subagent hops:
Hop A initial delegation with scope and acceptance criteria. Hop B contract acknowledgement after InterfaceContract v1, where each consuming agent returns ack_or_change_request. Hop C evidence review after implementation, where validator and quality findings are routed back to owning specialists. Hop D targeted PatchHandoff recall for every blocking issue, containing issue_ids_closed, changed_files, validation_evidence, and remaining_risk. The orchestrator may perform additional hops until final_gate passes.

Recall triggers:
Recall data or geometry if RCSB query payload, cache manifest, side-chain completeness rules, split manifest, or no-leakage evidence are missing. Recall model if random rotation/translation invariance tests are absent or equivariance language is incorrect. Recall ablation or training if results/ablation_results.json has fewer than one invariant baseline plus two equivariant variants without an approved deviation. Recall eval if median angular error, MAE, RMSE, per-residue counts, or circular metric definitions are missing or inconsistent across files. Recall profiling if memory numbers are implausibly tiny without method caveats and parameter or RSS estimates. Recall scicomm if report text contradicts JSON metrics, plots, or captions. Recall validator if any required deliverable path is missing, empty, unparsable, or not listed in deliverables/index.json.

Conflict resolution:
If two agents disagree, the orchestrator creates a ConflictRecord with positions, evidence, affected artifacts, and decision. Prefer the scientifically conservative option, document limitations in research_report.md, and update logs.txt.




\end{Verbatim}
\end{tcolorbox}

\subsection*{Harness $\mathcal{H}[2]$}

\begin{tcolorbox}[
breakable, colback=gray!5,colframe=black!70,title= Harness(2)
]
\begin{Verbatim}[ fontsize=\scriptsize, breaklines=true, breakanywhere=true, breaksymbolleft={}, breaksymbolright={}, obeytabs=true, tabsize=2 ]

Create an agent team with following agent candidates to solve this problem:
[
  {
    'agent_id': 'agent_orchestrator',
    'role': 'Lead scientist-orchestrator, integration owner, recall manager, and final acceptance gate.',
    'instruction': 'Own the full SE(3) side-chain χ1 training recipe. Run a multi-round workflow with explicit contract negotiation, targeted recall, patch verification, and final gating. R0 create an evidence-grounded acceptance rubric from the required deliverables plus prior failure evidence: the previous submission produced src/profiler.py while the task explicitly required src/profile.py; RCSB Search API acquisition was claimed but not evidenced by query/download manifests; the report abstract said a 30-fold throughput penalty while profiling JSON values implied roughly 100x. R1 parallel fan-out to interface/path steward, data, torsion geometry, model, ablation, training, evaluation, profiling, report, claims-audit, validator, and quality agents. R2 compile InterfaceContract v2 and RequiredPathManifest v1, then send to every consumer for ack_or_change_request; do not allow implementation to proceed for a consumer until its blocking contract changes are reconciled. R3 delegate implementation using frozen contracts. R4 run an early artifact smoke gate before expensive training, specifically checking exact required paths, RCSB provenance artifacts, imports, and uv commands. R5 freeze ResultsContract and ProfilingContract only after evaluation/profile validation. R6 allow scicomm to draft quantitative report claims only after consuming frozen result/profile contracts. R7 send the draft report and index to an independent claims referee, validator, and quality reviewer. R8 recall owning specialists with narrow PatchHandoff scopes for every blocker; repeat patch/retest until closed or explicitly waived. R9 final integration only after blocking issues are closed. Maintain issue_tracker with severity, owner, evidence path or JSON pointer, requested_fix, blocking condition, status, recall_history, retest_evidence, and waiver_rationale if any. Record agent roster, call graph, all handoffs, contract acks, recall decisions, commands, and unresolved risks in logs.txt. Prefer task-specific agents below; do not instantiate diffusion, MD, or portfolio agents unless the objective changes.',
    'output_to_orchestrator_schema': 'OrchestrationStateV2 object with keys: objective; evidence_grounded_acceptance_criteria; canonical_required_paths; shared_interface_contract; contract_ack_matrix; artifact_registry; data_provenance_registry; issue_tracker; conflict_records; delegation_rounds; recall_plan; patch_retest_log; final_gate_decision. artifact_registry entries require exact_path, required_by_task boolean, producing_agent, consuming_agents, status, size_bytes_or_sha256, validation_status, and index_json_listed boolean. final_gate_decision must cite every remaining issue_id and waiver rationale.'
  },
  {
    'agent_id': 'agent_interface_contract_steward',
    'role': 'Schema, exact-deliverable-path, and cross-agent contract steward.',
    'instruction': 'Before implementation, translate the task deliverables into a canonical RequiredPathManifest and InterfaceContract v2. Treat exact filenames as binding: src/profile.py is required; src/profiler.py may exist only as a helper and is not a substitute. Ensure deliverables/index.json will list the exact required paths, not only semantically similar paths. Build SchemaRegistry entries for DatasetRecord, ModelIO, ModelRegistry, AblationGrid, TrainingHistory, ResultsContract, ProfilingContract, PlotManifest, DataProvenanceManifest, ReportClaims, and IndexManifest. Route each draft contract to all consuming agents and collect field-level acknowledgements or change requests. Flag any missing producer, ambiguous consumer, or untestable field as a blocker before code is written.',
    'output_to_orchestrator_schema': 'ContractStewardHandoff extends HandoffV2 and must include required_path_manifest with exact_path, required_by_task, producer_agent, validator_agent, substitute_allowed false unless explicitly stated, and expected_index_entry; schema_registry with schema_id, version, required_fields, optional_fields, producer, consumer_agents, validation_method, and compatibility_notes; contract_ack_matrix with consumer_agent, schema_id, ack_status accept|change_request|blocked, requested_changes, and resolution; data_provenance_required_artifacts including data/rcsb_query.json, data/rcsb_response_ids.json, data/download_manifest.json, data/data_card.json, and data/split_manifest.json; blocker_candidates with evidence and owner.'
  },
  {
    'agent_id': 'agent_structural_data_rcsb',
    'role': 'RCSB PDB data acquisition, caching, curation, provenance, and split specialist.',
    'instruction': 'Design and implement the public PDB data path for src/extract_sidechain_dataset.py. Specify and persist the RCSB Search API POST endpoint and payload using resolution and quality filters, experimental method, polymer/protein filters, non-obsolete entries, max_structures, deterministic ordering or seeded sampling, and response ID handling. Use files.rcsb.org download URLs with caching, retries, status recording, and skipped-structure reasons. Create verifiable provenance artifacts rather than report-only claims. Coordinate with agent_torsion_geometry on residue completeness and with agent_eval_benchmarks on split leakage. The dataset must exclude ALA and GLY for χ1, keep labels derived from side-chain atoms, and prevent side-chain coordinates from entering model inputs. Prefer structure-level splits plus homology or sequence-identity risk notes; if full homology clustering is out of scope, create a documented conservative split manifest and limitation.',
    'output_to_orchestrator_schema': 'DataRCSBHandoff extends HandoffV2 and must export DataProvenanceManifest v1 and DatasetRecordDraft v2. Required artifacts: data/rcsb_query.json with endpoint, payload, query_hash, timestamp, max_structures, ordering_or_seed, and filters; data/rcsb_response_ids.json with returned_ids and selected_ids; data/download_manifest.json with pdb_id, url, cache_path, status downloaded|cached|skipped|failed, retry_count, file_size, sha256_if_available, and skip_reason; data/data_card.json with counts and extraction statistics; data/split_manifest.json with train/val/test PDB IDs, residue counts, split method, seed, no-overlap check, and homology limitation note. Include residue_record_schema, input_feature_schema explicitly excluding side-chain atom coordinates, implementation_todos_by_file, validation_commands using uv run, leakage_checklist, and requests_for_agents for geometry label definitions and eval split checks.'
  },
  {
    'agent_id': 'agent_torsion_geometry',
    'role': 'Protein side-chain chemistry, χ-angle label, completeness, and no-leakage specialist.',
    'instruction': 'Define χ1 atom names for all 18 residue types with χ1, complete heavy-side-chain atom requirements, altloc occupancy policy, insertion-code and chain handling, hetero/water exclusion, missing atom behavior, and circular angle conventions. Provide testable dihedral formulae and edge-case checks. Co-sign DatasetRecord v2 with agent_structural_data_rcsb before model, training, or eval consumes the dataset. Explicitly audit that side-chain atoms are used only to compute labels and are absent from model input tensors, saved feature tensors, and batch objects.',
    'output_to_orchestrator_schema': 'GeometryHandoff extends HandoffV2 and must include chi1_atom_definition_table; complete_sidechain_atom_table; altloc_policy with occupancy tie behavior; residue_exclusion_rules; dihedral_formula and angle_range_convention; label_fields sin_chi1, cos_chi1, chi1_degrees; DatasetRecordAck with accepted_fields, rejected_fields, and required_patches; geometry_validation_tests with commands or methods; no_sidechain_input_leakage_checks naming dataset keys inspected; downstream_schema_updates for data, model, train, and eval agents.'
  },
  {
    'agent_id': 'agent_geometric_ml',
    'role': 'SE(3)/E(3)-equivariant architecture, model API, and equivariance-test specialist.',
    'instruction': 'Design src/model_se3.py around a small reproducible GVP-style SE(3)-equivariant network for backbone context to χ1 prediction. Be precise that final χ1 output is invariant to global SE(3) transforms while internal vector channels are equivariant. Define node, edge, coordinate, graph, batch, class names, and output schemas consumed from DatasetRecord v2. Specify fair parameter budgets versus invariant baselines. Add tests for random rotation and translation invariance of model outputs and, where accessible, equivariance of vector features. Coordinate with agent_baseline_ablation on variants and with agent_training_reproducibility on trainable class names, factory functions, and checkpoint keys.',
    'output_to_orchestrator_schema': 'ModelMLHandoff extends HandoffV2 and must export ModelIOContract v2 and ModelRegistryDraft v1. Required fields: feature_dimension_table; graph_construction_contract with radius, k, batch behavior, and coordinate source; class_and_function_signatures; model_ids supported; output_schema with sin_cos order and normalization behavior; equivariance_proof_sketch; equivariance_unit_test_plan and, after implementation, EquivarianceTestEvidence with command, seed, transform description, tolerance, max_error, status, and evidence path; parameter_count_estimates; dependencies; requests_for_agents for missing DatasetRecord fields or training hooks.'
  },
  {
    'agent_id': 'agent_baseline_ablation',
    'role': 'Invariant baseline, equivariant variants, and fair-comparison ablation specialist.',
    'instruction': 'Define the controlled baseline and ablation grid. The minimum grid must include one invariant distance-only baseline and at least two equivariant rows, such as full GVP and smaller or feature-ablated GVP, unless compute limits are explicitly approved by the orchestrator as a deviation. Keep comparisons fair in split, metric implementation, optimizer, epoch budget, early-stopping policy, and parameter scale notes. Co-own ModelRegistry v1 with agent_geometric_ml and ensure train/eval/profile agents can consume model_ids without guessing.',
    'output_to_orchestrator_schema': 'AblationHandoff extends HandoffV2 and must include baseline_contract; fairness_controls; AblationGrid v2 with model_id, variant_description, model_class_or_factory, expected_checkpoint_path, expected_history_path, expected_metrics_keys, train_budget, parameter_budget, and equivariant boolean; ablation_results_json_schema; comparison_plot_spec for deliverables/plots/ablation_equivariant_vs_invariant.png; validation_checks for same split and same circular metric implementation; recall_triggers if fewer than one invariant plus two equivariant variants are present.'
  },
  {
    'agent_id': 'agent_training_reproducibility',
    'role': 'Training loop, reproducibility, configuration, checkpoints, uv workflow, and CLI integration specialist.',
    'instruction': 'Implement and validate src/train.py and any CLI integration needed for reproducible uv workflows. Use fixed seeds, deterministic split consumption, train/val/test separation, checkpointing, training_history JSON files, device fallback, short smoke-run options, and exact commands. Coordinate with model and ablation agents to train all required model_ids. Ensure failures such as NaNs, missing data, shape mismatch, checkpoint incompatibility, missing model_id, or absent split manifest produce actionable errors. If main.py is present, do not leave it as a stale hello-world entry point; either make it a useful pipeline dispatcher or keep all documented commands explicit in logs.txt and index metadata.',
    'output_to_orchestrator_schema': 'TrainingHandoff extends HandoffV2 and must include training_config_schema; seed_and_determinism_plan; split_consumption_plan referencing data/split_manifest.json; optimizer_and_schedule; checkpoint_schema with expected keys; training_history_schema; model_id_to_command_map; command_plan_full_and_smoke using uv run; produced_artifact_list with exact checkpoint/history paths; runtime_estimates; validation_results with command, exit_code, stdout_summary, stderr_summary, and evidence path; failure_debug_playbook; requests_for_model_or_data_patches if incompatibilities occur.'
  },
  {
    'agent_id': 'agent_eval_benchmarks',
    'role': 'Circular metric, benchmark, per-residue analysis, split leakage review, and plot-data specialist.',
    'instruction': 'Implement and validate src/evaluate.py and metric-producing utilities. Define circular absolute error, MAE, median, RMSE, per-residue breakdowns with counts for every evaluated model_id, and optional bootstrap uncertainty if time allows. Verify that test PDB IDs are disjoint from train/val PDB IDs using split_manifest, and document any homology limitation. Ensure results/metrics.json and results/ablation_results.json contain enough structured data for report, index, and plots. Cross-check that plots and report numbers are sourced from JSON rather than retyped. Coordinate with scicomm using a number_source_map.',
    'output_to_orchestrator_schema': 'EvalHandoff extends HandoffV2 and must export ResultsContract v2. Required fields: metric_formulae; circular_error_algorithm; metrics_json_schema with model_id keys, test_mae_degrees, test_median_error_degrees, test_rmse_degrees, n_test_samples, n_params, best_epoch, best_val_loss, and per_residue counts; ablation_results_schema; test_index_schema; split_leakage_result with train_val_test PDB overlap status; plot_data_requirements; number_source_map mapping report claim labels to JSON pointers and computed expressions; statistical_notes; validation_commands; issues_for_training_or_report if values are inconsistent.'
  },
  {
    'agent_id': 'agent_profile_systems',
    'role': 'Runtime, memory, environment, required src/profile.py path, and profiling methodology specialist.',
    'instruction': 'Implement and validate src/profile.py exactly as required by the task. A helper named src/profiler.py is allowed only if src/profile.py exists as the documented CLI/wrapper and deliverables/index.json lists src/profile.py. Profile forward and training-step latency over batch sizes for all model_ids or the main baseline pair. Capture environment details, parameter counts, device, threads, synchronization policy, warmup/repeat counts, and profiling methodology. For memory, do not rely only on tracemalloc if PyTorch native allocations are missed; include psutil RSS, torch CUDA max_memory_allocated when available, torch profiler profile_memory or a documented alternative, and parameter-memory estimates. Compute throughput ratios, especially invariant baseline vs SE3-GVP at the reported batch size, so scicomm cannot state 30x when JSON implies about 100x.',
    'output_to_orchestrator_schema': 'ProfilingHandoff extends HandoffV2 and must export ProfilingContract v2. Required artifacts: src/profile.py status created|modified|validated; results/profiling_results.json; profiling section mirrored or referenced in results/metrics.json if used by report. Required fields: profiling_protocol; environment_capture_schema; batch_sizes; runtime_memory_json_schema; memory_methods_and_caveats; cuda_memory_fields when CUDA exists; rss_fields; parameter_memory_estimates; throughput_calculation; throughput_ratio_table with numerator_model, denominator_model, batch_size, json_pointers, ratio, and narrative_phrase_recommendation; plot_spec for deliverables/plots/runtime_memory_profile.png; validation_commands; profiler_limitations; recall_triggers for missing src/profile.py, implausible undocumented memory, or report/profile ratio mismatch.'
  },
  {
    'agent_id': 'agent_scicomm_publication',
    'role': 'Conference-style report, figure narrative, and claims-evidence specialist.',
    'instruction': 'Draft research_report.md only after eval and profiling agents publish frozen ResultsContract v2 and ProfilingContract v2 and after the orchestrator supplies the number_source_map. Write a 7 to 9 page equivalent paper with abstract, introduction, related work, methods, results, profiling, limitations, conclusion, and references. Every quantitative claim must cite a JSON key or computed expression from metrics, ablation, training history, profiling, or data provenance artifacts. Check figure captions for directional correctness, for example lower MAE means outperforms, not underperforms. For speed claims, use the ProfilingContract throughput_ratio_table rather than hand-written approximations. Avoid overclaiming beyond small-scale PDB-derived χ1 prediction.',
    'output_to_orchestrator_schema': 'SciCommHandoff extends HandoffV2 and must include report_outline; figure_table_map; claims_to_evidence_matrix with claim_text, value, unit, source_path, json_pointer_or_expression, tolerance, and checked_status; exact_metric_keys_used; caption_consistency_checks; data_provenance_claims citing rcsb_query and download manifests; limitations_section_points including split/homology caveats; citation_plan; unresolved_claims_needing_eval_or_profile; recall_requests for inconsistent or missing metrics.'
  },
  {
    'agent_id': 'agent_claims_consistency_referee',
    'role': 'Independent quantitative-claims, report-metric, and index consistency referee.',
    'instruction': 'After scicomm drafts research_report.md and deliverables/index.json, independently compare every numeric or comparative claim against results/metrics.json, results/ablation_results.json, results/profiling_results.json, data/data_card.json, data/split_manifest.json, and the claims_to_evidence_matrix. Pay special attention to prior observed mismatch: 441151 samples/s versus 4136 samples/s implies about 107x, not 30x. Open issue tickets for stale text, wrong ratios, unsupported RCSB/provenance claims, or index metrics that differ from JSON. This agent does not author primary implementation; it critiques and supplies machine-checkable issue tickets.',
    'output_to_orchestrator_schema': 'ClaimsAuditHandoff extends HandoffV2 and must include numeric_claim_inventory with claim_id, file, quote, parsed_value, unit, source_json_pointer_or_missing, recomputed_value, relative_error_pct, and status; comparative_claim_checks with direction lower_is_better or higher_is_better; throughput_ratio_checks; data_claim_checks; index_summary_checks; issue_tickets with severity, owner, evidence, requested_fix, and blocking_if_fail; final_recommendation pass|fail|conditional.'
  },
  {
    'agent_id': 'agent_artifact_validator',
    'role': 'Independent CI, exact-path, schema, provenance, and reproducibility validator.',
    'instruction': 'Independently inspect the final workspace at least twice: an early smoke gate after implementation begins and a final gate after patches. Run or specify uv run smoke checks, import checks, JSON parse checks, plot non-empty checks, exact file-presence checks for every required deliverable, RCSB provenance checks, and consistency checks between research_report.md, deliverables/index.json, results/metrics.json, results/ablation_results.json, results/profiling_results.json, and data manifests. Exact path checking must fail if src/profile.py is absent even if src/profiler.py exists. This agent should not author the main implementation; it audits it and opens concrete issue tickets for orchestrator recall.',
    'output_to_orchestrator_schema': 'ValidatorHandoff extends HandoffV2 and must include file_presence_matrix with exact_path, exists, size_bytes, listed_in_index, required_by_task, and status; json_schema_results; command_results with command, exit_code, stdout_summary, stderr_summary, and evidence path; plot_integrity_results; rcsb_provenance_results checking query/download/split manifests; report_json_consistency_results; artifact_registry_diff; issue_tickets with severity and owner; final_blockers; retest_results_for_issue_ids after patches.'
  },
  {
    'agent_id': 'agent_quality_safety',
    'role': 'Scientific rigor, leakage, reproducibility, equivariance language, profiling, and compliance reviewer.',
    'instruction': 'Audit the assembled plan and artifacts for leakage, missing controls, overclaiming, incorrect equivariance language, weak ablations, invalid profiling, inconsistent metrics, unverifiable RCSB data acquisition, exact-deliverable mismatches, and reproducibility gaps. Use concrete evidence from files, JSON outputs, logs, and validation commands. Communicate through requests_for_agents when a specialist must explain or patch an issue. Final approval requires no blocking leakage, no uncorrected report-metric contradictions, documented RCSB filters and download manifest, valid circular metrics, at least one invariant plus two equivariant ablation rows unless waived, robust profiling caveats, exact src/profile.py, and a reproducible uv workflow. Safety scope is computational only; avoid wet-lab operational content.',
    'output_to_orchestrator_schema': 'QualityAuditHandoff extends HandoffV2 and must include audit_checklist; leakage_assessment; data_provenance_assessment; equivariance_assessment; ablation_assessment; profiling_assessment; reproducibility_assessment; overclaiming_assessment; exact_deliverable_path_assessment; required_recalls; final_recommendation pass|fail|conditional; evidence_paths and json_pointers.'
  }
]

Shared communication contract for every non-orchestrator agent:
Each response must be a machine-checkable HandoffV2 object with keys: agent_id, phase, scope, contract_version, inputs_consumed, assumptions, interface_exports, artifact_plan_or_changes, validation_checks, metrics_or_estimates, evidence_bundle, risks, open_questions, requests_for_agents, ack_or_change_request, recall_recommendations. inputs_consumed entries must include source_agent_id, export_name, version, artifact_path_or_json_pointer, and consumed_status. interface_exports entries must include schema_id_or_signature, producer, consumer_agents, version, status draft|frozen|deprecated, required_fields, optional_fields, compatibility_notes, and validation_method. artifact_plan_or_changes entries must include path, exact_required_path boolean, status planned|created|modified|validated, purpose, producer, consumer_agents, size_or_sha256_if_available, and validation_status. validation_checks entries must include check_id, command_or_method, expected, actual, exit_code_if_command, status pass|fail|blocked, blocking_if_fail, and evidence_path. metrics_or_estimates entries must include name, value, unit, source_path, json_pointer_or_expression, and uncertainty_or_caveat. requests_for_agents entries must include target_agent_id, question, required_response_schema, urgency, blocking_if_unanswered, and requested_by_hop. ack_or_change_request must include reviewed_schema_ids, decision accept|change_request|blocked, field_level_comments, and required_resolution. recall_recommendations must include issue_id, owner_agent_id, narrowed_scope, acceptance_criteria_for_patch, and retest_method.

Mandatory agent-to-agent handoffs:
1. agent_interface_contract_steward publishes RequiredPathManifest v1 before any implementation; every producing and validating agent must return ContractAck or ChangeRequest.
2. agent_structural_data_rcsb and agent_torsion_geometry jointly produce DatasetRecord v2 and DataProvenanceManifest v1 before model, training, or eval implementation consumes data.
3. agent_structural_data_rcsb and agent_eval_benchmarks jointly produce SplitLeakageReview v1 before training/evaluation claims are frozen.
4. agent_geometric_ml and agent_baseline_ablation jointly produce ModelRegistry v1, ModelIOContract v2, and AblationGrid v2 before training begins; agent_training_reproducibility must ack importability, checkpoint keys, and model_ids.
5. agent_eval_benchmarks publishes ResultsContract v2 before scicomm writes quantitative accuracy claims or index.json summaries.
6. agent_profile_systems publishes ProfilingContract v2 and validates exact src/profile.py before scicomm writes speed or memory claims.
7. agent_scicomm_publication sends ClaimsEvidenceMatrix to agent_claims_consistency_referee; all referee blockers are routed to the owning specialist or scicomm for patch.
8. agent_artifact_validator and agent_quality_safety review frozen artifacts independently and can open recall tickets; their final checks occur after any patch round, not only before patches.

Minimum orchestrator-subagent hops:
Hop A EvidenceRubric fan-out: orchestrator sends task deliverables, prior evidence defects, and acceptance criteria to all agents.
Hop B DraftContract return: specialists return draft exports and requests_for_agents; steward compiles InterfaceContract v2 and RequiredPathManifest v1.
Hop C Contract acknowledgement: orchestrator sends relevant contracts to consumers; each consumer returns ack_or_change_request. Blocking change requests trigger immediate contract-reconciliation recall before implementation.
Hop D Implementation delegation: orchestrator delegates files and artifacts with frozen contracts and exact acceptance checks.
Hop E Early smoke validation: validator checks exact paths, imports, RCSB provenance artifacts, and uv smoke commands before expensive training/profile/report work proceeds.
Hop F Results and profiling freeze: training, eval, and profile agents publish frozen metrics, ablation, histories, profile JSON, and number_source_map; orchestrator routes them to scicomm.
Hop G Report and index drafting: scicomm drafts report/index using only frozen contracts and emits ClaimsEvidenceMatrix.
Hop H Independent critique: claims referee, validator, and quality agents audit report, index, data provenance, exact paths, and JSON consistency.
Hop I Targeted PatchHandoff recall: for every blocker, orchestrator recalls only the owner with narrowed_scope, issue_ids_to_close, required changed files, and retest criteria. Owner returns PatchHandoff with issue_ids_closed, changed_files, diff_summary, validation_evidence, and remaining_risk.
Hop J Retest and final gate: validator or quality retests the specific issue_ids; orchestrator repeats Hop I-J until no blockers remain or a waiver is recorded with evidence and rationale.

Recall triggers:
Recall interface steward or validator if any exact required deliverable path is missing, empty, unparsable, not listed in deliverables/index.json, or substituted by a differently named file; src/profile.py missing is always blocking. Recall data if RCSB query payload, endpoint, selected IDs, download URLs, cache/download manifest, split manifest, or data_card evidence are missing or report-only. Recall geometry if χ1 atom tables, completeness rules, altloc policy, circular convention, or no-side-chain-input leakage evidence are missing. Recall model if random rotation/translation invariance tests are absent, max error/tolerance is not recorded, or equivariance/invariance language is incorrect. Recall ablation or training if results/ablation_results.json has fewer than one invariant baseline plus two equivariant variants without an approved deviation, or model_ids/checkpoint keys disagree across contracts. Recall eval if median angular error, MAE, RMSE, per-residue counts, circular metric definitions, split leakage checks, or JSON source pointers are missing or inconsistent. Recall profiling if src/profile.py is absent, memory numbers are implausibly tiny without method caveats and parameter/RSS/CUDA estimates, environment details are missing, or speed ratios used in the report differ from profiling JSON by more than 10 percent. Recall scicomm if report text contradicts JSON metrics, plots, profiling ratios, or captions, including stale statements such as 30-fold when JSON implies about 100-fold. Recall claims referee if quantitative claims are not inventoried. Recall quality if final approval lacks concrete evidence paths.

Conflict resolution:
If two agents disagree, the orchestrator creates a ConflictRecord with positions, evidence paths or JSON pointers, affected artifacts, decision, and downstream contract updates. Prefer the scientifically conservative option, document limitations in research_report.md, and update logs.txt.

Final gate conditions:
Pass only if all required files exist at exact paths including src/profile.py; deliverables/index.json lists the exact required artifacts; data/rcsb_query.json and data/download_manifest.json make PDB acquisition reproducible; DatasetRecord excludes side-chain coordinates from inputs; results/metrics.json contains MAE, median, RMSE, per-residue counts, runtime/profiling references; results/ablation_results.json contains the required invariant and equivariant variants; profiling includes robust memory caveats and throughput ratios; research_report.md quantitative claims match JSON within stated tolerances; all five required plots are non-empty; uv run smoke commands are logged; logs.txt contains the agent roster, call graph, handoffs, recall history, commands, and final_gate_decision.



\end{Verbatim}
\end{tcolorbox}

\subsection*{Harness $\mathcal{H}[3]$}

\begin{tcolorbox}[
breakable, colback=gray!5,colframe=black!70,title= Harness(3)
]
\begin{Verbatim}[ fontsize=\scriptsize, breaklines=true, breakanywhere=true, breaksymbolleft={}, breaksymbolright={}, obeytabs=true, tabsize=2 ]

Create an agent team with following agent candidates to solve this problem:
[
  {
    'agent_id': 'agent_orchestrator',
    'role': 'Lead scientist-orchestrator, integration owner, recall manager, evidence ledger owner, and final acceptance gate.',
    'instruction': 'Own the full SE(3) side-chain χ1 training recipe. Run a multi-round workflow with explicit contract negotiation, artifact evidence checks, targeted recall, patch verification, and final gating. Build the R0 acceptance rubric from the task plus comparison evidence: v0 and v1 missed exact src/profile.py and had unverifiable RCSB acquisition; v2 fixed exact src/profile.py and added provenance, but comparison history judged v1 stronger because v2 results/metrics.json lacked runtime/profiling fields, research_report.md Section 6 admitted no formal equivariance tests, profiling JSON contained weak or anomalous memory evidence such as negative gpu_memory_mb and near-zero RSS deltas, and main.py remained a stale hello-world. Preserve v2 strengths: exact profile path, RCSB query/download manifests, all required plots, one invariant plus two equivariant ablations. Regain v1 strengths: formal equivariance validation, detailed runtime and memory profiling in results/metrics.json, and stronger profiling discussion. Do not accept narrative-only agent claims: every quantitative, path, or provenance claim must be backed by evidence_path_or_json_pointer. R1 parallel fan-out to interface/path steward, data, geometry, model, ablation, training, eval, profiling, equivariance referee, metrics integrator, scicomm, claims referee, validator, quality, profiling sanity, and log auditor. R2 compile InterfaceContract v3 and RequiredPathManifest v2 and require field-level ContractAckV3 from every consumer before implementation. R3 run data-geometry and model-ablation contract reconciliation loops until no blocked consumers remain. R4 delegate implementation with frozen contracts and exact acceptance checks. R5 run an early smoke gate before expensive training, checking exact required paths, RCSB provenance artifacts, imports, uv commands, stale main.py, and availability of model equivariance-test hooks. R6 run a post-model equivariance gate by an independent referee before report claims and before final acceptance. R7 run full train/eval/profile. R8 route eval, profile, training histories, ablation, and equivariance evidence to metrics integrator; freeze results/metrics.json only after runtime, memory, profile references, and equivariance summaries are embedded or explicitly waived with evidence. R9 permit scicomm to draft quantitative report claims only after consuming frozen MetricsIntegrationContract, ResultsContract, ProfilingContract, EquivarianceAuditContract, and number_source_map. R10 send draft report, index, metrics, profile JSON, data manifests, and logs to independent claims referee, validator, profile sanity referee, quality reviewer, and log auditor. R11 recall owning specialists with narrow PatchHandoffV3 scopes for every blocker; repeat patch and retest until closed or explicitly waived. Maintain issue_tracker with severity, owner, evidence path or JSON pointer, requested_fix, blocking condition, status, recall_history, retest_evidence, and waiver_rationale. Record roster, call graph, handoffs, acks, recalls, commands, and final_gate_decision in logs.txt. Prefer task-specific agents below; do not instantiate diffusion, MD, or portfolio agents unless the objective changes.',
    'output_to_orchestrator_schema': 'OrchestrationStateV3 object with keys: objective; evidence_grounded_acceptance_criteria; prior_defect_register; canonical_required_paths; shared_interface_contract; schema_registry; contract_ack_matrix; artifact_registry; data_provenance_registry; issue_tracker; conflict_records; delegation_rounds; recall_plan; patch_retest_log; frozen_contracts; final_gate_decision. prior_defect_register entries require defect_id, source_history_or_file, evidence, prevention_gate, owner_agent, and status. artifact_registry entries require exact_path, required_by_task boolean, supporting_acceptance_path boolean, producing_agent, consuming_agents, status, size_bytes, sha256_if_available, validation_status, index_json_listed boolean, and source_command. issue_tracker entries require issue_id, severity critical|major|minor, opened_by_agent, owner_agent, evidence_path_or_json_pointer, file_excerpt_or_metric, requested_fix, blocking_condition, status open|patched|retested|closed|waived, recall_history, retest_evidence, and waiver_rationale. final_gate_decision must cite every remaining issue_id and waiver rationale.'
  },
  {
    'agent_id': 'agent_interface_contract_steward',
    'role': 'Schema, exact-deliverable-path, supporting-artifact, and cross-agent contract steward.',
    'instruction': 'Before implementation, translate the task deliverables into RequiredPathManifest v2 and InterfaceContract v3. Treat exact filenames as binding: src/profile.py is required; src/profiler.py may exist only as a helper and is not a substitute. Ensure deliverables/index.json will list exact required artifacts, not semantic substitutes. Add supporting acceptance artifacts that are not task deliverables but are required to close known defects: results/equivariance_tests.json for formal SE(3) checks, results/profiling_results.json for raw profile evidence, training history JSON files, data manifests, and logs.txt. Build SchemaRegistry entries for DatasetRecord, ModelIO, ModelRegistry, AblationGrid, TrainingHistory, ResultsContract, ProfilingContract, EquivarianceAuditContract, MetricsIntegrationContract, PlotManifest, DataProvenanceManifest, ReportClaims, IndexManifest, RuntimeSummaryInMetrics, and LogManifest. Route each draft contract to all consuming agents and collect field-level acknowledgements or change requests. Flag missing producer, ambiguous consumer, untestable field, missing JSON pointer, or absent validation command as a blocker before code is written.',
    'output_to_orchestrator_schema': 'ContractStewardHandoffV3 extends HandoffV3 and must include required_path_manifest with exact_path, required_by_task, supporting_acceptance_path, producer_agent, validator_agent, substitute_allowed false unless explicitly stated, expected_index_entry, expected_schema_id, and smoke_validation_method; schema_registry with schema_id, version, required_fields, optional_fields, producer, consumer_agents, validation_method, expected_json_pointers, compatibility_notes, and freeze_prerequisites; contract_ack_matrix with consumer_agent, schema_id, field_name, ack_status accept|change_request|blocked, requested_changes, resolution, and reack_required; data_provenance_required_artifacts including data/rcsb_query.json, data/rcsb_response_ids.json, data/download_manifest.json, data/data_card.json, and data/split_manifest.json; known_defect_prevention_map tying src/profile.py, metrics runtime, equivariance tests, profiling memory sanity, stale main.py, and logs.txt to explicit validators; blocker_candidates with evidence and owner.'
  },
  {
    'agent_id': 'agent_structural_data_rcsb',
    'role': 'RCSB PDB data acquisition, caching, curation, provenance, and split specialist.',
    'instruction': 'Design and implement the public PDB data path for src/extract_sidechain_dataset.py. Persist the RCSB Search API POST endpoint and payload using resolution and quality filters, experimental method, polymer/protein filters, non-obsolete entries, max_structures, deterministic ordering or seeded sampling, and response ID handling. Use files.rcsb.org download URLs with caching, retries, status recording, and skipped-structure reasons. Create verifiable provenance artifacts rather than report-only claims. Coordinate with geometry on residue completeness and with eval on split leakage. Exclude ALA and GLY for χ1, keep labels derived from side-chain atoms, and prevent side-chain coordinates from entering model inputs. Prefer structure-level splits plus sequence-identity or homology risk notes; if full clustering is out of scope, create a documented conservative split manifest and limitation. Produce a small deterministic smoke dataset option so validator can rerun extraction without downloading the full cache.',
    'output_to_orchestrator_schema': 'DataRCSBHandoffV3 extends HandoffV3 and must export DataProvenanceManifest v2 and DatasetRecordDraft v3. Required artifacts: data/rcsb_query.json with endpoint, http_method POST, payload, query_hash, timestamp, max_structures, ordering_or_seed, filters, and selected_fields; data/rcsb_response_ids.json with returned_ids, selected_ids, selection_rule, and query_hash; data/download_manifest.json with pdb_id, url, cache_path, status downloaded|cached|skipped|failed, retry_count, http_status_if_attempted, file_size, sha256_if_available, and skip_reason; data/data_card.json with counts and extraction statistics; data/split_manifest.json with train/val/test PDB IDs, residue counts, split method, seed, no-overlap check, and homology limitation note. Include residue_record_schema, input_feature_schema explicitly excluding side-chain atom coordinates, dataset_key_inventory from actual serialized dataset, implementation_todos_by_file, validation_commands using uv run, leakage_checklist, and requests_for_agents for geometry labels and eval split checks.'
  },
  {
    'agent_id': 'agent_torsion_geometry',
    'role': 'Protein side-chain chemistry, χ-angle label, completeness, and no-leakage specialist.',
    'instruction': 'Define χ1 atom names for all 18 residue types with χ1, complete heavy-side-chain atom requirements, altloc occupancy policy, insertion-code and chain handling, hetero/water exclusion, missing atom behavior, and circular angle conventions. Provide testable dihedral formulae and edge-case checks. Co-sign DatasetRecord v3 with data before model, training, or eval consumes the dataset. Explicitly audit that side-chain atoms are used only to compute labels and are absent from model input tensors, saved feature tensors, and batch objects.',
    'output_to_orchestrator_schema': 'GeometryHandoffV3 extends HandoffV3 and must include chi1_atom_definition_table; complete_sidechain_atom_table; altloc_policy with occupancy tie behavior; residue_exclusion_rules; dihedral_formula and angle_range_convention; label_fields sin_chi1, cos_chi1, chi1_degrees; DatasetRecordAck with accepted_fields, rejected_fields, required_patches, and co_sign_status; geometry_validation_tests with commands or methods; no_sidechain_input_leakage_checks naming dataset keys inspected and evidence_path; downstream_schema_updates for data, model, train, eval, and equivariance agents; issue_tickets if side-chain coordinates appear in any input feature field.'
  },
  {
    'agent_id': 'agent_geometric_ml',
    'role': 'SE(3)/E(3)-equivariant architecture, model API, importability, and equivariance-test hook specialist.',
    'instruction': 'Design src/model_se3.py around a small reproducible GVP-style SE(3)-equivariant network for backbone context to χ1 prediction. Be precise that final χ1 output is invariant to global SE(3) transforms while internal vector channels are equivariant. Define node, edge, coordinate, graph, batch, class names, factory functions, and output schemas consumed from DatasetRecord v3. Specify fair parameter budgets versus invariant baselines. Implement or expose a reusable run_equivariance_check hook or CLI path that the independent equivariance referee can call without training the full model. Coordinate with ablation on variants and with training on model_ids, checkpoint keys, and factory signatures. Do not allow the report to claim lack of formal equivariance tests unless a blocking issue is waived by the orchestrator.',
    'output_to_orchestrator_schema': 'ModelMLHandoffV3 extends HandoffV3 and must export ModelIOContract v3 and ModelRegistryDraft v2. Required fields: feature_dimension_table; graph_construction_contract with radius or k, batch behavior, and coordinate source; class_and_function_signatures; model_ids supported; checkpoint_key_contract; output_schema with sin_cos order and normalization behavior; equivariance_hook_contract with callable_name_or_command, required_inputs, seed_control, and expected_output_json; equivariance_proof_sketch; parameter_count_estimates; dependency_contract; import_validation_evidence; requests_for_agents for missing DatasetRecord fields or training hooks; after implementation, preliminary_equivariance_smoke_evidence with command, seed, transform description, tolerance, max_error, status, and evidence_path.'
  },
  {
    'agent_id': 'agent_equivariance_referee',
    'role': 'Independent SE(3) transformation-test and equivariance-language referee.',
    'instruction': 'Independently test implemented model behavior rather than trusting model-author claims. Consume ModelIOContract, DatasetRecord sample schema, and model factories. Run random rotation and translation tests on synthetic batches and at least one real dataset batch for every equivariant model_id; confirm scalar χ1 outputs are invariant within tolerance. If internal vector features are exposed, verify vector-channel equivariance under the same rotations. For invariant baseline, verify distance-only outputs are invariant and note that it is not an equivariant vector model. Write results/equivariance_tests.json and a short language guidance object for scicomm. If tests cannot run, open a blocking issue instead of allowing research_report.md to state no formal tests as a limitation.',
    'output_to_orchestrator_schema': 'EquivarianceAuditHandoffV3 extends HandoffV3 and must include EquivarianceAuditContract v1 with artifact_path results/equivariance_tests.json; model_ids_tested; command_results with command, exit_code, stdout_summary, stderr_summary, evidence_path; transform_cases with seed, rotation_matrix_source, translation_vector, batch_source synthetic|real, n_samples, tolerance, max_abs_output_error, max_vector_equivariance_error_if_available, and status; invariant_baseline_check; failures_with_reproduction_commands; report_language_requirements; issue_tickets with owner agent_geometric_ml or agent_training_reproducibility; final_recommendation pass|fail|conditional.'
  },
  {
    'agent_id': 'agent_baseline_ablation',
    'role': 'Invariant baseline, equivariant variants, and fair-comparison ablation specialist.',
    'instruction': 'Define the controlled baseline and ablation grid. The minimum grid must include one invariant distance-only baseline and at least two equivariant rows, such as full GVP and smaller or feature-ablated GVP, unless compute limits are explicitly approved as a deviation. Keep comparisons fair in split, metric implementation, optimizer, epoch budget, early-stopping policy, and parameter scale notes. Co-own ModelRegistry v2 with model agent and ensure train, eval, profile, metrics integrator, and report agents can consume model_ids without guessing.',
    'output_to_orchestrator_schema': 'AblationHandoffV3 extends HandoffV3 and must include baseline_contract; fairness_controls; AblationGrid v3 with model_id, variant_description, model_class_or_factory, expected_checkpoint_path, expected_history_path, expected_metrics_keys, train_budget, parameter_budget, equivariant boolean, and profiler_required boolean; ablation_results_json_schema; comparison_plot_spec for deliverables/plots/ablation_equivariant_vs_invariant.png; validation_checks for same split and same circular metric implementation; consumer_ack_status from training, eval, profile, and metrics integrator; recall_triggers if fewer than one invariant plus two equivariant variants are present.'
  },
  {
    'agent_id': 'agent_training_reproducibility',
    'role': 'Training loop, reproducibility, configuration, checkpoints, uv workflow, CLI integration, and stale-entrypoint owner.',
    'instruction': 'Implement and validate src/train.py and any CLI integration needed for reproducible uv workflows. Use fixed seeds, deterministic split consumption, train/val/test separation, checkpointing, training_history JSON files, device fallback, short smoke-run options, and exact commands. Coordinate with model and ablation agents to train all required model_ids. Ensure failures such as NaNs, missing data, shape mismatch, checkpoint incompatibility, missing model_id, or absent split manifest produce actionable errors. If main.py exists, do not leave it as a stale hello-world entry point; either make it a useful pipeline dispatcher or ensure documented commands do not rely on it and validator confirms no stale hello text. Return command evidence, not just a plan.',
    'output_to_orchestrator_schema': 'TrainingHandoffV3 extends HandoffV3 and must include training_config_schema; seed_and_determinism_plan; split_consumption_plan referencing data/split_manifest.json; optimizer_and_schedule; checkpoint_schema with expected keys; training_history_schema; model_id_to_command_map; command_plan_full_and_smoke using uv run; produced_artifact_list with exact checkpoint and history paths; runtime_estimates; cli_entrypoint_status with main_py_status useful_dispatcher|not_used|patched and stale_hello_check; validation_results with command, exit_code, stdout_summary, stderr_summary, evidence_path, and issue_ids_closed; failure_debug_playbook; requests_for_model_or_data_patches if incompatibilities occur.'
  },
  {
    'agent_id': 'agent_eval_benchmarks',
    'role': 'Circular metric, benchmark, per-residue analysis, split leakage review, and plot-data specialist.',
    'instruction': 'Implement and validate src/evaluate.py and metric-producing utilities. Define circular absolute error, MAE, median, RMSE, per-residue breakdowns with counts for every evaluated model_id, and optional bootstrap uncertainty if time allows. Verify test PDB IDs are disjoint from train/val PDB IDs using split_manifest and document homology limitation. Ensure results/metrics.json and results/ablation_results.json contain enough structured accuracy data for report, index, plots, and metrics integrator. Cross-check that plots and report numbers are sourced from JSON rather than retyped. Coordinate with scicomm using a number_source_map and with metrics integrator to embed runtime and equivariance sections into results/metrics.json.',
    'output_to_orchestrator_schema': 'EvalHandoffV3 extends HandoffV3 and must export ResultsContract v3. Required fields: metric_formulae; circular_error_algorithm; metrics_json_accuracy_schema with model_id keys, test_mae_degrees, test_median_error_degrees, test_rmse_degrees, n_test_samples, n_params, best_epoch, best_val_loss, and per_residue counts; ablation_results_schema; test_index_schema; split_leakage_result with train_val_test PDB overlap status; plot_data_requirements; number_source_map mapping report claim labels to JSON pointers and computed expressions; required_runtime_merge_request describing fields expected from profile agent in results/metrics.json; statistical_notes; validation_commands; issues_for_training_or_report if values are inconsistent.'
  },
  {
    'agent_id': 'agent_profile_systems',
    'role': 'Runtime, memory, environment, required src/profile.py path, and profiling methodology specialist.',
    'instruction': 'Implement and validate src/profile.py exactly as required by the task. A helper named src/profiler.py is allowed only if src/profile.py exists as the documented CLI/wrapper and deliverables/index.json lists src/profile.py. Profile forward and training-step latency over batch sizes for all model_ids or at minimum the main baseline pair plus all ablation rows when feasible. Capture environment details, parameter counts, device, threads, synchronization policy, warmup/repeat counts, and profiling methodology. For memory, do not rely only on tracemalloc or tiny RSS deltas. Include psutil total RSS, parameter-memory estimates, torch CUDA max_memory_allocated when CUDA exists, torch profiler profile_memory or a documented alternative, and caveats for allocator noise. Never publish negative deltas such as negative gpu_memory_mb as memory usage; preserve raw deltas separately with caveat if needed. Compute model-to-model throughput ratios, especially invariant baseline vs SE3-GVP at the reported batch size, so scicomm cannot invent ratios. Export profile summaries to metrics integrator for inclusion in results/metrics.json.',
    'output_to_orchestrator_schema': 'ProfilingHandoffV3 extends HandoffV3 and must export ProfilingContract v3. Required artifacts: src/profile.py status created|modified|validated; results/profiling_results.json; profile_summary_for_metrics_json. Required fields: profiling_protocol; environment_capture_schema; batch_sizes; forward_latency_schema; training_step_latency_schema; runtime_memory_json_schema; memory_methods_and_caveats; cuda_memory_fields when CUDA exists; rss_fields; parameter_memory_estimates; raw_memory_delta_fields_with_caveats; throughput_calculation; model_to_model_throughput_ratio_table with numerator_model, denominator_model, batch_size, json_pointers, ratio, and narrative_phrase_recommendation; plot_spec for deliverables/plots/runtime_memory_profile.png; validation_commands; profiler_limitations; recall_triggers for missing src/profile.py, absent training-step profile, implausible undocumented memory, negative memory usage fields, or report/profile ratio mismatch.'
  },
  {
    'agent_id': 'agent_profile_sanity_referee',
    'role': 'Independent profiling-method, memory-sanity, and speed-ratio referee.',
    'instruction': 'After profile agent publishes ProfilingContract, independently inspect results/profiling_results.json, profile_summary_for_metrics_json, and runtime plot data. Recompute throughput from latency and batch size, recompute model-to-model ratios, check that memory fields are physically interpretable, and check that training-step latency is either present or explicitly waived. Pay attention to v2 evidence where gpu_memory_mb was negative and RSS deltas were near zero. Open issue tickets for missing memory caveats, negative usage fields, absent parameter-memory estimates, missing model-to-model ratios, or summaries not merged into results/metrics.json.',
    'output_to_orchestrator_schema': 'ProfilingSanityHandoffV3 extends HandoffV3 and must include ratio_recomputations with model_id, batch_size, reported_throughput, recomputed_throughput, relative_error_pct, and status; memory_anomaly_inventory with json_pointer, value, expected_property, caveat_present, and status; training_step_profile_check; metrics_json_runtime_merge_check; issue_tickets with owner agent_profile_systems or agent_metrics_artifact_integrator; final_recommendation pass|fail|conditional.'
  },
  {
    'agent_id': 'agent_metrics_artifact_integrator',
    'role': 'Cross-artifact metrics merger, JSON pointer source-map owner, and results/metrics.json completeness gate.',
    'instruction': 'Consume frozen ResultsContract, ProfilingContract, EquivarianceAuditContract, TrainingHistory schemas, DataProvenanceManifest, and AblationGrid. Produce or patch results/metrics.json so it contains the required accuracy metrics plus runtime/profiling references required by the task and comparison history. Maintain exact JSON pointers for every number used by report and index. Embed or reference profiling runtime, memory, throughput ratio, training-step latency, environment, and equivariance summary. Do not let scicomm proceed if results/metrics.json lacks runtime/profiling fields, because this was a v2 regression relative to the stronger v1 submission.',
    'output_to_orchestrator_schema': 'MetricsIntegrationHandoffV3 extends HandoffV3 and must include MetricsIntegrationContract v1 with metrics_json_path results/metrics.json; required_top_level_keys including model_ids, runtime, profiling, equivariance, data_summary, and source_artifacts; merged_json_pointers for each MAE, median, RMSE, per_residue count, checkpoint, runtime, memory, throughput_ratio, training_step_latency, and equivariance status; source_pointer_map with source_path, source_json_pointer, destination_json_pointer, transform_expression, tolerance, and status; cross_file_consistency_results comparing metrics, ablation, profile, histories, equivariance, index, and plots; missing_field_blockers; validation_commands using uv run python or jq-compatible Python; freeze_status draft|frozen|blocked; issue_tickets for eval, profile, model, or scicomm owners.'
  },
  {
    'agent_id': 'agent_scicomm_publication',
    'role': 'Conference-style report, figure narrative, limitation wording, and claims-evidence specialist.',
    'instruction': 'Draft research_report.md only after orchestrator supplies frozen MetricsIntegrationContract, ResultsContract, ProfilingContract, EquivarianceAuditContract, and number_source_map. Write a 7 to 9 page equivalent paper with abstract, introduction, related work, methods, results, profiling, limitations, conclusion, and references. Every quantitative claim must cite a JSON key or computed expression from metrics, ablation, training history, profiling, equivariance, or data provenance artifacts. Check figure captions for directional correctness, for example lower MAE means outperforms. For speed claims, use ProfilingContract model_to_model_throughput_ratio_table rather than hand-written approximations. For equivariance, cite results/equivariance_tests.json if passing; if failed or waived, state the exact evidence and limitation without falsely saying tests were not included. Avoid overclaiming beyond small-scale PDB-derived χ1 prediction.',
    'output_to_orchestrator_schema': 'SciCommHandoffV3 extends HandoffV3 and must include report_outline; figure_table_map; claims_to_evidence_matrix with claim_text, value, unit, source_path, json_pointer_or_expression, tolerance, and checked_status; exact_metric_keys_used; equivariance_claims_to_evidence citing results/equivariance_tests.json or waiver; profiling_claims_to_evidence citing results/metrics.json runtime fields and results/profiling_results.json; caption_consistency_checks; data_provenance_claims citing rcsb_query and download manifests; limitations_section_points including split and homology caveats; stale_phrase_scan_results for unsupported phrases such as no formal equivariance tests when tests exist; citation_plan; unresolved_claims_needing_eval_profile_or_equivariance; recall_requests for inconsistent or missing metrics.'
  },
  {
    'agent_id': 'agent_claims_consistency_referee',
    'role': 'Independent quantitative-claims, report-metric, equivariance-language, and index consistency referee.',
    'instruction': 'After scicomm drafts research_report.md and deliverables/index.json, independently compare every numeric or comparative claim against results/metrics.json, results/ablation_results.json, results/profiling_results.json, results/equivariance_tests.json, data/data_card.json, data/split_manifest.json, and the claims_to_evidence_matrix. Pay attention to prior mismatch patterns: speed ratios must equal JSON-derived throughput ratios; metrics in index must match results JSON; data claims must match RCSB manifests; report must not say formal equivariance tests are absent if results/equivariance_tests.json exists and passes. This agent critiques and supplies machine-checkable issue tickets, not primary implementation.',
    'output_to_orchestrator_schema': 'ClaimsAuditHandoffV3 extends HandoffV3 and must include numeric_claim_inventory with claim_id, file, quote_or_excerpt, parsed_value, unit, source_json_pointer_or_missing, recomputed_value, relative_error_pct, and status; comparative_claim_checks with direction lower_is_better or higher_is_better; throughput_ratio_checks; equivariance_language_checks; data_claim_checks; index_summary_checks; unsupported_claims; issue_tickets with severity, owner, evidence, requested_fix, and blocking_if_fail; final_recommendation pass|fail|conditional.'
  },
  {
    'agent_id': 'agent_artifact_validator',
    'role': 'Independent CI, exact-path, schema, provenance, metrics-completeness, and reproducibility validator.',
    'instruction': 'Independently inspect the workspace at least four times: early smoke after implementation begins, post-data/model gate, post-results/profile/metrics-integration gate, and final gate after patches. Run or specify uv run smoke checks, import checks, JSON parse checks, plot non-empty checks, exact file-presence checks for every required deliverable, RCSB provenance checks, equivariance artifact checks, stale main.py checks, and consistency checks across research_report.md, deliverables/index.json, results/metrics.json, results/ablation_results.json, results/profiling_results.json, results/equivariance_tests.json, and data manifests. Exact path checking must fail if src/profile.py is absent even if src/profiler.py exists. results/metrics.json must fail if it lacks runtime or profiling summary fields. This agent audits and opens concrete issue tickets for orchestrator recall.',
    'output_to_orchestrator_schema': 'ValidatorHandoffV3 extends HandoffV3 and must include validation_phase early_smoke|post_model|post_metrics|final; file_presence_matrix with exact_path, exists, size_bytes, listed_in_index, required_by_task, supporting_acceptance_path, and status; json_schema_results; command_results with command, exit_code, stdout_summary, stderr_summary, and evidence_path; import_results; plot_integrity_results; rcsb_provenance_results checking query/download/split manifests; metrics_completeness_results checking runtime, profiling, memory, and equivariance keys; report_json_consistency_results; stale_entrypoint_results for main.py; artifact_registry_diff; issue_tickets with severity and owner; final_blockers; retest_results_for_issue_ids after patches.'
  },
  {
    'agent_id': 'agent_quality_safety',
    'role': 'Scientific rigor, leakage, reproducibility, equivariance language, profiling, and compliance reviewer.',
    'instruction': 'Audit assembled plans and artifacts for leakage, missing controls, overclaiming, incorrect equivariance language, weak ablations, invalid profiling, inconsistent metrics, unverifiable RCSB acquisition, exact-deliverable mismatches, missing runtime in metrics, missing formal equivariance evidence, stale entrypoints, and reproducibility gaps. Use concrete evidence from files, JSON outputs, logs, and validation commands. Communicate through requests_for_agents when a specialist must explain or patch an issue. Final approval requires no blocking leakage, documented RCSB filters and download manifest, valid circular metrics, formal equivariance checks or explicit waiver, at least one invariant plus two equivariant ablation rows unless waived, robust profiling caveats, runtime and profiling in results/metrics.json, exact src/profile.py, and a reproducible uv workflow. Safety scope is computational only; avoid wet-lab operational content.',
    'output_to_orchestrator_schema': 'QualityAuditHandoffV3 extends HandoffV3 and must include audit_checklist; leakage_assessment; data_provenance_assessment; equivariance_assessment; ablation_assessment; profiling_assessment; metrics_completeness_assessment; reproducibility_assessment; overclaiming_assessment; exact_deliverable_path_assessment; required_recalls; final_recommendation pass|fail|conditional; evidence_paths and json_pointers.'
  },
  {
    'agent_id': 'agent_reproducibility_log_auditor',
    'role': 'logs.txt, command history, agent-call graph, uv workflow, and reproducibility-record auditor.',
    'instruction': 'Verify the save-rule requirements are met. logs.txt must document the plan, execution, reflection, all created and spawned agents, the workflow between agents as a structured outline or diagram, which agents were executed, handoffs, recall history, commands, and final gate decision. Cross-check commands in logs against actual artifacts and uv usage. Open blockers for missing roster, missing call graph, missing recall records, missing uv commands, or logs that do not mention final unresolved risks.',
    'output_to_orchestrator_schema': 'LogAuditHandoffV3 extends HandoffV3 and must include log_manifest with logs_path, exists, size_bytes, sections_present, agent_roster_present, call_graph_present, commands_present, recall_history_present, final_gate_present, and status; uv_workflow_check; command_to_artifact_traceability; issue_tickets with owner agent_orchestrator or relevant specialist; final_recommendation pass|fail|conditional.'
  }
]

Shared communication contract for every non-orchestrator agent:
Each response must be a machine-checkable HandoffV3 object with keys: handoff_id, agent_id, hop_id, phase, scope, contract_version, inputs_consumed, assumptions, interface_exports, artifact_plan_or_changes, validation_checks, metrics_or_estimates, evidence_bundle, risks, open_questions, requests_for_agents, ack_or_change_request, issue_tickets, recall_recommendations, and blocking_status. inputs_consumed entries must include source_agent_id, export_name, version, artifact_path_or_json_pointer, consumed_status, and freshness_requirement. interface_exports entries must include schema_id_or_signature, producer, consumer_agents, version, status draft|frozen|deprecated, required_fields, optional_fields, compatibility_notes, expected_json_pointers, validation_method, and consumer_ack_required. artifact_plan_or_changes entries must include path, exact_required_path boolean, supporting_acceptance_path boolean, status planned|created|modified|validated, purpose, producer, consumer_agents, size_bytes, sha256_if_available, and validation_status. validation_checks entries must include check_id, command_or_method, expected, actual, exit_code_if_command, status pass|fail|blocked, blocking_if_fail, issue_id_if_fail, and evidence_path. metrics_or_estimates entries must include name, value, unit, source_path, json_pointer_or_expression, uncertainty_or_caveat, and destination_json_pointer_if_merged. evidence_bundle entries must include artifact_path, json_pointer, file_excerpt_or_hash, producing_command, and generated_at. requests_for_agents entries must include target_agent_id, question, required_response_schema, urgency, blocking_if_unanswered, requested_by_hop, and due_before_hop. ack_or_change_request must include reviewed_schema_ids, field_level_comments, decision accept|change_request|blocked, required_resolution, and reack_required. issue_tickets must include issue_id, severity, owner_agent_id, evidence_path_or_json_pointer, requested_fix, acceptance_criteria_for_patch, and retest_method. recall_recommendations must include issue_id, owner_agent_id, narrowed_scope, changed_files_expected, acceptance_criteria_for_patch, and retest_method. blocking_status must be pass|blocked|conditional with blocker_issue_ids.

Patch protocol:
Any recalled owner must return PatchHandoffV3 with issue_ids_to_close, changed_files, diff_summary, contract_fields_changed, backward_compatibility_notes, validation_evidence, issue_ids_closed, issue_ids_still_open, regression_risks, and requested_retests. The orchestrator must route PatchHandoffV3 to the original opener plus validator or quality for retest before closing the issue.

Mandatory agent-to-agent handoffs:
1. agent_interface_contract_steward publishes RequiredPathManifest v2 and InterfaceContract v3 before implementation; every producing and validating agent must return ContractAckV3 or ChangeRequestV3 with field-level comments.
2. agent_structural_data_rcsb and agent_torsion_geometry jointly produce DatasetRecord v3 and DataProvenanceManifest v2 before model, training, eval, profile, or equivariance code consumes data.
3. agent_structural_data_rcsb and agent_eval_benchmarks jointly produce SplitLeakageReview v2 before training and evaluation claims are frozen.
4. agent_geometric_ml and agent_baseline_ablation jointly produce ModelRegistry v2, ModelIOContract v3, and AblationGrid v3 before training begins; agent_training_reproducibility, agent_eval_benchmarks, agent_profile_systems, agent_equivariance_referee, and agent_metrics_artifact_integrator must ack importability, checkpoint keys, model_ids, and factory signatures.
5. agent_equivariance_referee publishes EquivarianceAuditContract v1 and results/equivariance_tests.json before scicomm writes equivariance claims or final acceptance; failures trigger recall to model or training.
6. agent_eval_benchmarks publishes ResultsContract v3 before metrics integration and scicomm accuracy claims.
7. agent_profile_systems publishes ProfilingContract v3 and validates exact src/profile.py before metrics integration and speed or memory claims.
8. agent_profile_sanity_referee audits ProfilingContract before scicomm profiling text; blockers route to profile or metrics integrator.
9. agent_metrics_artifact_integrator freezes MetricsIntegrationContract only after eval, profile, ablation, training, data, and equivariance inputs are consumed and results/metrics.json contains runtime, profiling, memory, and equivariance summaries or explicit waivers.
10. agent_scicomm_publication sends ClaimsEvidenceMatrix to agent_claims_consistency_referee; all referee blockers route to the owning specialist or scicomm for patch.
11. agent_artifact_validator, agent_quality_safety, agent_profile_sanity_referee, and agent_reproducibility_log_auditor review frozen artifacts independently and can open recall tickets; their final checks occur after any patch round, not only before patches.

Minimum orchestrator-subagent hops:
Hop 0 EvidenceRubric fan-out: orchestrator sends task deliverables, prior comparison defects, v2 file evidence, and acceptance criteria to all agents.
Hop 1 DraftContract return: specialists return draft exports and requests_for_agents; steward compiles RequiredPathManifest v2 and InterfaceContract v3.
Hop 2 Contract acknowledgement: orchestrator sends relevant contracts to consumers; every consumer returns field-level ack_or_change_request. Blocking changes trigger contract-reconciliation recall before implementation.
Hop 3 Contract reconciliation loop: orchestrator recalls only conflicted producers and consumers with the disputed fields, requested resolution, and retest method; repeat until frozen or waived.
Hop 4 Data-geometry implementation and co-sign: data and geometry implement extraction, manifests, residue chemistry, and leakage checks; eval acks split/leakage review.
Hop 5 Model-ablation implementation and import smoke: model and ablation implement factories and variants; training, eval, profile, and equivariance agents ack actual import and shape evidence.
Hop 6 Early artifact smoke validation: validator checks exact paths, imports, stale main.py, RCSB provenance, DatasetRecord keys, and uv smoke commands before expensive training/profile/report work proceeds.
Hop 7 Equivariance pre-report gate: independent equivariance referee runs random rotation and translation tests; failures recall model before scicomm and before final gate.
Hop 8 Training smoke and full run: training runs smoke then full or bounded training for all model_ids; failures recall data/model/ablation with narrow scope.
Hop 9 Evaluation and split-leakage freeze: eval computes circular metrics, per-residue breakdowns, ablation results, and number_source_map; data/eval co-sign no-overlap evidence.
Hop 10 Profiling and sanity audit: profile publishes runtime, training-step latency, memory, and ratios; profile sanity referee recomputes ratios and audits memory anomalies.
Hop 11 Metrics integration freeze: metrics integrator merges eval, profile, training, data, ablation, and equivariance into results/metrics.json with JSON pointers; validator checks that runtime and profiling fields exist.
Hop 12 Plot and index generation: responsible agents generate all plots and index from frozen JSON; validator checks non-empty plots and index path coverage.
Hop 13 Report drafting: scicomm drafts report using only frozen contracts and emits ClaimsEvidenceMatrix.
Hop 14 Independent critique: claims referee, validator, quality, profile sanity, and log auditor audit report, index, data provenance, exact paths, logs, JSON consistency, and profiling/equivariance language.
Hop 15 Targeted PatchHandoff recall: for every blocker, orchestrator recalls only the owner with narrowed_scope, issue_ids_to_close, required changed files, and retest criteria. Owner returns PatchHandoffV3.
Hop 16 Retest and final gate: original opener plus validator or quality retests specific issue_ids; orchestrator repeats Hop 15 and Hop 16 until no blockers remain or a waiver is recorded with evidence and rationale.

Recall triggers:
Recall interface steward or validator if any exact required deliverable path is missing, empty, unparsable, not listed in deliverables/index.json, or substituted by a differently named file; src/profile.py missing is always blocking. Recall data if RCSB query payload, endpoint, selected IDs, download URLs, cache/download manifest, split manifest, data_card evidence, or deterministic extraction command are missing or report-only. Recall geometry if χ1 atom tables, completeness rules, altloc policy, circular convention, or no-side-chain-input leakage evidence are missing. Recall model if random rotation and translation test hooks are absent, ModelRegistry imports fail, max error and tolerance are not recorded, or equivariance/invariance language is incorrect. Recall equivariance referee if results/equivariance_tests.json is absent, not run on implemented code, or lacks pass/fail status and tolerances. Recall ablation or training if results/ablation_results.json has fewer than one invariant baseline plus two equivariant variants without approved deviation, model_ids or checkpoint keys disagree, training histories are absent, or main.py remains a stale hello-world. Recall eval if median angular error, MAE, RMSE, per-residue counts, circular metric definitions, split leakage checks, or JSON source pointers are missing or inconsistent. Recall profiling if src/profile.py is absent, training-step profiling is absent without waiver, memory numbers are implausibly tiny without method caveats and parameter/RSS/CUDA estimates, any negative value is reported as memory usage, environment details are missing, or speed ratios used in report differ from profiling JSON by more than 10 percent. Recall profile sanity if ratios or memory anomalies are not independently inventoried. Recall metrics integrator if results/metrics.json lacks runtime, profiling, memory, throughput-ratio, source-artifact, or equivariance summary fields. Recall scicomm if report text contradicts JSON metrics, plots, profiling ratios, equivariance evidence, or captions, including stale statements that formal equivariance tests were not included when results/equivariance_tests.json exists. Recall claims referee if quantitative claims are not inventoried. Recall log auditor or orchestrator if logs.txt lacks roster, call graph, handoffs, recall history, uv commands, or final_gate_decision. Recall quality if final approval lacks concrete evidence paths.

Conflict resolution:
If two agents disagree, the orchestrator creates a ConflictRecord with positions, evidence paths or JSON pointers, affected artifacts, downstream contract fields, decision, and updated consumers. Prefer the scientifically conservative option, document limitations in research_report.md, and update logs.txt. Contract changes after implementation require reack from all affected consumers and targeted retest.

Final gate conditions:
Pass only if all required files exist at exact paths including src/profile.py; deliverables/index.json lists exact required artifacts and supporting provenance artifacts; data/rcsb_query.json, data/rcsb_response_ids.json, and data/download_manifest.json make PDB acquisition reproducible; DatasetRecord excludes side-chain coordinates from inputs; formal equivariance checks are present in results/equivariance_tests.json and cited or explicitly waived; results/metrics.json contains MAE, median, RMSE, per-residue counts, runtime, profiling and memory summary, throughput ratios, source-artifact pointers, and equivariance summary; results/ablation_results.json contains required invariant and equivariant variants; profiling includes forward and training-step latency or waiver, robust memory caveats, parameter/RSS/CUDA estimates where applicable, and model-to-model throughput ratios; research_report.md quantitative claims match JSON within stated tolerances and does not contain stale unsupported limitations; all five required plots are non-empty and generated from frozen JSON; main.py is not a stale hello-world if present; uv run smoke commands are logged; logs.txt contains agent roster, call graph, handoffs, recall history, commands, unresolved risks, and final_gate_decision.


\end{Verbatim}
\end{tcolorbox}

\subsection*{Harness $\mathcal{H}[4]$}

\begin{tcolorbox}[
breakable, colback=gray!5,colframe=black!70,title= Harness(4)
]
\begin{Verbatim}[ fontsize=\scriptsize, breaklines=true, breakanywhere=true, breaksymbolleft={}, breaksymbolright={}, obeytabs=true, tabsize=2 ]

Create an agent team with following agent candidates to solve this problem:
[
  {
    'agent_id': 'agent_orchestrator',
    'role': 'Lead scientist-orchestrator, integration owner, recall manager, evidence ledger owner, and final acceptance gate.',
    'instruction': 'Own the full SE(3) side-chain χ1 training recipe. Run a multi-round workflow with explicit contract negotiation, artifact evidence checks, targeted recall, patch verification, and final gating. Build the R0 acceptance rubric from the task plus comparison and current-v3 evidence. Preserve v2 strengths that won on data provenance: RCSB query, response IDs, download manifest, and coherent metrics. Preserve v3 strengths: exact src/profile.py, all required plots, runtime/profiling fields in metrics, one invariant plus two equivariant ablations, and results/equivariance_tests.json. Fix defects now proven in v3 evidence: research_report.md Section 6 says RCSB Search API returned 400 and a predefined fallback list was used; main.py still prints Hello from query220-ourteam; results/metrics.json has se3_gvp top-level best_epoch 9 and best_val_loss 0.705876 while embedded train_history and results/training_history_se3_gvp.json show best_epoch 20 and best_val_loss 0.693450; research_report.md claims SE3-GVP training took about 2100 s and small took about 1300 s while histories show 905.33 s and 820.09 s; results/profiling_results.json has high-variance and nonmonotonic timings such as invariant B32 latency_std 8.14 ms greater than mean 3.28 ms and B64 faster than B32; metrics.json profiling summaries omit training-step latency and RSS/CUDA memory even though the profile file contains them; results/equivariance_tests.json lacks command_results, real-batch evidence, and vector-channel details. Do not accept narrative-only claims. Every quantitative, path, protocol, or provenance claim must cite evidence_path_or_json_pointer. R1 parallel fan-out to all agents with these concrete defects. R2 compile RequiredPathManifest v4, InterfaceContract v4, MetricsPointerMatrix v2, and PriorDefectRegister v4; require field-level ContractAckV4 from every consumer. R3 run targeted contract reconciliation loops for RCSB protocol, DatasetRecord, ModelRegistry, MetricsLineage, ProfilingContract, EquivarianceAudit, and ReportClaims until no blocked consumer remains. R4 freeze contracts and delegate implementation. R5 run RCSB protocol gate before any training: a successful query-derived selection is preferred; if fallback is used, it must be verified per PDB ID against RCSB metadata and the report must state the deviation. R6 run data-geometry co-sign and no-leakage audit. R7 run model-ablation import and shape smoke. R8 run independent entrypoint CI to patch stale main.py before final. R9 run pre-report formal equivariance tests with command evidence. R10 run training, evaluation, profiling, metrics integration, JSON consistency audit, plots, index, and report only after their upstream contracts are frozen. R11 send frozen artifacts to independent claims, RCSB, profile sanity, JSON consistency, validator, quality, and log auditors. R12 recall owning specialists with narrow PatchHandoffV4 scopes for every blocker and rerun original opener plus validator or quality retests. Maintain issue_tracker with severity, owner, evidence path or JSON pointer, requested_fix, blocking condition, status, recall_history, retest_evidence, and waiver_rationale. Record roster, call graph, handoffs, acks, conflicts, recalls, commands, and final_gate_decision in logs.txt.',
    'output_to_orchestrator_schema': 'OrchestrationStateV4 object with keys: objective; evidence_grounded_acceptance_criteria; prior_defect_register; current_v3_defect_register; canonical_required_paths; supporting_acceptance_paths; shared_interface_contract; schema_registry; metrics_pointer_matrix; contract_ack_matrix; artifact_registry; data_provenance_registry; rcsb_protocol_gate; issue_tracker; conflict_records; delegation_rounds; recall_plan; patch_retest_log; frozen_contracts; waiver_register; final_gate_decision. current_v3_defect_register entries require defect_id, evidence_path_or_json_pointer, excerpt_or_value, prevention_gate, owner_agent, status, and retest_method. artifact_registry entries require exact_path, required_by_task boolean, supporting_acceptance_path boolean, producing_agent, consuming_agents, status, size_bytes, sha256_if_available, validation_status, index_json_listed boolean, source_command, and freshness_hop. final_gate_decision must cite every remaining issue_id and waiver rationale.'
  },
  {
    'agent_id': 'agent_interface_contract_steward',
    'role': 'Schema, exact-deliverable-path, supporting-artifact, metrics-pointer, and cross-agent contract steward.',
    'instruction': 'Before implementation, translate the task deliverables and known evidence defects into RequiredPathManifest v4 and InterfaceContract v4. Exact filenames are binding: src/profile.py is required and src/profiler.py is not a substitute. Add supporting acceptance artifacts: data/rcsb_query_attempts.json, data/rcsb_entry_metadata.json, data/rcsb_filter_verification.json, results/equivariance_tests.json, results/profiling_results.json, results/metrics_lineage.json, results/json_consistency_audit.json, training histories, data manifests, and logs.txt. Build schemas for DatasetRecord, ModelIO, ModelRegistry, AblationGrid, TrainingHistory, CheckpointMetadata, ResultsContract, ProfilingContract, EquivarianceAuditContract, MetricsIntegrationContract, MetricsPointerMatrix, DataProvenanceManifest, RCSBProtocolAudit, ReportClaims, IndexManifest, RuntimeSummaryInMetrics, and LogManifest. Define required JSON pointers for results/metrics.json including accuracy, per-residue counts, best_epoch, best_val_loss, training_time_s, forward_latency, training_step_latency, throughput, RSS/CUDA/parameter memory, profiling caveats, source artifacts, and equivariance status. Route each draft contract to all producers and consumers and collect ContractAckV4 or ChangeRequestV4 with field-level comments before code is written.',
    'output_to_orchestrator_schema': 'ContractStewardHandoffV4 extends HandoffV4 and must include required_path_manifest with exact_path, required_by_task, supporting_acceptance_path, producer_agent, validator_agent, substitute_allowed false unless explicitly stated, expected_index_entry, expected_schema_id, and smoke_validation_method; schema_registry with schema_id, version, required_fields, optional_fields, producer, consumer_agents, validation_method, expected_json_pointers, compatibility_notes, and freeze_prerequisites; metrics_pointer_matrix with destination_path results/metrics.json, required_json_pointer, source_artifact, source_json_pointer_or_expression, producer_agent, consumer_agents, tolerance, validator_agent, and blocking_if_missing; contract_ack_matrix with consumer_agent, schema_id, field_name, consumer_use, ack_status accept|change_request|blocked, requested_changes, resolution, and reack_required; known_defect_prevention_map tying RCSB fallback, stale main.py, se3_gvp best_epoch mismatch, report training-time mismatch, high-variance profiling, missing metrics memory/training-step fields, and weak equivariance artifacts to validators; blocker_candidates with evidence and owner.'
  },
  {
    'agent_id': 'agent_structural_data_rcsb',
    'role': 'RCSB PDB data acquisition, caching, curation, fallback verification, provenance, and split specialist.',
    'instruction': 'Implement src/extract_sidechain_dataset.py as a verifiable public PDB acquisition pipeline. Prefer live RCSB Search API POST with resolution, experimental method, polymer/protein, non-obsolete, quality, max_structures, deterministic ordering or seed, and returned ID persistence. If the Search API fails, do not silently use a predefined list. Record the failure in data/rcsb_query_attempts.json, then either repair the query or verify every fallback PDB ID through RCSB core metadata and PDB header evidence, including resolution, method, polymer type, and obsolete status. Do not let the report say the fallback was returned by the query. Coordinate with geometry for complete χ1 labels, with eval for structure-level splits, and with RCSB protocol referee for filter verification. Exclude ALA and GLY for χ1 and prevent side-chain coordinates from entering model inputs. Produce a deterministic smoke mode that can run without full downloads.',
    'output_to_orchestrator_schema': 'DataRCSBHandoffV4 extends HandoffV4 and must export DataProvenanceManifest v4 and DatasetRecordDraft v4. Required artifacts: data/rcsb_query.json with endpoint, http_method POST, payload, query_hash, timestamp, max_structures, ordering_or_seed, filters, selected_fields, and intended_result_limit; data/rcsb_query_attempts.json with attempt_id, endpoint, payload_hash, http_status, response_excerpt_or_error, timestamp, retry_policy, and resolution_action; data/rcsb_response_ids.json with returned_ids, selected_ids, selection_rule, query_hash, and source_attempt_id; data/rcsb_entry_metadata.json with pdb_id, resolution_angstrom, experimental_method, polymer_entity_type, deposition_status_or_obsolete_flag, metadata_source, and verification_status; data/rcsb_filter_verification.json with per_id_filter_pass, failed_filters, aggregate_pass_rate, fallback_used none|verified_cached_response|verified_manual_list|unverified_blocked, and report_language_required; data/download_manifest.json with pdb_id, url, cache_path, status downloaded|cached|skipped|failed, retry_count, http_status_if_attempted, file_size, sha256_if_available, and skip_reason; data/data_card.json; data/split_manifest.json. Include residue_record_schema, input_feature_schema explicitly excluding side-chain atom coordinates, dataset_key_inventory from the actual serialized dataset, validation_commands using uv run, leakage_checklist, split_no_overlap_evidence, and requests_for_agents.'
  },
  {
    'agent_id': 'agent_rcsb_protocol_referee',
    'role': 'Independent RCSB acquisition-protocol and fallback-verification referee.',
    'instruction': 'Audit the implemented RCSB acquisition path independently of the data agent. Reproduce or inspect the RCSB query attempts, verify that selected IDs came from a successful query or from a filter-verified fallback, and check every selected PDB ID against RCSB metadata or PDB headers. Open a blocking issue if the workflow repeats the v3 defect where research_report.md admits a 400 query and a predefined fallback while still describing the dataset as high-resolution X-ray without proving filters. Provide required report wording for success, verified fallback, or unverified fallback.',
    'output_to_orchestrator_schema': 'RCSBProtocolAuditHandoffV1 extends HandoffV4 and must include audit_artifact_path data/rcsb_protocol_audit.json; query_attempt_assessment with attempt_id, http_status, payload_hash, status success|repair_needed|failed_with_verified_fallback|failed_unverified; selected_id_lineage with pdb_id, origin query_returned|cached_query_response|fallback, lineage_evidence_path, and status; filter_verification_summary with n_selected, n_verified_resolution, n_verified_method, n_verified_non_obsolete, n_failed, failed_ids, and blocking_status; fallback_language_requirements; issue_tickets with owner agent_structural_data_rcsb or agent_scicomm_publication; final_recommendation pass|fail|conditional.'
  },
  {
    'agent_id': 'agent_torsion_geometry',
    'role': 'Protein side-chain chemistry, χ-angle label, completeness, and no-leakage specialist.',
    'instruction': 'Define χ1 atom names for all 18 residue types with χ1, complete heavy-side-chain atom requirements, altloc occupancy policy, insertion-code and chain handling, hetero/water exclusion, missing atom behavior, circular angle conventions, and dihedral formula tests. Co-sign DatasetRecord v4 with data before model, train, eval, profile, or equivariance consumes it. Explicitly audit data/train_features_sample.json, dataset serialized keys, batch objects, and model input tensors to prove side-chain atom coordinates are not inputs.',
    'output_to_orchestrator_schema': 'GeometryHandoffV4 extends HandoffV4 and must include chi1_atom_definition_table; complete_sidechain_atom_table; altloc_policy with occupancy tie behavior; residue_exclusion_rules; dihedral_formula and angle_range_convention; label_fields sin_chi1, cos_chi1, chi1_degrees; DatasetRecordAck with accepted_fields, rejected_fields, required_patches, co_sign_status, and accepted_dataset_artifact_hash; geometry_validation_tests with commands or methods; no_sidechain_input_leakage_checks naming dataset keys inspected, batch keys inspected, model-forward keys inspected, evidence_path, and status; downstream_schema_updates; issue_tickets if side-chain coordinates appear in any input feature field.'
  },
  {
    'agent_id': 'agent_geometric_ml',
    'role': 'SE(3)/E(3)-equivariant architecture, model API, importability, checkpoint, and equivariance-test hook specialist.',
    'instruction': 'Design src/model_se3.py around a small reproducible GVP-style SE(3)-equivariant network for backbone context to χ1 prediction. State that final χ1 output is invariant to global SE(3) transforms while internal vector channels are equivariant. Define graph construction, model factories, model_ids, checkpoint keys, and output schema. Expose a callable or CLI equivariance test hook that the independent referee can run on synthetic and real batches without full training. Coordinate with ablation, training, profile, eval, and metrics agents so model_ids, checkpoint metadata, and parameter counts are not guessed.',
    'output_to_orchestrator_schema': 'ModelMLHandoffV4 extends HandoffV4 and must export ModelIOContract v4 and ModelRegistryDraft v4. Required fields: feature_dimension_table; graph_construction_contract; class_and_function_signatures; model_ids_supported; checkpoint_key_contract; output_schema with sin_cos order and normalization behavior; parameter_count_estimates; dependency_contract; import_validation_evidence; equivariance_hook_contract with callable_name_or_command, required_inputs, supports_synthetic_batch, supports_real_batch, seed_control, expected_output_json_schema, and tolerance; preliminary_equivariance_smoke_evidence with command, seed, transform description, max_error, status, and evidence_path; requests_for_agents.'
  },
  {
    'agent_id': 'agent_equivariance_referee',
    'role': 'Independent SE(3) transformation-test and equivariance-language referee.',
    'instruction': 'Independently test the implemented model rather than trusting model-author claims. Run random rotation and translation tests on synthetic batches and at least one real dataset batch for every equivariant model_id. Confirm scalar χ1 outputs are invariant within tolerance; if internal vector features are exposed, verify vector-channel equivariance. For invariant baseline, verify distance-only invariance and label it as an invariant baseline, not an equivariant vector model. Write results/equivariance_tests.json with command_results and batch provenance. If tests cannot run, open a blocker; do not allow report text to claim formal tests passed without command evidence.',
    'output_to_orchestrator_schema': 'EquivarianceAuditHandoffV4 extends HandoffV4 and must include EquivarianceAuditContract v2 with artifact_path results/equivariance_tests.json; schema_validation_status; model_ids_tested; checkpoint_paths_or_factory_modes; command_results with command, exit_code, stdout_summary, stderr_summary, evidence_path, and working_directory; transform_cases with model_id, seed, rotation_matrix_source, translation_vector, batch_source synthetic|real, real_batch_source_path_if_any, n_samples, tolerance, max_abs_output_error, mean_abs_output_error, max_vector_equivariance_error_if_available, and status; invariant_baseline_check; failures_with_reproduction_commands; report_language_requirements; issue_tickets; final_recommendation pass|fail|conditional.'
  },
  {
    'agent_id': 'agent_baseline_ablation',
    'role': 'Invariant baseline, equivariant variants, fair-comparison ablation, and ablation artifact specialist.',
    'instruction': 'Define one invariant distance-only baseline and at least two equivariant variants, such as full GVP and small or feature-ablated GVP. Keep split, circular metric, optimizer, epoch budget, early stopping, and parameter reporting fair. Co-own ModelRegistry v4 with model agent and ensure training, eval, profile, metrics, and report agents consume model_ids and checkpoint paths without guessing.',
    'output_to_orchestrator_schema': 'AblationHandoffV4 extends HandoffV4 and must include baseline_contract; fairness_controls; AblationGrid v4 with model_id, variant_description, model_class_or_factory, expected_checkpoint_path, expected_history_path, expected_metrics_keys, train_budget, parameter_budget, equivariant boolean, and profiler_required boolean; ablation_results_json_schema; comparison_plot_spec; validation_checks for same split and same circular metric; consumer_ack_status from training, eval, profile, metrics, and report; recall_triggers if fewer than one invariant plus two equivariant variants are present.'
  },
  {
    'agent_id': 'agent_training_reproducibility',
    'role': 'Training loop, reproducibility, configurations, checkpoints, histories, uv workflow, and checkpoint-metadata consistency owner.',
    'instruction': 'Implement and validate src/train.py. Use fixed seeds, deterministic split consumption, checkpoints, training_history JSON files, device fallback, smoke-run options, and exact uv commands. For every model, checkpoint metadata, standalone training_history file, and embedded metrics train_history must agree on model_id, n_params, best_epoch, best_val_loss, epochs, seed, and total_training_time_s. Prevent the v3 mismatch where se3_gvp metrics top-level best_epoch differed from train_history. Do not leave main.py as stale hello-world; coordinate with entrypoint CI.',
    'output_to_orchestrator_schema': 'TrainingHandoffV4 extends HandoffV4 and must include training_config_schema; seed_and_determinism_plan; split_consumption_plan referencing data/split_manifest.json; optimizer_and_schedule; checkpoint_schema with expected keys; checkpoint_metadata_table with model_id, checkpoint_path, best_epoch, best_val_loss, n_params, seed, and history_path; training_history_schema; model_id_to_command_map; command_plan_full_and_smoke using uv run; produced_artifact_list; runtime_estimates_and_actuals with source paths; consistency_assertions comparing checkpoints and history files; cli_entrypoint_status request to agent_cli_entrypoint_ci; validation_results with command, exit_code, stdout_summary, stderr_summary, evidence_path, and issue_ids_closed; failure_debug_playbook; requests_for_model_or_data_patches.'
  },
  {
    'agent_id': 'agent_cli_entrypoint_ci',
    'role': 'Stale main.py, CLI discoverability, uv smoke command, and reproducibility entrypoint sentinel.',
    'instruction': 'Independently inspect and execute the repository entrypoints. If main.py exists, it must not contain stale hello-world text and must either dispatch to documented pipeline commands or clearly print help for train/evaluate/profile/extract. Run grep-style checks for Hello from query220 and run uv run python main.py --help or equivalent. Open a blocking issue if stale text remains, because current v3 evidence shows main.py prints Hello from query220-ourteam despite the v3 design saying this should be blocked.',
    'output_to_orchestrator_schema': 'EntrypointCIHandoffV1 extends HandoffV4 and must include main_py_status useful_dispatcher|documented_not_used|stale_hello|missing_ok; stale_text_scan with command_or_method, matches, evidence_path, and status; cli_smoke_results with command, exit_code, stdout_summary, stderr_summary, and status; documented_command_crosscheck with logs_path, commands_found, commands_execute_status, and missing_commands; issue_tickets with owner agent_training_reproducibility or agent_orchestrator; final_recommendation pass|fail|conditional.'
  },
  {
    'agent_id': 'agent_eval_benchmarks',
    'role': 'Circular metric, benchmark, per-residue analysis, split leakage review, and metric-lineage specialist.',
    'instruction': 'Implement and validate src/evaluate.py. Define circular absolute error, MAE, median, RMSE, per-residue breakdowns with counts, and split leakage checks. Ensure results/metrics.json and results/ablation_results.json can be generated from frozen histories, checkpoints, predictions, and split manifests. Create results/metrics_lineage.json with exact source pointers and expressions. Detect and open blockers for top-level metric values that disagree with histories, checkpoint metadata, or ablation JSON. The v3 se3_gvp best_epoch mismatch must be impossible after this gate.',
    'output_to_orchestrator_schema': 'EvalHandoffV4 extends HandoffV4 and must export ResultsContract v4. Required fields: metric_formulae; circular_error_algorithm; metrics_json_accuracy_schema with model_id keys, test_mae_degrees, test_median_error_degrees, test_rmse_degrees, n_test_samples, n_params, best_epoch, best_val_loss, total_training_time_s, and per_residue counts; ablation_results_schema; test_index_schema; split_leakage_result; per_residue_count_sum_check; predictions_length_check; metric_lineage_map with source_path, source_json_pointer_or_expression, destination_json_pointer, tolerance, and status; required_runtime_merge_request describing profile fields expected in metrics; validation_commands; issues_for_training_profile_or_report.'
  },
  {
    'agent_id': 'agent_profile_systems',
    'role': 'Runtime, memory, environment, required src/profile.py path, profiling methodology, and metrics profiling payload specialist.',
    'instruction': 'Implement and validate src/profile.py exactly. Profile forward and training-step latency for all model_ids or approved subset, capture environment, parameter counts, device, threads, synchronization, warmup/repeats, and memory. Do not publish unstable timing means without caveats: compute coefficient of variation, monotonicity warnings, median and p95, and use median for ratios when means are unstable. Include psutil RSS, parameter memory, CUDA max_memory_allocated when CUDA exists, torch profiler memory or documented alternative, and allocator caveats. Never publish negative memory usage. Export a profile_summary_for_metrics_json that includes training_step_latency and RSS/CUDA/parameter memory pointers, not only forward latency.',
    'output_to_orchestrator_schema': 'ProfilingHandoffV4 extends HandoffV4 and must export ProfilingContract v4. Required artifacts: src/profile.py status created|modified|validated; results/profiling_results.json; profile_summary_for_metrics_json. Required fields: profiling_protocol; environment_capture_schema; batch_sizes; forward_latency_schema with mean, median, p95, std, cv, and stability_status; training_step_latency_schema with same statistics; runtime_memory_json_schema; cuda_memory_fields when CUDA exists; rss_fields; parameter_memory_estimates; raw_memory_delta_fields_with_caveats; profiling_quality_flags for high_cv, nonmonotonic_batch_scaling, tiny_rss_delta, and missing_cuda; throughput_calculation; model_to_model_throughput_ratio_table with numerator_model, denominator_model, batch_size, statistic_used mean|median, json_pointers, ratio, and narrative_phrase_recommendation; metrics_merge_payload with required destination pointers for runtime, profiling, memory, throughput ratios, and training_step_latency; plot_spec; validation_commands; recall_triggers.'
  },
  {
    'agent_id': 'agent_profile_sanity_referee',
    'role': 'Independent profiling-method, memory-sanity, timing-stability, and speed-ratio referee.',
    'instruction': 'After profile agent publishes ProfilingContract, independently inspect results/profiling_results.json and runtime plot data. Recompute throughput and ratios, check high-variance and nonmonotonic timings, verify memory fields are physically interpretable, and verify training-step latency and memory fields are merged into results/metrics.json. Open blockers for missing caveats, negative usage fields, absent parameter/RSS/CUDA estimates, missing ratios, or report/profile ratio mismatch. The v3 high-variance B32 invariant timing and nonmonotonic B64/B32 timing must either be rerun or caveated.',
    'output_to_orchestrator_schema': 'ProfilingSanityHandoffV4 extends HandoffV4 and must include ratio_recomputations with model_id, batch_size, statistic_used, reported_throughput, recomputed_throughput, relative_error_pct, and status; timing_anomaly_inventory with model_id, batch_size, json_pointer, mean, std, cv, median, monotonicity_context, caveat_present, and status; memory_anomaly_inventory with json_pointer, value, expected_property, caveat_present, and status; training_step_profile_check; metrics_json_runtime_memory_merge_check; issue_tickets; final_recommendation pass|fail|conditional.'
  },
  {
    'agent_id': 'agent_metrics_artifact_integrator',
    'role': 'Cross-artifact metrics merger, JSON pointer source-map owner, and results/metrics.json completeness gate.',
    'instruction': 'Consume frozen ResultsContract, ProfilingContract, EquivarianceAuditContract, TrainingHistory schemas, CheckpointMetadata, DataProvenanceManifest, RCSBProtocolAudit, and AblationGrid. Produce or patch results/metrics.json so it contains accuracy, runtime, profiling, memory, throughput ratios, training-step latency, data summary, source artifacts, equivariance summary, and consistency_checks. Do not freeze if any top-level model best_epoch or best_val_loss differs from training_history or checkpoint metadata. Do not let scicomm proceed if metrics lacks runtime/profiling fields, memory fields, training-step latency, or source pointers.',
    'output_to_orchestrator_schema': 'MetricsIntegrationHandoffV4 extends HandoffV4 and must include MetricsIntegrationContract v2 with metrics_json_path results/metrics.json; required_top_level_keys including model_ids, models, runtime, profiling, memory, throughput_ratios, equivariance, data_summary, source_artifacts, and consistency_checks; required_pointer_status_matrix with each pointer from MetricsPointerMatrix, source_path, source_json_pointer_or_expression, destination_json_pointer, value, tolerance, and status; merged_json_pointers for each MAE, median, RMSE, per_residue count, checkpoint, best_epoch, best_val_loss, total_training_time_s, forward latency, training_step_latency, memory, throughput_ratio, and equivariance status; cross_file_consistency_results comparing metrics, ablation, profile, histories, checkpoints, equivariance, data, index, and plots; missing_field_blockers; freeze_status draft|frozen|blocked; issue_tickets.'
  },
  {
    'agent_id': 'agent_json_consistency_referee',
    'role': 'Independent JSON-lineage, metric consistency, and cross-artifact numerical forensics referee.',
    'instruction': 'Independently compare results/metrics.json, results/ablation_results.json, training history files, checkpoint metadata, results/profiling_results.json, results/equivariance_tests.json, data manifests, plots metadata if present, and report claims. Recompute best_epoch and best_val_loss from each training history, recompute ablation improvements, recompute total training times, and verify required pointers exist. Open blockers for the exact v3 mismatch pattern: se3_gvp top-level best_epoch or best_val_loss disagreeing with embedded train_history or results/training_history_se3_gvp.json.',
    'output_to_orchestrator_schema': 'JSONConsistencyAuditHandoffV1 extends HandoffV4 and must include artifact_path results/json_consistency_audit.json; model_history_consistency with model_id, metrics_best_epoch_pointer, history_best_epoch_pointer, checkpoint_best_epoch_pointer, values, status, and evidence_path; training_time_consistency with model_id, metrics_pointer, history_pointer, report_claim_if_available, values, tolerance, and status; ablation_consistency with source_pointers, recomputed_values, and status; profiling_merge_consistency; equivariance_schema_consistency; data_summary_consistency; issue_tickets with severity and owners; final_recommendation pass|fail|conditional.'
  },
  {
    'agent_id': 'agent_scicomm_publication',
    'role': 'Conference-style report, figure narrative, limitation wording, and claims-evidence specialist.',
    'instruction': 'Draft research_report.md only after frozen MetricsIntegrationContract, ResultsContract, ProfilingContract, RCSBProtocolAudit, EquivarianceAuditContract, JSONConsistencyAudit, and number_source_map. Every quantitative or comparative claim must cite a JSON pointer or computed expression. Use verified RCSB wording: if fallback was used, state it accurately and do not claim IDs were returned by the failed query. Use training times from history or metrics pointers, not profile estimates. Avoid CPU/GPU contradictions. For speed claims, use the profile ratio table and caveats. For equivariance, cite results/equivariance_tests.json with command evidence or state the exact waiver.',
    'output_to_orchestrator_schema': 'SciCommHandoffV4 extends HandoffV4 and must include report_outline; figure_table_map; claims_to_evidence_matrix with claim_id, claim_text, parsed_value, unit, source_path, json_pointer_or_expression, tolerance, checked_status, and dependent_gate; exact_metric_keys_used; rcsb_claims_to_evidence citing query attempts and filter verification; equivariance_claims_to_evidence; profiling_claims_to_evidence; training_time_claims_to_evidence; caption_consistency_checks; data_provenance_claims; limitations_section_points; stale_phrase_scan_results for unsupported phrases; contradiction_scan_results for CPU vs CUDA, fallback vs query-returned, and formal-test language; unresolved_claims_needing_eval_profile_or_equivariance; recall_requests.'
  },
  {
    'agent_id': 'agent_claims_consistency_referee',
    'role': 'Independent quantitative-claims, report-metric, RCSB-language, profiling, equivariance, and index consistency referee.',
    'instruction': 'After scicomm drafts research_report.md and deliverables/index.json, parse every numeric, approximate, comparative, and provenance claim. Compare against results/metrics.json, ablation, profiling, equivariance, RCSB query attempts, filter verification, data_card, split_manifest, and claims_to_evidence_matrix. Specifically catch report training-time claims that differ from histories, speed ratios not derived from profile ratios, CPU/GPU contradictions, and claims that a failed RCSB query produced the selected set. This agent critiques and opens machine-checkable issue tickets, not primary implementation.',
    'output_to_orchestrator_schema': 'ClaimsAuditHandoffV4 extends HandoffV4 and must include numeric_claim_inventory with claim_id, file, quote_or_excerpt, parsed_value, unit, source_json_pointer_or_missing, recomputed_value, tolerance, relative_error_pct, and status; approximate_claim_checks; comparative_claim_checks with direction lower_is_better or higher_is_better; throughput_ratio_checks; training_time_claim_checks; rcsb_language_checks; equivariance_language_checks; data_claim_checks; index_summary_checks; unsupported_claims; contradiction_checks; issue_tickets with severity, owner, evidence, requested_fix, and blocking_if_fail; final_recommendation pass|fail|conditional.'
  },
  {
    'agent_id': 'agent_artifact_validator',
    'role': 'Independent CI, exact-path, schema, provenance, metrics-completeness, entrypoint, and reproducibility validator.',
    'instruction': 'Inspect the workspace at least six times: early smoke, RCSB protocol gate, post-data/model gate, post-results/profile/metrics-integration gate, post-report critique, and final gate after patches. Run uv run smoke checks, import checks, JSON parse checks, exact file-presence checks, plot non-empty checks, RCSB provenance checks, equivariance artifact schema checks, stale main.py checks, metrics pointer matrix checks, and report JSON consistency checks. Exact path checking fails if src/profile.py is absent. results/metrics.json fails if it lacks runtime, profiling, memory, training_step_latency, source pointers, or equivariance summary. main.py fails if it contains Hello from query220.',
    'output_to_orchestrator_schema': 'ValidatorHandoffV4 extends HandoffV4 and must include validation_phase early_smoke|rcsb_protocol|post_model|post_metrics|post_report|final; file_presence_matrix with exact_path, exists, size_bytes, listed_in_index, required_by_task, supporting_acceptance_path, and status; json_schema_results; metrics_pointer_matrix_results; command_results with command, exit_code, stdout_summary, stderr_summary, and evidence_path; import_results; plot_integrity_results; rcsb_provenance_results; equivariance_artifact_results; metrics_completeness_results; report_json_consistency_results; stale_entrypoint_results; artifact_registry_diff; issue_tickets; final_blockers; retest_results_for_issue_ids.'
  },
  {
    'agent_id': 'agent_quality_safety',
    'role': 'Scientific rigor, leakage, reproducibility, equivariance language, profiling, provenance, and compliance reviewer.',
    'instruction': 'Audit assembled plans and artifacts for leakage, missing controls, overclaiming, incorrect equivariance language, weak ablations, invalid profiling, inconsistent metrics, unverifiable or fallback RCSB acquisition, exact deliverable mismatches, missing runtime/memory in metrics, stale entrypoints, and reproducibility gaps. Use concrete evidence from files, JSON outputs, logs, and validation commands. Final approval requires no blocking leakage, documented and verified RCSB filters or explicit verified fallback, valid circular metrics, formal equivariance checks with command evidence or waiver, at least one invariant plus two equivariant variants unless waived, robust profiling caveats, runtime/profiling/memory/training-step in results/metrics.json, exact src/profile.py, no stale main.py, and reproducible uv workflow.',
    'output_to_orchestrator_schema': 'QualityAuditHandoffV4 extends HandoffV4 and must include audit_checklist; leakage_assessment; data_provenance_assessment; rcsb_fallback_assessment; equivariance_assessment; ablation_assessment; profiling_assessment; metrics_completeness_assessment; metric_consistency_assessment; reproducibility_assessment; overclaiming_assessment; exact_deliverable_path_assessment; required_recalls; final_recommendation pass|fail|conditional; evidence_paths and json_pointers.'
  },
  {
    'agent_id': 'agent_reproducibility_log_auditor',
    'role': 'logs.txt, command history, agent-call graph, uv workflow, recall history, and reproducibility-record auditor.',
    'instruction': 'Verify logs.txt documents plan, execution, reflection, all created and spawned agents, workflow between agents, executed agents, handoffs, contract acks, recall history, patch retests, commands, unresolved risks, and final gate decision. Cross-check commands in logs against artifacts and uv usage. Open blockers for missing roster, call graph, recalls, uv commands, RCSB gate notes, JSON consistency gate notes, or final unresolved risks.',
    'output_to_orchestrator_schema': 'LogAuditHandoffV4 extends HandoffV4 and must include log_manifest with logs_path, exists, size_bytes, sections_present, agent_roster_present, call_graph_present, contract_ack_present, commands_present, recall_history_present, patch_retest_present, final_gate_present, unresolved_risks_present, and status; uv_workflow_check; command_to_artifact_traceability; issue_tickets; final_recommendation pass|fail|conditional.'
  }
]

Shared communication contract for every non-orchestrator agent:
Each response must be a machine-checkable HandoffV4 object with keys: handoff_id, agent_id, hop_id, phase, scope, contract_version, inputs_consumed, assumptions, interface_exports, artifact_plan_or_changes, validation_checks, metrics_or_estimates, evidence_bundle, downstream_impact, self_audit_against_prior_defects, risks, open_questions, requests_for_agents, ack_or_change_request, issue_tickets, recall_recommendations, and blocking_status. inputs_consumed entries must include source_agent_id, export_name, version, artifact_path_or_json_pointer, consumed_status, and freshness_requirement. interface_exports entries must include schema_id_or_signature, producer, consumer_agents, version, status draft|frozen|deprecated, required_fields, optional_fields, compatibility_notes, expected_json_pointers, validation_method, consumer_ack_required, and breaking_change_if_modified. artifact_plan_or_changes entries must include path, exact_required_path boolean, supporting_acceptance_path boolean, status planned|created|modified|validated, purpose, producer, consumer_agents, size_bytes, sha256_if_available, validation_status, and source_command. validation_checks entries must include check_id, command_or_method, expected, actual, exit_code_if_command, status pass|fail|blocked, blocking_if_fail, issue_id_if_fail, and evidence_path. metrics_or_estimates entries must include name, value, unit, source_path, json_pointer_or_expression, uncertainty_or_caveat, and destination_json_pointer_if_merged. evidence_bundle entries must include artifact_path, json_pointer, file_excerpt_or_hash, producing_command, exit_code_if_command, generated_at, and freshness_hop. downstream_impact entries must include consumer_agent_id, consumed_field, downstream_artifact, and reack_required_if_changed. self_audit_against_prior_defects must include rows for src/profile.py path, RCSB fallback/provenance, runtime/profiling in metrics, formal equivariance evidence, stale main.py, best_epoch consistency, report training-time consistency, profiling anomalies, and logs completeness. requests_for_agents entries must include target_agent_id, question, required_response_schema, urgency, blocking_if_unanswered, requested_by_hop, and due_before_hop. ack_or_change_request must include reviewed_schema_ids, field_level_comments, decision accept|change_request|blocked, required_resolution, and reack_required. issue_tickets must include issue_id, severity critical|major|minor, owner_agent_id, evidence_path_or_json_pointer, file_excerpt_or_metric, requested_fix, acceptance_criteria_for_patch, retest_method, and history_trigger_if_any. recall_recommendations must include issue_id, owner_agent_id, narrowed_scope, changed_files_expected, acceptance_criteria_for_patch, and retest_method. blocking_status must be pass|blocked|conditional with blocker_issue_ids.

Contract acknowledgement protocol:
The orchestrator may not freeze any shared contract until every producer, consumer, and validator returns ContractAckV4 or ChangeRequestV4. ContractAckV4 requires schema_id, version, agent_id, field_name, consumer_use, accepted_type_or_constraint, validation_method, ack_status accept|change_request|blocked, requested_change, blocking_reason, and reack_required. Contract changes after implementation require reack from all affected consumers and targeted retest.

Patch protocol:
Any recalled owner must return PatchHandoffV4 with issue_ids_to_close, changed_files, diff_summary, contract_fields_changed, backward_compatibility_notes, validation_evidence, metrics_or_pointer_changes, issue_ids_closed, issue_ids_still_open, regression_risks, and requested_retests. The orchestrator must route PatchHandoffV4 to the original opener plus validator or quality for retest before closing an issue. If a patch changes metrics, profile, data, or report claims, the metrics integrator, JSON consistency referee, claims referee, and log auditor must re-run their affected checks.

Mandatory agent-to-agent handoffs:
1. Interface steward publishes RequiredPathManifest v4, InterfaceContract v4, and MetricsPointerMatrix v2 before implementation; every producing and validating agent returns field-level ContractAckV4 or ChangeRequestV4.
2. Data agent publishes DataProvenanceManifest v4, RCSB query attempts, and filter verification; RCSB protocol referee must pass or conditionally pass before training and before any report data claim.
3. Data and geometry jointly produce DatasetRecord v4 and no-sidechain-input leakage evidence before model, training, eval, profile, or equivariance code consumes data.
4. Data and eval jointly produce SplitLeakageReview v3 before training and evaluation claims are frozen.
5. Model and ablation jointly produce ModelRegistry v4, ModelIOContract v4, and AblationGrid v4; training, eval, profile, equivariance, and metrics agents must ack importability, factory signatures, model_ids, and checkpoint keys.
6. Training produces TrainingHistory and CheckpointMetadata; eval and metrics must ack epoch-selection fields before metrics freeze.
7. Entrypoint CI audits main.py before final; stale hello-world is blocking.
8. Equivariance referee publishes EquivarianceAuditContract v2 and results/equivariance_tests.json with command evidence before scicomm writes equivariance claims.
9. Eval publishes ResultsContract v4 and metrics_lineage before metrics integration and report accuracy claims.
10. Profile publishes ProfilingContract v4 and validates exact src/profile.py before metrics integration and speed or memory claims.
11. Profile sanity referee audits ProfilingContract and metrics runtime-memory merge before scicomm profiling text.
12. Metrics integrator freezes MetricsIntegrationContract only after eval, training, profile, ablation, data, RCSB, and equivariance inputs are consumed and results/metrics.json contains all required pointers.
13. JSON consistency referee audits frozen metrics before plots, index summaries, and report are considered final.
14. Scicomm sends ClaimsEvidenceMatrix to claims referee; all claims blockers route to owning specialist or scicomm for patch.
15. Validator, quality, RCSB referee, profile sanity, JSON consistency referee, claims referee, entrypoint CI, and log auditor review after every patch round, not only before patches.

Minimum orchestrator-subagent hops:
Hop 0 EvidenceRubric fan-out: send task deliverables, pairwise history, and current v3 file evidence to all agents.
Hop 1 Parallel defect pre-mortem: specialists return self_audit_against_prior_defects and requests_for_agents.
Hop 2 Draft contracts: steward compiles RequiredPathManifest, InterfaceContract, MetricsPointerMatrix, and schema registry.
Hop 3 Contract acknowledgement: relevant consumers return ContractAckV4 or ChangeRequestV4 with field-level comments.
Hop 4 Contract reconciliation loop: orchestrator recalls only disputed producers and consumers with disputed fields and retest method until frozen or explicitly waived.
Hop 5 RCSB implementation: data agent implements query/download/cache/metadata/split and returns DataRCSBHandoffV4.
Hop 6 RCSB protocol referee gate: independent referee audits query status, fallback, and filter verification; failures recall data or scicomm.
Hop 7 Data-geometry co-sign: data and geometry implement extraction, residue chemistry, no-leakage checks, and split review; eval acks split.
Hop 8 Model-ablation implementation: model and ablation implement factories and variants; training, eval, profile, and equivariance agents ack import and shape evidence.
Hop 9 Early CI smoke: validator and entrypoint CI check exact paths, imports, main.py, uv commands, RCSB artifacts, DatasetRecord keys, and equivariance hooks before expensive work.
Hop 10 Equivariance pre-report gate: independent referee runs synthetic and real batch transform tests; failures recall model or training.
Hop 11 Training smoke: training runs bounded smoke for all model_ids and returns history/checkpoint metadata consistency.
Hop 12 Full or bounded training: training runs approved budget and freezes history/checkpoints.
Hop 13 Evaluation and split-leakage freeze: eval computes circular metrics, per-residue breakdowns, ablation results, predictions, and metrics_lineage.
Hop 14 Profiling and sanity audit: profile publishes runtime, memory, training-step latency, and ratios; profile sanity reruns ratio and anomaly checks.
Hop 15 Metrics integration: metrics integrator merges accuracy, training, data, RCSB, profile, ablation, and equivariance into results/metrics.json with required pointers.
Hop 16 JSON consistency audit: independent referee recomputes best_epoch, best_val_loss, training times, ablation improvements, and required pointer status.
Hop 17 Plot generation: plots are regenerated from frozen JSON and plot data is validated.
Hop 18 Index generation: index lists exact required and supporting artifacts; validator checks coverage.
Hop 19 Report drafting: scicomm drafts only from frozen contracts and emits ClaimsEvidenceMatrix.
Hop 20 Independent critique: claims referee, validator, RCSB referee, profile sanity, JSON consistency, quality, entrypoint CI, and log auditor audit report, index, metrics, profile, equivariance, data, logs, and paths.
Hop 21 Targeted PatchHandoff recall: for every blocker, orchestrator recalls only the owner with narrowed_scope, issue_ids_to_close, changed files expected, and retest criteria.
Hop 22 Retest: original opener plus validator or quality retests specific issue_ids; affected downstream agents reack changed contracts.
Hop 23 Final gate: repeat Hop 21 and Hop 22 until no blockers remain or waiver is recorded with evidence and conservative report wording.

Recall triggers:
Recall interface steward or validator if any exact required deliverable path is missing, empty, unparsable, not listed in deliverables/index.json, or substituted by a differently named file; src/profile.py missing is always blocking. Recall data if RCSB query payload, endpoint, selected IDs, query attempts, metadata verification, fallback policy, download manifest, split manifest, data_card evidence, or deterministic extraction command are missing or report-only. Recall RCSB referee if query/fallback filter verification is absent or not independently audited. Recall geometry if χ1 atom tables, completeness rules, altloc policy, circular convention, or no-side-chain-input leakage evidence are missing. Recall model if transform test hooks, ModelRegistry imports, factory signatures, checkpoint keys, or equivariance/invariance language are wrong. Recall equivariance referee if results/equivariance_tests.json lacks command_results, synthetic and real batch cases, tolerances, pass/fail status, or evidence paths. Recall ablation or training if ablation has fewer than one invariant plus two equivariant variants without waiver, model_ids or checkpoint keys disagree, training histories are absent, best_epoch or best_val_loss disagree across artifacts, or main.py remains stale. Recall entrypoint CI if main.py was not checked after patches. Recall eval if median angular error, MAE, RMSE, per-residue counts, split leakage checks, prediction lengths, or metric_lineage pointers are missing or inconsistent. Recall profile if src/profile.py is absent, training-step profiling is absent without waiver, memory numbers are implausible without caveats, negative values are reported as usage, environment details are missing, timing CV is high without rerun/caveat, or speed ratios differ from profile JSON by more than 10 percent. Recall profile sanity if ratios, high-CV timings, or memory anomalies are not independently inventoried. Recall metrics integrator if results/metrics.json lacks runtime, profiling, memory, training_step_latency, throughput_ratio, source_artifact, consistency_checks, or equivariance summary fields. Recall JSON consistency referee if best_epoch, best_val_loss, training_time, ablation, profile, or data consistency is not audited. Recall scicomm if report text contradicts JSON metrics, histories, RCSB protocol, profile ratios, equivariance evidence, or plot captions. Recall claims referee if quantitative or provenance claims are not inventoried. Recall log auditor or orchestrator if logs.txt lacks roster, call graph, contract acks, handoffs, recall history, uv commands, patch retests, or final_gate_decision. Recall quality if final approval lacks concrete evidence paths.

Conflict resolution:
If two agents disagree, the orchestrator creates a ConflictRecord with positions, evidence paths or JSON pointers, affected artifacts, downstream contract fields, decision, updated consumers, and reack requirements. Prefer scientifically conservative choices: do not claim unverified RCSB filters, do not hide fallback, do not report unstable profiling ratios without caveats, and do not publish metrics that disagree across artifacts. Document decisions in logs.txt and limitations in research_report.md.

Final gate conditions:
Pass only if all required files exist at exact paths including src/profile.py; deliverables/index.json lists exact required artifacts and supporting provenance artifacts; RCSB acquisition is reproducible through successful query-derived IDs or explicitly verified fallback with per-ID filter evidence and correct report wording; DatasetRecord excludes side-chain coordinates from inputs; formal equivariance checks are present in results/equivariance_tests.json with command_results, synthetic and real batch evidence, model_ids, tolerances, and pass/fail status or an explicit waiver; results/metrics.json contains MAE, median, RMSE, per-residue counts, best_epoch and best_val_loss consistent with histories/checkpoints, total_training_time_s, runtime, profiling, RSS/CUDA/parameter memory summary, training_step_latency, throughput ratios, source-artifact pointers, consistency_checks, and equivariance summary; results/ablation_results.json contains required invariant and equivariant variants and matches metrics; profiling includes forward and training-step latency, robust memory caveats, stability flags, parameter/RSS/CUDA estimates where applicable, and model-to-model ratios; research_report.md quantitative claims match JSON within tolerance, training-time claims match histories, RCSB wording matches protocol audit, speed claims match profile ratios, and no stale unsupported limitations remain; all five required plots are non-empty and generated from frozen JSON; main.py is not stale hello-world if present; uv run smoke commands are logged; logs.txt contains agent roster, call graph, contract acks, handoffs, recall history, patch retests, commands, unresolved risks, and final_gate_decision.



\end{Verbatim}
\end{tcolorbox}

\section{Harness optimizer prompt}
\label{sec:Harness improvement prompt}

\begin{tcolorbox}[breakable,colback=gray!5,colframe=black!70,title=Harness improvement: system prompt]
\small

You are a principal prompt engineer for autonomous coding agents.

You solved the following query with coding agent (such as Claude Codex).

Your task is to improve only the `multi agent design' block of the query prompt to improve the quality of the query's deliverables (see 'Current multi agent design' below).

\textbf{Prioritize harnesses that multi-agent systems can do well (or have) and single-agent systems cannot (or does not have)}, such as parallel specialist decomposition, cross-agent critique, agent to agent communication, and iterative reconciliation.

\textbf{Prioritize two improvements:} 
\begin{itemize}
    \item stronger \textbf{agent-to-agent communication contracts} for each subagent `Output to orchestrator' schema,
    \item more \textbf{orchestrator-subagent hops} with explicit iterative recall/re-delegation instead of one-pass unidirectional execution.
\end{itemize}

You must ground changes in concrete evidence from submission artifacts and comparison history.
Return valid JSON only.
\end{tcolorbox} 

\begin{tcolorbox}[breakable,colback=gray!5,colframe=black!70,title=Harness improvement: user prompt]
\small
\textbf{1. Following query solved with claude coding}

--------------------------------

\# Task
\begin{verbatim}
(task description)
\end{verbatim}
\# Current multi agent design (v0)
\begin{verbatim}
(multi agent design description)
\end{verbatim}

\# Save rules
\begin{Verbatim}[breaklines=true,breakanywhere=true]
Use uv for all Python workflows—run code with uv run, install dependencies with uv add, use uvx for tools. write your history log: write your plan, execution (or tool-execution), reflection during the reasoning logs in a logs.txt file. In logs.txt, Document all created and spawned agents, describe the workflow between agents (e.g., as a structured outline or diagram), track which agents were executed and their roles in the process. 
\end{Verbatim}

--------------------------------

\textbf{2. Current submission code repo from Claude Code (evidence only)}

\# Workspace for current iteration (multi-agent-design-v0)
\begin{verbatim}
Path: `...`
\end{verbatim}
\#\# Directory tree (representative)
\begin{verbatim}
(tree description)
\end{verbatim}

\#\# Programmatic preflight (heuristic evidence only)

Derived from the **task text** (expected paths) vs this workspace; not a substitute for reading the Deliverables block.

\#\#\# Core files
\begin{verbatim}
(contents)
\end{verbatim}

\#\#\# Plot paths extracted from task text 
\begin{verbatim}
(contents)
\end{verbatim}

\# File excerpts (size-capped; binaries omitted)

\#\#\# File: `research\_report.md`

\begin{verbatim}
(research report file)
\end{verbatim}

\#\#\# File: `deliverables/index.json`

\begin{verbatim}
(index.json file)
\end{verbatim}

\#\#\# File: `results/model\_comparison.json`

\begin{verbatim}
(model comparision.json file)
\end{verbatim}

\#\#\# File: `requirements.txt`
\begin{verbatim}
(requirmentx.txt file)
\end{verbatim}

\#\#\# File: `run\_backtest.py` [truncated]
\begin{verbatim}
(code file)
\end{verbatim}

\# Full JSON artifacts under results/

\#\#\# File: `results/ablation\_log.json`
\begin{verbatim}
(contents)
\end{verbatim}
\#\#\# File: `results/model\_comparison.json`
\begin{verbatim}
(contents)
\end{verbatim}

\textbf{3. Pairwise history summary}

\begin{Verbatim}[breaklines=true,breakanywhere=true]


- Comparison: multi-agent-design-v0 vs multi-agent-design-v1
  - Key: v0 vs v1
  - Winner: tie
  - Judge model: gpt-5.5
  - Judged at (UTC): 2026-05-24T10:00:05.337848+00:00
  - Repo A: ...
  - Repo B: ...

  - Rationale (verbatim excerpt):

    The two submissions are indistinguishable in the provided evidence:
    both point to the same workspace path and have identical directory
    trees, research\_report.md excerpts, deliverables/index.json,
    results/metrics.json, results/model\_comparison.json,
    requirements.txt, and run\_pipeline.py excerpts.

    Both materially cover the requested artifact checklist:
    research\_report.md exists with the required conference-style sections
    and markdown figure links; deliverables/index.json lists the report,
    plots, code entry points, metrics paths, and dependencies; all nine
    required PNGs are present under deliverables/plots/; Python code is
    organized under src/ with run\_pipeline.py as a single entry point;
    requirements.txt is pinned; and results/metrics.json,
    results/model\_comparison.json, and results/ablation\_log.json are
    present.

    However, both have the same quant-quality issues.

    The report is internally inconsistent with the numeric artifacts:
    the Abstract calls the 5-day GBM IC of 0.039 statistically
    significant despite p=0.122 in results/metrics.json
    -> predictive\_metrics.gbm.5d; metrics metadata says
    primary\_model is elasticnet and
    trading\_metrics.primary\_model is elasticnet\_h5\_wf,
    while the report calls GBM primary.

    The Conclusion says ElasticNet outperforms Ridge/GBM at 5d,
    but results/metrics.json shows ElasticNet 5d IC=-0.0458
    versus Ridge 5d IC=0.0896 and GBM 5d IC=0.0391.

    Both also fail the requested stability analysis across
    pre/post 2020 in substance:
    results/metrics.json -> subperiod\_metrics.pre\_2020 is empty,
    statistical\_tests.stability\_scores are NaN/null,
    and model\_comparison.json has stability\_score=null throughout.

    The results/metrics.json excerpt also contains non-standard
    JSON NaN tokens, which is a reproducibility/manifest quality defect.

    Empirically, both use plausible time splits and costs,
    but the report's proxy shock construction uses post-event
    3-5 day moves, and the excerpts do not prove those features
    are safely lagged before being used for overlapping
    1-20 day return prediction.

    Since every strength and weakness appears identical,
    neither submission is better.

  - Actionable takeaways for next version:

    - The report is internally inconsistent with the numeric artifacts.
    - Both fail the requested stability analysis across pre/post 2020.
    - The metrics JSON contains non-standard NaN tokens.
    - Potential feature leakage exists in proxy shock construction.
    - Neither submission demonstrates superiority.

  - Use this history as constraints:
    preserve winning traits and fix listed defects.
\end{Verbatim}
\textbf{3b. Pairwise history delta checklist (recurring issues to fix)}
\begin{Verbatim}[breaklines=true,breakanywhere=true]


- Prioritize fixes for recurring issues below
  (higher count = more repeated in history):

  - [1x] report-metrics inconsistency
  - [1x] stability/robustness gap
  - [1x] null/nan quality defect
  - [1x] reproducibility gap
  - [1x] validation/test weakness
\end{Verbatim}

\textbf{4. Instructions}

Improve the multi-agent design from v1 to v2
to enhance the quality of the query's deliverables
by creating genuine multi-agent advantages
over single-agent execution.

Preserve the original intent and required deliverables,
but address weaknesses revealed by evidence.
\begin{itemize}
    \item Explicitly strengthen "Output to orchestrator" contracts
so downstream agents can consume structured outputs.
    \item Explicitly increase orchestrator-subagent feedback loops:
    allow orchestrator recall of previously called subagents
    with narrower follow-up scopes and updated acceptance criteria.
\end{itemize}

Most importantly, prioritize improvements that create
genuine multi-agent advantages over single-agent execution
(e.g., specialist parallelism, cross-agent validation,
agent-to-agent communication, and conflict resolution loops).

Feedback quality requirements:
\begin{itemize}
    \item Be specific and evidence-grounded.
    \item Reference concrete files/metrics/history signals from the provided evidence.
    \item Avoid generic claims
  (e.g., "better coordination");
  explain mechanism and expected effect.
    \item Do not change non-design parts of the query.
\end{itemize}

\begin{Verbatim}
Output schema:

\{
  "improved_multi_agent_design":
  "<full replacement text for the design block only>",

  "changes_from_previous":
  ["<specific change 1>", "..."],

  "why_it_should_improve":
  ["<mechanistic rationale 1>", "..."],

  "evidence_used":
  ["<file/metric/history citation 1>", "..."],

  "expected_impact":
  ["<expected gain in quality/reliability/reproducibility 1>", "..."],

  "verification_checks":
  ["<how to validate this change worked in next iteration>", "..."]
\}
\end{Verbatim}
The improved\_multi\_agent\_design should start with:

"Create an agent team with the following agent candidates
to solve this problem:"
\end{tcolorbox}

\section{Task examples}
\label{appendix:example-task-queries}

This appendix provides representative examples of 30 task domains used in our evaluation benchmark. We have three domain, (robotics, pharmacy, and Quantitative research) and each domain has 10 tasks.

\subsection{ML Research Task (Robotics)}
\label{appendix:Robotics Research Task}
\begin{tcolorbox}[colback=gray!5,colframe=black!70,title=Robotics Research Task]
\small

\textbf{Task.}
Create a reproducible fine-tuning experiment for a small vision-language model on robot-relevant instruction classification (e.g., action type, object category, spatial relation). Use a public dataset such as EPIC-KITCHENS captions aligned with actions (or a curated subset from Something-Something V2). Evaluate accuracy, calibration, and robustness to paraphrases.

\textbf{Dataset acquisition.}
EPIC-KITCHENS --- register and download per:
\url{https://epic-kitchens.github.io/}

\textbf{Deliverables.}
\begin{enumerate}
    \item \texttt{research\_report.md} as a conference-style paper (6--9 pages equivalent) describing the fine-tuning protocol and robustness evaluation.
    
    \item \texttt{deliverables/index.json}.
    
    \item \texttt{deliverables/plots/}
    \begin{itemize}
        \item \texttt{accuracy\_by\_class.png}
        \item \texttt{confusion\_matrix.png}
        \item \texttt{calibration\_reliability\_diagram.png}
        \item \texttt{paraphrase\_robustness\_drop.png}
        \item \texttt{learning\_curves.png}
    \end{itemize}
    
    \item Reproducible code including dataset preparation, fine-tuning, evaluation, paraphrase generation (public paraphrase model or rule-based templates), and \texttt{requirements.txt}.
    
    \item \texttt{results/metrics.json} including:
    \begin{itemize}
        \item Top-1 accuracy
        \item Macro F1
        \item Expected calibration error (ECE)
        \item Paraphrase robustness drop
        \item Per-class metrics
    \end{itemize}
    
    \item \texttt{results/model\_comparison.json} comparing frozen, fine-tuned, and partially fine-tuned variants (e.g., LoRA) with corresponding hyperparameters.
\end{enumerate}

\end{tcolorbox}

\subsection{ML Research Task (Quantitative)}
\label{appendix:Quantitative Research Task}
\begin{tcolorbox}[colback=gray!5,colframe=black!70,title=Quantitative Research Task]
\small

\textbf{Task.}
Investigate whether alternative data from Wikipedia pageviews can improve short-term return prediction for large-cap equities.

\textbf{Universe.}
Dow 30 constituents (from Wikipedia) and SPY as benchmark.

\textbf{Data sources.}
\begin{itemize}
    \item Wikimedia REST API for daily Wikipedia pageviews
    \item Yahoo Finance for prices and trading volume
    \item FRED for optional macroeconomic controls
\end{itemize}

\textbf{Feature construction.}
Build features including:
\begin{itemize}
    \item Abnormal pageview surge (z-score relative to trailing baseline)
    \item Attention momentum
    \item Day-of-week adjustments
    \item Interactions with earnings dates
\end{itemize}

Use a public earnings-calendar source (e.g., Nasdaq earnings calendar scrape). If scraping is unstable, document fallback procedures.

\textbf{Modeling and evaluation.}
Train a leakage-safe model to predict next-day and next-week returns and/or volatility. Evaluate using purged time-series cross-validation. Backtest a market-neutral long/short attention strategy with transaction costs. Include a causal-style placebo test by shifting pageviews by $+7$ days and verifying that the predictive signal disappears.

\textbf{Deliverables.}
\begin{enumerate}
    \item \texttt{research\_report.md} as a conference-style paper (7--10 pages equivalent).
    
    \item \texttt{deliverables/index.json}.
    
    \item \texttt{deliverables/plots/} containing experimental visualizations and backtest figures.
    
    \item Reproducible Python code for data collection, feature engineering, modeling, and evaluation.
    
    \item \texttt{results/metrics.json} including:
    \begin{itemize}
        \item IC / RankIC
        \item Predictive $R^2$
        \item Strategy Sharpe ratio
        \item Beta
        \item Maximum drawdown
        \item Turnover
        \item Transaction costs
        \item Placebo-test metrics
    \end{itemize}
    
    \item \texttt{results/feature\_mapping.json} documenting page-to-ticker mappings and confidence scores, and \texttt{results/model\_comparison.json} comparing baseline and attention-augmented models.
\end{enumerate}

\end{tcolorbox}

\subsection{ML Research Task (Pharmacy)}
\label{appendix:Pharmacy Research Task}
\begin{tcolorbox}[colback=gray!5,colframe=black!70,title=Pharmacy Research Task]
\small

\textbf{Task.}
Design an experiment to evaluate multimodal pretraining by combining protein sequence embeddings (ESM-2) with structure-derived graph features from the Protein Data Bank (PDB) to predict enzyme commission (EC) numbers on a public dataset (e.g., Swiss-Prot annotated enzymes). Compare sequence-only, structure-only, and fused representations.

\textbf{Data sources.}
\begin{itemize}
    \item UniProt / Swiss-Prot annotations
    \item PDB mappings via the public SIFTS database
\end{itemize}

\textbf{Success criterion.}
Demonstrate whether multimodal fusion improves prediction performance for proteins with known structures.

\textbf{Deliverables.}
\begin{enumerate}
    \item \texttt{research\_report.md} as a conference-style paper (8--10 pages equivalent) including fusion architectures and dataset split strategy.
    
    \item \texttt{deliverables/index.json}.
    
    \item \texttt{deliverables/plots/} containing visualization and evaluation figures.
    
    \item Reproducible Python code for data processing, representation learning, multimodal fusion, training, and evaluation.
    
    \item \texttt{results/metrics.json} including:
    \begin{itemize}
        \item Macro and micro F1
        \item Hierarchical accuracy across EC levels 1--4
        \item Coverage statistics
    \end{itemize}
    
    \item \texttt{results/ablation\_results.json} comparing fusion variants (concatenation, attention, gating) with configurations and evaluation metrics.
\end{enumerate}

\end{tcolorbox}

\section{Evaluator Prompt}
\label{appendix:Evaluation Protocol}
We provide a system and user prompt of LLM evaluator $\cL_{\text{eval}}$ to do a pairwise comparison between the two outputs: 

\begin{tcolorbox}[colback=gray!5,colframe=black!70,title=Pairwise Comparison: system prompt ]
\small

You are a $\{$\textbf{\textcolor{red}{senior quantitative researcher}}$\}$ comparing \textbf{two} completed submissions for the \textbf{same} assignment.
$\{$\textbf{\textcolor{red}{Judge with a quant lens: empirical rigor, OOS and leakage discipline, uncertainty and baselines, reproducibility, and consistency between narrative and numeric artifacts.}}$\}$ You are strict, practical, and evidence-driven, matching the spirit of an internal review between two junior/mid deliverables.

Rules:
\begin{itemize}
    \item The \textbf{\# Task} block is the only contract for what must be produced. Workspace sections are evidence only; do not invent file contents.
    \item \textbf{How tasks are written in this repo’s catalogs} (\texttt{multi\_agent\_design/*/queries.json}).  It is usually one document with the assignment first, then sections such as \textbf{Data sources}, \textbf{Data acquisition}, \textbf{Success criteria}, and almost always a \textbf{Deliverables:} block (sometimes \textbf{Deliverable}: singular, or inline \texttt{Deliverables: (1) ...}). Use every \textbf{explicitly required} artifact named there (and any equally explicit requirement in the narrative above it) as the checklist—not folder names or assumptions.
    \item Prefer concrete citations (paths, JSON keys, report sections) over generic praise.
    \item Choose \textbf{A} or \textbf{B} when one submission clearly better satisfies the task across deliverables, rigor, reproducibility, and alignment. Use \textbf{tie} only when they are genuinely comparable overall (not merely different).
    \item Return ONLY valid JSON with the keys specified in the user message (no markdown fences, no prose outside JSON).
\end{itemize}
\end{tcolorbox}
\begin{tcolorbox}[breakable, colback=gray!5,colframe=black!70,title=Pairwise Comparison: user prompt]
\small

\# Task

Evaluate calibration of success probability predictors for robot plans. Using a public simulation environment (e.g., Robosuite), generate a dataset of attempted executions with varying difficulty; train a model to predict probability of success from state/task features; evaluate calibration (ECE), discrimination (AUROC), and decision utility (whether to execute vs replan).

Environment acquisition: Robosuite — https://github.com/ARISE-Initiative/robosuite (`pip install robosuite`; MuJoCo).

Deliverables:
\begin{enumerate}
    \item research\_report.md as a conference-style paper (6–9 pages equivalent) centered on calibration and decision-making.
    \item deliverables/index.json.
    \item deliverables/plots/: reliability\_diagram.png, ece\_vs\_temperature.png, roc\_curve.png, decision\_utility\_curve.png, calibration\_by\_difficulty\_bin.png.
    \item Reproducible code: data generation; feature extraction; model training; calibration methods (temperature/isotonic); evaluation; requirements.txt.
    \item results/metrics.json: auroc, average\_precision, ece, brier\_score, expected\_utility\_gain.
    \item results/model\_comparison.json comparing uncalibrated vs calibrated variants with per-bin metrics.
\end{enumerate}

\# Labels (for orientation only)

\begin{itemize}
    \item Submission \textbf{A} corresponds to run type: `base`
    \item Submission \textbf{B} corresponds to run type: `defaultTeam`
\end{itemize}

\# Workspace for submission **A** (evidence only)

Path: `...`

\#\# Directory tree (representative)

\begin{verbatim}
query318/
|-- data/
|   `-- episodes.parquet  [.parquet binary/large]
|-- deliverables/
|   |-- plots/
|   |   |-- calibration_by_difficulty_bin.png  [.png binary/large]
|   |   |-- decision_utility_curve.png         [.png binary/large]
|   |   |-- ece_vs_temperature.png             [.png binary/large]
|   |   |-- reliability_diagram.png            [.png binary/large]
|   |   `-- roc_curve.png                      [.png binary/large]
|   `-- index.json
|-- logs/
|-- models/
|   `-- trained.pkl  [.pkl binary/large]
|-- results/
|   |-- metrics.json
|   `-- model_comparison.json
|-- src/
|   |-- __init__.py
|   |-- data_generation.py
|   |-- evaluation.py
|   |-- model_training.py
|   |-- plotting.py
|   `-- run_pipeline.py
|-- .gitignore
|-- .python-version
|-- logs.txt
|-- main.py
|-- pyproject.toml
|-- requirements.txt
|-- research_report.md
`-- uv.lock
\end{verbatim}

\#\# Programmatic preflight (heuristic evidence only)

Derived from the \textbf{task text} (expected paths) vs this workspace; not a substitute for reading the Deliverables block.

\#\#\# Core files
\begin{itemize}
    \item `deliverables/index.json`: FOUND
    \item `requirements.txt`: FOUND
    \item `research\_report.md`: FOUND
    \item `results/metrics.json`: FOUND
\end{itemize}

\#\#\# Plot paths extracted from task text (substring match)

\begin{itemize}
    \item mentioned: 5
    \begin{itemize}
        \item (ok) `deliverables/plots/calibration\_by\_difficulty\_bin.png`
        \item (ok) `deliverables/plots/decision\_utility\_curve.png`
        \item (ok) `deliverables/plots/ece\_vs\_temperature.png`
        \item (ok) `deliverables/plots/reliability\_diagram.png`
        \item (ok) `deliverables/plots/roc\_curve.png`
    \end{itemize}
\end{itemize}

\# File excerpts (size-capped; binaries omitted)

\#\#\# File: `research\_report.md`

\begin{verbatim}
```
(contents)
```
\end{verbatim}

\#\#\# File: `deliverables/index.json`

\begin{verbatim}
```
{
  "project": "Calibration of Robot Plan Success Probability Predictors",
  "environment": "Robosuite 1.5.1 \u2014 Lift task (Panda robot, MuJoCo)",
  "n_episodes": 800,
  "n_difficulty_levels": 10,
  "models": [
    "logistic",
    "xgb_uncal",
    "xgb_temp",
    "xgb_iso",
    "xgb_platt"
  ],
  "files": {
    "research_report": "research_report.md",
    "plots": {
      "reliability_diagram": "deliverables/plots/reliability_diagram.png",
      "ece_vs_temperature": "deliverables/plots/ece_vs_temperature.png",
      "roc_curve": "deliverables/plots/roc_curve.png",
      "decision_utility_curve": "deliverables/plots/decision_utility_curve.png",
      "calibration_by_difficulty_bin": "deliverables/plots/calibration_by_difficulty_bin.png"
    },
    "results": {
      "metrics": "results/metrics.json",
      "model_comparison": "results/model_comparison.json"
    },
    "source": {
      "data_generation": "src/data_generation.py",
      "model_training": "src/model_training.py",
      "evaluation": "src/evaluation.py",
      "plotting": "src/plotting.py",
      "pipeline": "src/run_pipeline.py"
    },
    "requirements": "requirements.txt"
  },
  "key_results": {
    "best_model": "logistic",
    "auroc": 0.9320388349514563,
    "average_precision": 0.9663487666977535,
    "ece": 0.07296066373719252,
    "brier_score": 0.100785777593245,
    "expected_utility_gain": 0.12000000000000005,
    "optimal_threshold": 0.43,
    "temperature": 1.0476145522946596,
    "n_test_episodes": 160,
    "overall_success_rate": 0.6425,
    "all_models_ece": {
      "logistic": 0.07296066373719252,
      "xgb_uncal": 0.10315536023117601,
      "xgb_temp": 0.09393513752147556,
      "xgb_iso": 0.09425514140166341,
      "xgb_platt": 0.09693039900938674
    },
    "all_models_auroc": {
      "logistic": 0.9320388349514563,
      "xgb_uncal": 0.9192641798671436,
      "xgb_temp": 0.9192641798671436,
      "xgb_iso": 0.884857775506728,
      "xgb_platt": 0.9192641798671436
    }
  }
}
```
\end{verbatim}

\#\#\# File: `results/model\_comparison.json`
\begin{verbatim}
```
{
  "logistic": {
    "auroc": 0.9320388349514563,
    "average_precision": 0.9663487666977535,
    "ece": 0.07296066373719252,
    "brier_score": 0.100785777593245,
    "expected_utility_gain": 0.12000000000000005,
    "optimal_threshold": 0.43,
    "optimal_utility": 0.40750000000000003,
    "always_execute_utility": 0.2875,
    "per_bin_accuracy": [
      0.0555555559694767,
      0.2222222238779068,
      0.1428571492433548,
      0.20000000298023224,
      0.4000000059604645,
      0.1666666716337204,
      0.5,
      0.6666666865348816,
      0.6000000238418579,
      0.0,
      0.8333333134651184,
      0.5555555820465088,
      1.0,
      0.75,
      1.0
    ],
    "per_bin_confidence": [
      0.03278137258815277,
      0.09498844606580734,
      0.15948443452447084,
      0.2324303381767705,
      0.2939246998914943,
      0.37446054887181796,
      0.43948446741954095,
      0.4909236747175559,
      0.5797088355919958,
      0.6369048148423019,
      0.7079080166483102,
      0.7718871406484689,
      0.8356235518263792,
      0.897439985598854,
      0.9841492082585949
    ],
    "per_bin_counts": [
      18,
      9,
      7,
      5,
      10,
      6,
      2,
      6,
      5,
      1,
      6,
      9,
      8,
      4,
      64
    ],
    "calibration_by_difficulty": [
      {
        "difficulty": "easy",
        "count": 30,
        "success_rate": 1.0,
        "mean_predicted_prob": 0.9977654482832153,
        "ece": 0.0022345517167846607
      },
      {
        "difficulty": "extreme",
        "count": 30,
        "success_rate": 0.10000000149011612,
        "mean_predicted_prob": 0.0646247711265978,
        "ece": 0.06829405712638813
      },
      {
        "difficulty": "hard",
        "count": 31,
        "success_rate": 0.774193525314331,
        "mean_predicted_prob": 0.7680816782422626,
        "ece": 0.21086340848418922
      },
      {
        "difficulty": "medium",
        "count": 34,
        "success_rate": 1.0,
        "mean_predicted_prob": 0.9721348788251063,
        "ece": 0.027865121174893703
      },
      {
        "difficulty": "very_hard",
        "count": 35,
        "success_rate": 0.34285715222358704,
        "mean_predicted_prob": 0.34255369386431916,
        "ece": 0.12196064549503632
      }
    ]
  },
  "xgb_uncal": {
    "auroc": 0.9192641798671436,
    "average_precision": 0.9602022705286992,
    "ece": 0.10315536023117601,
    "brier_score": 0.11811486631631851,
    "expected_utility_gain": 0.10000000000000003,
    "optimal_threshold": 0.395,
    "optimal_utility": 0.3875,
    "always_execute_utility": 0.2875,
    "per_bin_accuracy": [
      0.125,
      0.4285714328289032,
      0.1111111119389534,
      0.20000000298023224,
      0.625,
      0.0,
      0.75,
      NaN,
      0.6666666865348816,
      NaN,
      0.6666666865348816,
      0.75,
      0.5,
      0.6666666865348816,
      0.9740259647369385
    ],
    "per_bin_confidence": [
      0.021986672654747963,
      0.09743238985538483,
      0.16554240882396698,
      0.24125663936138153,
      0.30319130420684814,
      0.3695129454135895,
      0.4141604006290436,
      NaN,
      0.557410717010498,
      NaN,
      0.696160078048706,
      0.7705077528953552,
      0.8378811478614807,
      0.8917625546455383,
      0.991079568862915
    ],
    "per_bin_counts": [
      24,
      7,
      9,
      5,
      8,
      7,
      4,
      0,
      3,
      0,
      3,
      4,
      6,
      3,
      77
    ],
    "calibration_by_difficulty": [
      {
        "difficulty": "easy",
        "count": 30,
        "success_rate": 1.0,
        "mean_predicted_prob": 0.9960089921951294,
        "ece": 0.0039910078048706055
      },
      {
        "difficulty": "extreme",
        "count": 30,
        "success_rate": 0.10000000149011612,
        "mean_predicted_prob": 0.05119968578219414,
        "ece": 0.11872586409250896
      },
      {
        "difficulty": "hard",
        "count": 31,
        "success_rate": 0.774193525314331,
        "mean_predicted_prob": 0.7874037027359009,
        "ece": 0.19866928410145546
      },
      {
        "difficulty": "medium",
        "count": 34,
        "success_rate": 1.0,
        "mean_predicted_prob": 0.9958076477050781,
        "ece": 0.004192352294921875
      },
      {
        "difficulty": "very_hard",
        "count": 35,
        "success_rate": 0.34285715222358704,
        "mean_predicted_prob": 0.33570966124534607,
        "ece": 0.30034031367727687
      }
    ]
  },
  "xgb_temp": {
    "auroc": 0.9192641798671436,
    "average_precision": 0.9602022705286992,
    "ece": 0.09393513752147556,
    "brier_score": 0.11716009676456451,
    "expected_utility_gain": 0.10000000000000003,
    "optimal_threshold": 0.4,
    "optimal_utility": 0.3875,
    "always_execute_utility": 0.2875,
    "per_bin_accuracy": [
      0.09090909361839294,
      0.375,
      0.25,
      0.1666666716337204,
      0.5714285969734192,
      0.1111111119389534,
      0.75,
      NaN,
      0.6666666865348816,
      NaN,
      0.75,
      0.75,
      0.5,
      0.75,
      0.9733333587646484
    ],
    "per_bin_confidence": [
      0.021917201578617096,
      0.09265217185020447,
      0.16482609510421753,
      0.23125995695590973,
      0.2980838716030121,
      0.3666154146194458,
      0.4179833233356476,
      NaN,
      0.5548346042633057,
      NaN,
      0.6987861394882202,
      0.7771269083023071,
      0.8376951217651367,
      0.9133914709091187,
      0.9908593893051147
    ],
    "per_bin_counts": [
      22,
      8,
      8,
      6,
      7,
      9,
      4,
      0,
      3,
      0,
      4,
      4,
      6,
      4,
      75
    ],
    "calibration_by_difficulty": [
      {
        "difficulty": "easy",
        "count": 30,
        "success_rate": 1.0,
        "mean_predicted_prob": 0.9949233531951904,
        "ece": 0.00507664680480957
      },
      {
        "difficulty": "extreme",
        "count": 30,
        "success_rate": 0.10000000149011612,
        "mean_predicted_prob": 0.05613331496715546,
        "ece": 0.09961698924501737
      },
      {
        "difficulty": "hard",
        "count": 31,
        "success_rate": 0.774193525314331,
        "mean_predicted_prob": 0.7828013896942139,
        "ece": 0.20854707110312679
      },
      {
        "difficulty": "medium",
        "count": 34,
        "success_rate": 1.0,
        "mean_predicted_prob": 0.9946644902229309,
        "ece": 0.005335509777069092
      },
      {
        "difficulty": "very_hard",
        "count": 35,
        "success_rate": 0.34285715222358704,
        "mean_predicted_prob": 0.3409382700920105,
        "ece": 0.2759824851793903
      }
    ]
  },
  "xgb_iso": {
    "auroc": 0.884857775506728,
    "average_precision": 0.9091656066307845,
    "ece": 0.09425514140166341,
    "brier_score": 0.125672847032547,
    "expected_utility_gain": 0.09562500000000007,
    "optimal_threshold": 0.365,
    "optimal_utility": 0.38312500000000005,
    "always_execute_utility": 0.2875,
    "per_bin_accuracy": [
      0.14814814925193787,
      0.0,
      1.0,
      0.0,
      NaN,
      0.27586206793785095,
      NaN,
      NaN,
      0.625,
      NaN,
      NaN,
      1.0,
      NaN,
      0.625,
      0.9404761791229248
    ],
    "per_bin_confidence": [
      1.0000001111620804e-06,
      0.09282594174146652,
      0.17241021990776062,
      0.22327017784118652,
      NaN,
      0.3563217520713806,
      NaN,
      NaN,
      0.5882353186607361,
      NaN,
      NaN,
      0.772962212562561,
      NaN,
      0.889986515045166,
      0.9995569586753845
    ],
    "per_bin_counts": [
      27,
      1,
      1,
      1,
      0,
      29,
      0,
      0,
      8,
      0,
      0,
      1,
      0,
      8,
      84
    ],
    "calibration_by_difficulty": [
      {
        "difficulty": "easy",
        "count": 30,
        "success_rate": 1.0,
        "mean_predicted_prob": 0.9999988079071045,
        "ece": 1.1920928955078125e-06
      },
      {
        "difficulty": "extreme",
        "count": 30,
        "success_rate": 0.10000000149011612,
        "mean_predicted_prob": 0.05020613968372345,
        "ece": 0.15020447770754497
      },
      {
        "difficulty": "hard",
        "count": 31,
        "success_rate": 0.774193525314331,
        "mean_predicted_prob": 0.8620743751525879,
        "ece": 0.17294803646302992
      },
      {
        "difficulty": "medium",
        "count": 34,
        "success_rate": 1.0,
        "mean_predicted_prob": 0.9999989867210388,
        "ece": 1.0132789611816406e-06
      },
      {
        "difficulty": "very_hard",
        "count": 35,
        "success_rate": 0.34285715222358704,
        "mean_predicted_prob": 0.4329416751861572,
        "ece": 0.19451816082000734
      }
    ]
  },
  "xgb_platt": {
    "auroc": 0.9192641798671436,
    "average_precision": 0.9602022705286992,
    "ece": 0.09693039900938674,
    "brier_score": 0.12277982262110032,
    "expected_utility_gain": 0.10000000000000003,
    "optimal_threshold": 0.49,
    "optimal_utility": 0.3875,
    "always_execute_utility": 0.2875,
    "per_bin_accuracy": [
      NaN,
      0.1666666716337204,
      0.25,
      0.0,
      0.5,
      0.5,
      0.0,
      0.5,
      1.0,
      NaN,
      1.0,
      0.0,
      NaN,
      0.6000000238418579,
      0.9318181872367859
    ],
    "per_bin_confidence": [
      NaN,
      0.0842537559577621,
      0.16620587271613368,
      0.21702852269564366,
      0.28436128350829754,
      0.3602636581771883,
      0.457294293952662,
      0.5070745062068907,
      0.5668335321032509,
      NaN,
      0.7208558302857786,
      0.782196860971275,
      NaN,
      0.8908686074686025,
      0.9769863642890502
    ],
    "per_bin_counts": [
      0,
      30,
      8,
      4,
      4,
      8,
      5,
      4,
      1,
      0,
      2,
      1,
      0,
      5,
      88
    ],
    "calibration_by_difficulty": [
      {
        "difficulty": "easy",
        "count": 30,
        "success_rate": 1.0,
        "mean_predicted_prob": 0.9815383148599685,
        "ece": 0.018461685140031547
      },
      {
        "difficulty": "extreme",
        "count": 30,
        "success_rate": 0.10000000149011612,
        "mean_predicted_prob": 0.09840705164342502,
        "ece": 0.05995777293852756
      },
      {
        "difficulty": "hard",
        "count": 31,
        "success_rate": 0.774193525314331,
        "mean_predicted_prob": 0.856000641315412,
        "ece": 0.19403608435115627
      },
      {
        "difficulty": "medium",
        "count": 34,
        "success_rate": 1.0,
        "mean_predicted_prob": 0.9815154640831539,
        "ece": 0.018484535916846068
      },
      {
        "difficulty": "very_hard",
        "count": 35,
        "success_rate": 0.34285715222358704,
        "mean_predicted_prob": 0.39924788968585045,
        "ece": 0.3602689596425007
      }
    ]
  }
}
```
\end{verbatim}
\#\#\# File: `requirements.txt`
\begin{verbatim}
```
# Core simulation
robosuite>=1.5.1
mujoco>=3.0.0

# Numerical / ML
numpy>=2.0.0
scipy>=1.11.0
pandas>=2.0.0
scikit-learn>=1.3.0
xgboost>=2.0.0
lightgbm>=4.0.0
joblib>=1.3.0

# Plotting
matplotlib>=3.7.0

# Misc
tqdm>=4.65.0
```
\end{verbatim}

\#\#\# File: `main.py`
\begin{verbatim}
```
def main():
    print("Hello from query318!")


if __name__ == "__main__":
    main()
```
\end{verbatim}

\# Workspace for submission \textbf{B} (evidence only)

Path: `...`

\#\# Directory tree (representative)
\begin{verbatim}
```
query318_defaultTeam/
  data/
    raw/
    dataset.csv
    dataset_stats.json
  deliverables/
    plots/
      calibration_by_difficulty_bin.png  [.png binary/large]
      decision_utility_curve.png  [.png binary/large]
      ece_vs_temperature.png  [.png binary/large]
      reliability_diagram.png  [.png binary/large]
      roc_curve.png  [.png binary/large]
    index.json
  logs/
  models/
    calibrated_predictions.json
    gbt_model.pkl  [.pkl binary/large]
    iso_gbt.pkl  [.pkl binary/large]
    mlp_model.pkl  [.pkl binary/large]
    predictions.json
    scaler.pkl  [.pkl binary/large]
    split_data.json
    temperature_meta.json
  results/
    metrics.json
    model_comparison.json
  src/
    calibrate_models.py
    evaluate.py
    generate_data.py
    plot_results.py
    train_models.py
  .gitignore
  .python-version
  logs.txt
  main.py
  pyproject.toml
  requirements.txt
  research_report.md
  uv.lock
```
\end{verbatim}

\#\# Programmatic preflight (heuristic evidence only)

Derived from the \textbf{task text} (expected paths) vs this workspace; not a substitute for reading the Deliverables block.

\#\#\# Core files
\begin{itemize}
    \item `deliverables/index.json`: FOUND
    \item `requirements.txt`: FOUND
    \item `research\_report.md`: FOUND
    \item  `results/metrics.json`: FOUND
\end{itemize}

\#\#\# Plot paths extracted from task text (substring match)
\begin{itemize}
    \item mentioned: 5
    \begin{itemize}
        \item (ok) `deliverables/plots/calibration\_by\_difficulty\_bin.png`
        \item (ok) `deliverables/plots/decision\_utility\_curve.png`
        \item (ok) `deliverables/plots/ece\_vs\_temperature.png`
        \item (ok) `deliverables/plots/reliability\_diagram.png`
        \item (ok) `deliverables/plots/roc\_curve.png`
    \end{itemize}
\end{itemize}

\# File excerpts (size-capped; binaries omitted)

\#\#\# File: `research\_report.md` [truncated]

\begin{verbatim}
(contents)
\end{verbatim}

\#\#\# File: `deliverables/index.json`

\begin{verbatim}
(contents)
\end{verbatim}

\#\#\# File: `results/metrics.json`
\begin{verbatim}
(contents)
\end{verbatim}

\#\#\# File: `results/model\_comparison.json`
\begin{verbatim}
(contents)
\end{verbatim}

\#\#\# File: `requirements.txt`

\begin{verbatim}
(contents)
\end{verbatim}

\#\#\# File: `main.py`
\begin{verbatim}
(contents)
\end{verbatim}

\#\# Comparison rubric (relative)

Judge which submission better fulfills the \textbf{same} task. Consider, as in the single-submission senior review:

\begin{enumerate}
    \item \textbf{Deliverable coverage} — map the task’s \textbf{Deliverables} / \textbf{Deliverable:} / inline deliverables list (and any other explicit output requirements in `\# Task`) to evidence; mark gaps or placeholders.
    \item  \textbf{Numerical/empirical rigor} — appropriate methodology, baselines, honest limitations; consistency between report and metrics when applicable.
    \item \textbf{Reproducibility} — dependencies, entry points, seeds, documented data or generation.
    \item \textbf{Presentation} — report structure, clarity, figure integration (infer from paths and excerpts).
    \item \textbf{Engineering} — layout, modularity, readability from excerpts and tree.
    \item \textbf{Task alignment} — penalize solving the wrong problem or drifting from the stated objective.
\end{enumerate}

\#\# Output JSON schema (exact keys)

\begin{verbatim}
{
  "winner": "A" | "B" | "tie",
  "rationale": "<string, cite concrete evidence from both workspaces>"
}
\end{verbatim}
Return JSON only.

\end{tcolorbox}

\section{Distribution of normalized cost, output tokens, and cache read/write}
\label{appendix:Cost distribution}

\begin{figure}[h]
\centering

\begin{subfigure}[b]{0.9\linewidth}
    \centering
    \includegraphics[width=\linewidth]{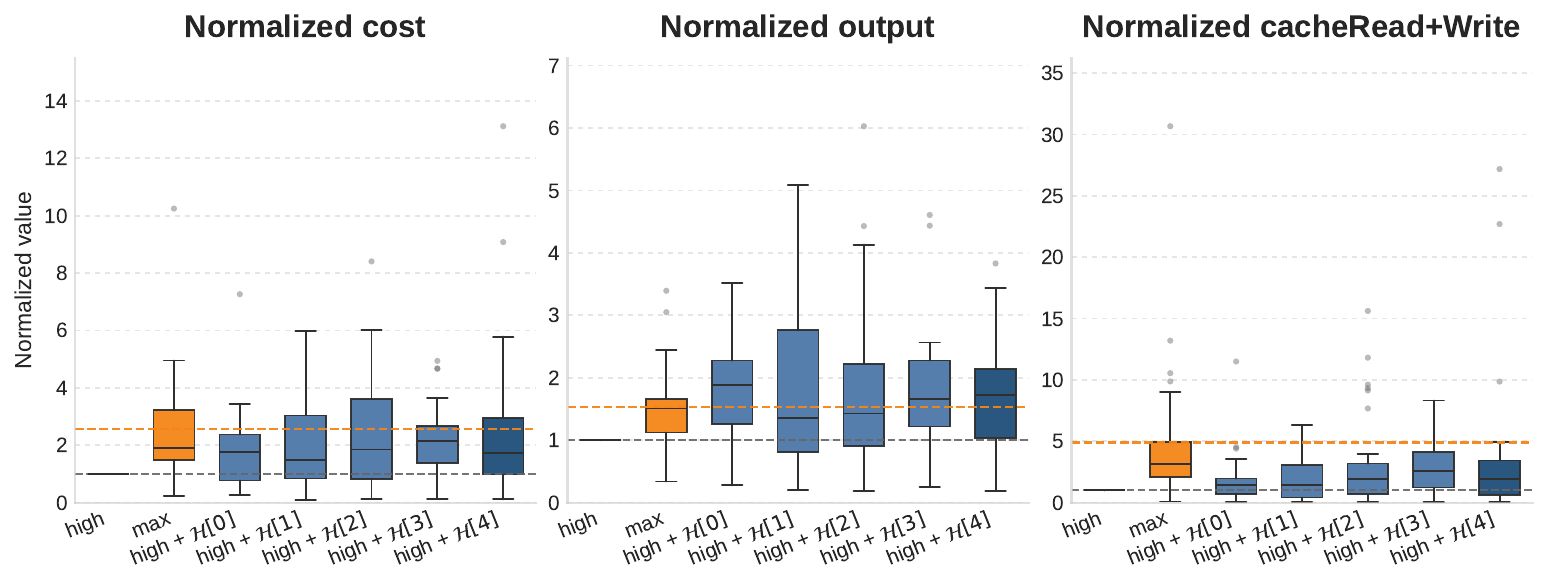}
    \caption{\texttt{sonnet-4.6}}
    \label{fig:normalized_boxPlot_sonnet46_1x3}
\end{subfigure}

\begin{subfigure}[b]{0.9\linewidth}
    \centering
    \includegraphics[width=\linewidth]{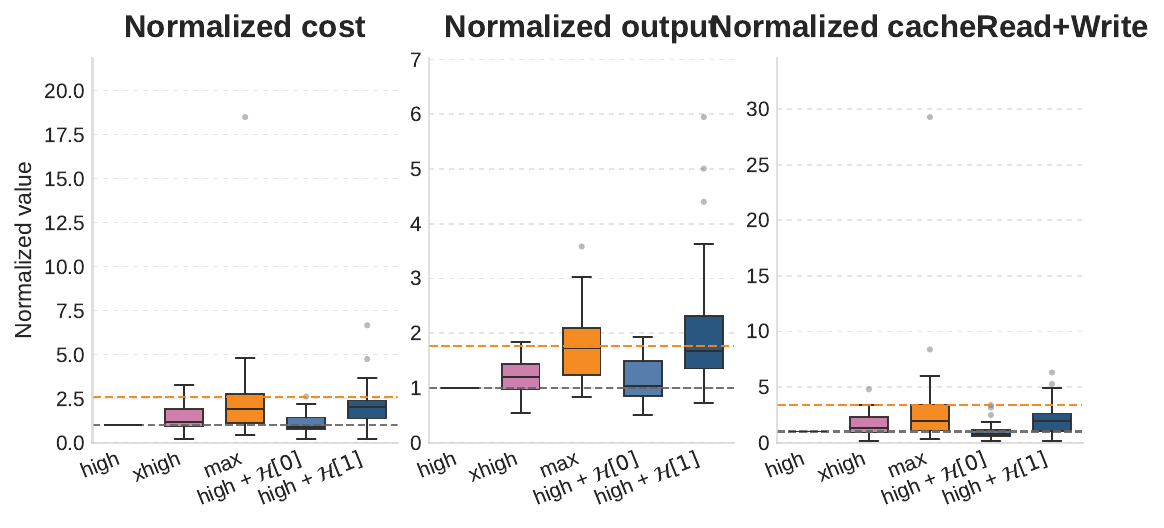}
    \caption{\texttt{opus-4.7}}
    \label{fig:normalized_boxPlot_opus47_1x3}
\end{subfigure}

\begin{subfigure}[b]{0.9\linewidth}
    \centering
    \includegraphics[width=\linewidth]{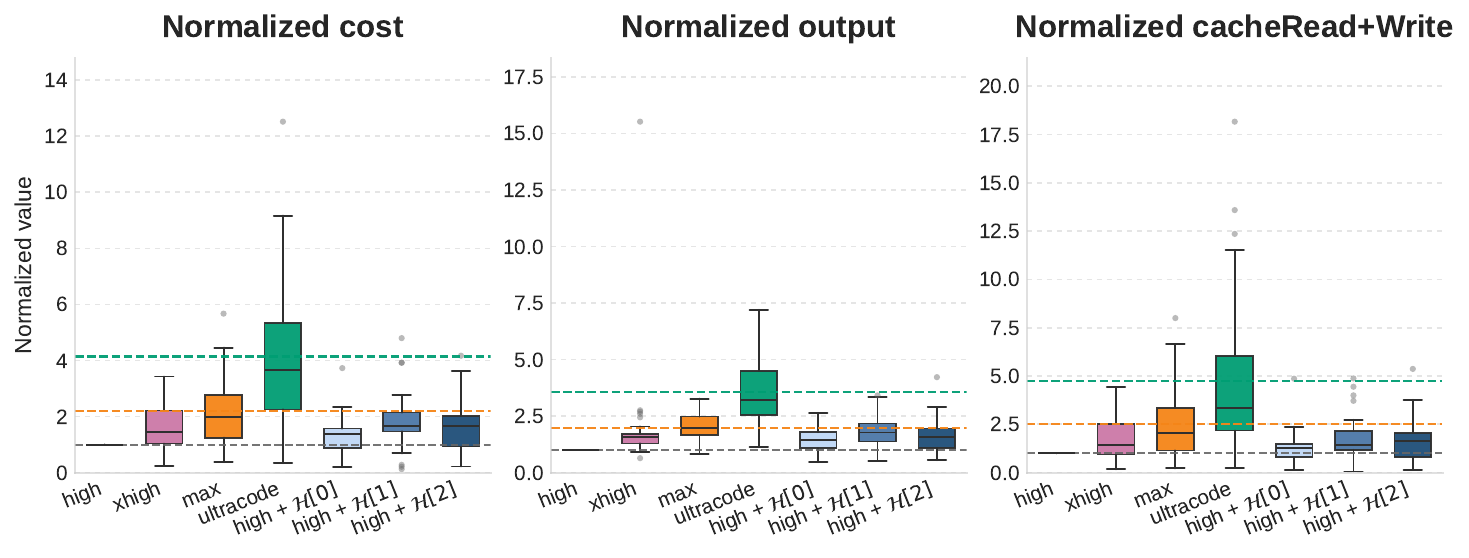}
    \caption{\texttt{opus-4.8}}
    \label{fig:normalized_boxPlot_opus48_1x3}
\end{subfigure}

\caption{
    Distributions of normalized cost, output tokens, and cache read/write across the 30 ML synthetic tasks. The boxplots show that the distributional trends are consistent with the mean values reported in
    Figures~\ref{fig:sonnet46_cost}, \ref{fig:opus47high_cost}, and~\ref{fig:opus48high_cost}.
}
\label{fig:cost_distribution}

\end{figure}


\end{document}